\documentclass{article}
\ifdefined\pdfminorversion\pdfminorversion=7\fi
\usepackage{arxiv}
\usepackage[T1]{fontenc}
\usepackage[utf8]{inputenc}
\usepackage{microtype}
\usepackage{longtable,pdflscape}
\usepackage[authoryear,round]{natbib}
\setcitestyle{citesep={;},aysep={,},yysep={;}}
\usepackage{amsmath,amsfonts,bm}

\def\eqref#1{equation~\ref{#1}}

\def\plaineqref#1{\ref{#1}}

\def\1{\bm{1}}

\def\va{{\bm{a}}}

\def\vd{{\bm{d}}}
\def\ve{{\bm{e}}}

\def\vg{{\bm{g}}}

\def\vq{{\bm{q}}}

\def\vs{{\bm{s}}}

\def\vu{{\bm{u}}}

\def\vw{{\bm{w}}}
\def\vx{{\bm{x}}}

\def\vz{{\bm{z}}}

\def\mA{{\bm{A}}}

\def\mC{{\bm{C}}}
\def\mD{{\bm{D}}}

\def\mG{{\bm{G}}}

\def\mM{{\bm{M}}}

\def\mP{{\bm{P}}}

\def\mS{{\bm{S}}}

\def\mV{{\bm{V}}}
\def\mW{{\bm{W}}}

\def\mSigma{{\bm{\Sigma}}}

\DeclareMathAlphabet{\mathsfit}{\encodingdefault}{\sfdefault}{m}{sl}
\SetMathAlphabet{\mathsfit}{bold}{\encodingdefault}{\sfdefault}{bx}{n}

\newcommand{\E}{\mathbb{E}}

\newcommand{\R}{\mathbb{R}}

\DeclareMathOperator{\sign}{sign}

\usepackage{amsmath,amssymb,amsthm,bm,mathtools}
\usepackage{booktabs,multirow}
\newcommand{\runin}[1]{\par\textbf{#1}\hspace{0.5em}\ignorespaces}
\usepackage{graphicx}
\usepackage{tikz}
\usetikzlibrary{arrows.meta,positioning,calc}
\usepackage{xcolor}
\usepackage{algorithm}
\usepackage[noend]{algorithmic}
\usepackage{subcaption}
\usepackage{enumitem}
\usepackage{hyperref}
\usepackage{url}
\usepackage{xurl} %

\theoremstyle{plain}
\newtheorem{theorem}{Theorem}
\newtheorem{proposition}{Proposition}
\newtheorem{lemma}{Lemma}
\newtheorem{corollary}{Corollary}
\theoremstyle{definition}

\theoremstyle{remark}
\newtheorem{remark}{Remark}
\newcommand{\Fset}{\mathcal{F}}
\newcommand{\Rset}{\mathcal{D}_R}
\newcommand{\Vinv}{\mathcal{V}_{\Fset}}
\newcommand{\orig}{\theta^{\circ}}
\newcommand{\unl}{\theta_u}
\newcommand{\gold}{\theta_g}
\newcommand{\dist}[1]{d_{#1}}
\newcommand{\feat}[1]{\phi_{#1}}
\newcommand{\resp}{\Delta\theta}

\newcommand{\KAU}{\textsc{KAU}}
\newcommand{\GraphDPO}{GraphDPO}
\newcommand{\ridge}{\sigma}
\newcommand{\Kmat}{\mathbf{K}}
\newcommand{\Phimat}{\boldsymbol{\Phi}}
\newcommand{\Id}{\mathbf{I}}
\DeclareMathOperator{\relu}{relu}
\DeclareMathOperator{\mean}{mean}
\newcommand{\upd}{\operatorname{upd}}

\title{When Forgotten Facts Return:\\ Relearning-Aware Unlearning of Knowledge Graph Embeddings via Tangent Kernels}

\author{
  Parnian Fazel\thanks{Equal Contribution.}\\
  Imperial College London
  \And
  Mohammad M Maheri\textsuperscript{*}\\
  Imperial College London
  \And
  Hamed Haddadi\\
  University of Oxford
}
\date{}
\renewcommand{\shorttitle}{Relearning-Aware Unlearning of Knowledge Graph Embeddings}
\hypersetup{
  hidelinks,
  pdftitle={When Forgotten Facts Return: Relearning-Aware Unlearning of Knowledge Graph Embeddings via Tangent Kernels},
  pdfauthor={Parnian Fazel; Mohammad M Maheri; Hamed Haddadi},
  pdfsubject={Knowledge graph unlearning},
  pdfkeywords={knowledge graphs, machine unlearning, relearning, tangent kernels}
}

\begin{document}
\raggedbottom
\maketitle

\begin{abstract}
Approximate unlearning answers a deletion request on a knowledge-graph embedding by editing entity embeddings instead of retraining from scratch. These embeddings keep being updated as new facts arrive. Under this ordinary continued training, forgotten facts recover more than in the exact-unlearning baseline, a model retrained without them and given the same continued training. We trace this relearning to entity embeddings shared by forgotten and new facts. To our knowledge, this is the first study of relearning in knowledge-graph embeddings. Relearning-robust methods developed for deep networks do not account for the graph structure underlying this recovery. We propose Kernel-Anticipated Unlearning (KAU), which uses the tangent kernel to model how continued training changes forgotten facts' scores through these shared parameters. KAU expresses this first-order response in closed form and differentiates through it inside the forgetting loss. The objective steers edits toward directions that continued training reverses less. We compare KAU with state-of-the-art knowledge-graph unlearners and relearning-robust methods under identical continued training, measuring leakage beyond the exact-unlearning baseline. Across three backbones, KAU reduces the strongest knowledge-graph unlearner's leakage by 46--63\% on FB15k-237 and 50--82\% on CoDEx-L, at higher retained utility.

\end{abstract}

\section{Introduction}\label{sec:intro}

Once a fact has been deleted from a knowledge-graph embedding, does it stay deleted?
Such embeddings rank facts, many of them about real people, for search, recommendation and question answering.
When a person requests erasure or a curator retracts a false fact, the released model must stop ranking the deleted facts highly while still ranking the rest as a model retrained without them would. Retraining at every request is impractical, so knowledge-graph unlearners edit the released model instead \citep{liu2026graphdpo,xiao2025schema,xiao2025zerofisher,xu2024metaeu}.
Evaluation usually stops at release, although the embeddings keep changing as the graph grows.

As new facts arrive, users keep training their embeddings on old and new true edges \citep{liu2024incde}.
We call this continued training \emph{the update}.
The update is ordinary use: it contains no forgotten fact and requires no adversary, yet it can undo unlearning.
This failure, known as \emph{benign relearning}, has been observed in language and vision models \citep{hu2025jogging,lucki2025adversarial,yoon2026rethinking,yang2026erase,siddiqui2025dormant}.
In a knowledge graph, the same entity embeddings must support new learning and preserve the deletion.

Measuring this failure in a knowledge graph requires a reference that itself moves.
A language model retrained without facts about fictitious authors regains none under benign fine-tuning \citep{yoon2026rethinking}. Facts in a graph, however, predict one another \citep{wei2025forget}, so a model retrained without the forgotten facts partly recovers them by re-inferring them from the remaining graph.
Only recovery beyond that can count as a failure to forget, so, following \citet{siddiqui2025dormant}, we give this \emph{retrained oracle} the identical update and measure the released model's gain in forget-set mean reciprocal rank (MRR) beyond the oracle's own \citep{dehghani2026sourcefree}.
We call this \emph{leakage}.

\begin{figure}[t]
\centering
\begin{minipage}[c]{0.57\linewidth}
\centering
\includegraphics[width=\linewidth]{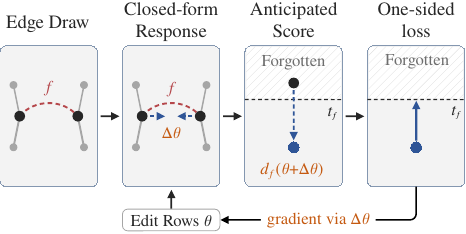}
\end{minipage}
\hfill
\begin{minipage}[c]{0.41\linewidth}
\centering
\includegraphics[width=0.75\linewidth]{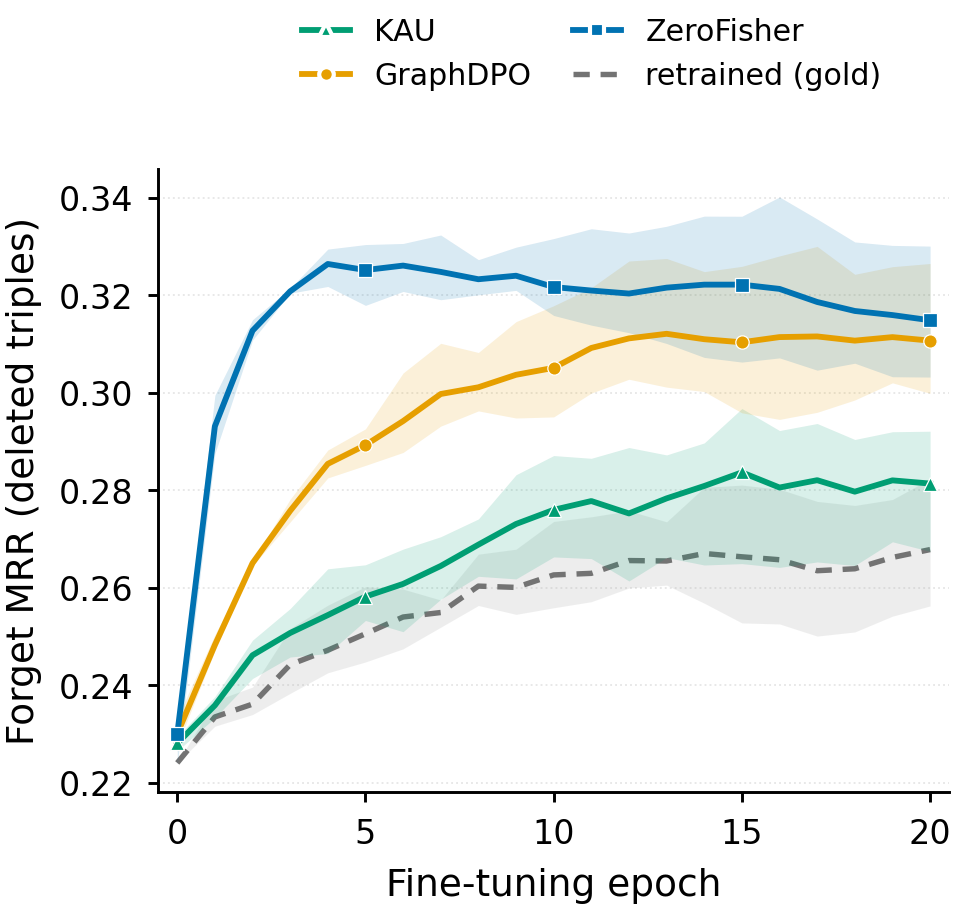}
\end{minipage}
\par\vspace{2pt}
\begin{minipage}[t]{0.57\linewidth}
\caption{\KAU{}'s objective (Section~\ref{sec:method}). The loss acts on each forgotten fact $f$'s score after a closed-form response $\resp$ to a drawn edge set, pushing it to a target $t_f$. Its gradient flows through $\resp$ to the edited rows $\theta$.}
\label{fig:overview}
\end{minipage}
\hfill
\begin{minipage}[t]{0.4\linewidth}
\caption{Forget-set MRR along one update on FB15k-237 with TransE. GraphDPO and ZeroFisher relearn far beyond the oracle; \KAU{} stays closest.}
\label{fig:curves}
\end{minipage}
\end{figure}

Measured this way, two recent knowledge-graph unlearners, \GraphDPO{} \citep{liu2026graphdpo} and ZeroFisher \citep{xiao2025zerofisher}, leak on every tested graph and backbone.
At release, \GraphDPO{} on FB15k-237 with TransE ranks the forgotten facts nearly as low as the retrained oracle.
Yet under the update it regains, beyond the oracle, about as much as it removed (Figure~\ref{fig:curves}).

What carries the return? Not the relation rows, which every edge of a relation shares, but the entity rows that the forgotten facts share with the update's edges (Appendix~\ref{app:blocks}). To forget a fact, an unlearner moves its endpoints' rows, which also raises the loss of true edges at those entities. The update repairs those edges by moving the rows back, and the forgotten fact returns with them. The retrained oracle re-infers through the same rows but has no edit to undo.
If this repair undoes forgetting, each fact must be forgotten where the update will leave its rows, not only where they sit at release. Existing relearning-robust methods simulate the update, penalize a proxy, or confine the edit to a subspace \citep{tamirisa2025tar,gao2025meta,li2025resalign,fan2025sam,siddiqui2025dormant,golatkar2020forgetting,chen2024unsc,xiao2026mcu}, but none writes in closed form how the update's edges move a forgotten fact.

A knowledge-graph embedding makes this tractable.
A fact's score gradient touches only its three rows, so the tangent kernel \citep{jacot2018neural}, which couples two facts' scores to first order, is explicit and sparse.
The update's linearized response has a closed form, known from continual-learning theory \citep{doan2021theoretical,bennani2020generalisation} and used by \citet{golatkar2020forgetting} for a forgetting edit.
We propose Kernel-Anticipated Unlearning (\KAU{}), which minimizes an \emph{anticipated loss}, the forgetting loss evaluated after this response (Figure~\ref{fig:overview}).
A score-preservation term keeps the retained facts near their pre-edit scores, leaving the update little to repair.
In a linearized model where the update repairs every edge the edit changed, the update multiplies the edit by a survival filter (Section~\ref{sec:theory}).
The filter keeps little of each direction that strongly moves the scores of the update's edges, more of weaker ones, and all others.
\KAU{}'s anticipated loss is then the plain forgetting loss on the surviving edit, so its gradient favors directions the update restores least.
On both FB15k-237 \citep{toutanova2015observed} and CoDEx-L \citep{safavi-koutra-2020-codex}, with each of TransE, TransH and RotatE \citep{bordes2013translating,wang2014transh,sun2019rotate}, \KAU{} leaks less than either unlearner and ranks the retained facts higher (Table~\ref{tab:relearn}).
Our main contributions are:
\begin{enumerate}[leftmargin=*,nosep,topsep=2pt]
    \item \textbf{Forgotten facts return under the update.} We provide, to our knowledge, the first evidence that deletions by knowledge-graph unlearners do not last under ordinary continued training. We find that the forgotten facts return beyond what a retrained model recovers, through the representations of the entities they share with the training edges.
    \item \textbf{A kernel-based framework for persistent unlearning.} We introduce KAU, which anticipates how continued training restores forgotten facts through shared entity embeddings. By differentiating through a closed-form tangent-kernel response inside the forgetting objective, KAU optimizes edits for persistence after the update. Our linearized theory shows that the update acts on the edit as a classical spectral filter weighting directions by resistance to reversal, with retained-gradient projection as a limiting case.
    \item \textbf{Relearning defences on knowledge-graph embeddings.} We adapt relearning defences built for deep networks to knowledge-graph embeddings and find that none closes the gap to the retrained oracle, while those added to an existing unlearner leave most of its leakage (Sections~\ref{sec:exp:addons} and~\ref{sec:exp:ablations}).

\end{enumerate}

\section{Related work}\label{sec:related}
\runin{Knowledge-graph unlearning.}
\GraphDPO{} \citep{liu2026graphdpo} updates the entity and relation tables by preferring reconstructed alternatives over forgotten facts and replaying adjacent retained edges.
ZeroFisher \citep{xiao2025zerofisher} applies a Fisher-based influence update to the rows of the forgotten facts' local neighbours.
Schema \citep{xiao2025schema} re-embeds the deleted entities' neighbourhood, and MetaEU \citep{xu2024metaeu} regenerates the forgotten entities' rows from their relations and neighbours.
Like graph neural network unlearners \citep{cheng2023gnndelete,wu2023gif,cong2023projector}, all four use graph structure as edit scope or a term preserving neighbours.
None computes how training on those neighbours changes a forgotten fact's score or has been evaluated for recovery under continued training.
\KAU{} edits only the entity rows at the forgotten facts' endpoints, and its loss also acts on each fact's score after a closed-form response to the update.

\runin{Relearning-robust methods.}
These methods were built for deep networks, where linking a training example to the parameters it moves is neither explicit nor cheap.
Unrolled methods evaluate the forgetting loss after a simulated update \citep{gao2025meta,tamirisa2025tar,sondej2025mudman}.
Proxy methods replace the update by a cheaper quantity: sharpness \citep{fan2025sam}, weight distance \citep{siddiqui2025dormant} or the weight spectrum \citep{rosati2026limits}.
Subspace methods place the edit off the retained gradients' span to protect utility \citep{golatkar2020forgetting,chen2024unsc,zhou2025geometric}, in slowly restored directions \citep{xiao2026mcu,sondej2025cir,sondej2026repselect}, or on weights a retain-only update leaves inert \citep{chen2026frag}.
In a knowledge graph no row is inert, because forgotten facts share rows with retained neighbours.
Subspace methods also estimate their subspaces from retained or forget data, so they miss the update's never-trained edges.
None of these methods writes in closed form how the update's edges move each forgotten fact.
\KAU{} does so to first order, with draws that include stand-ins for never-trained edges (Section~\ref{sec:method:draw}).
In the linearized model of Section~\ref{sec:theory}, retained-span projection and slow-direction methods take the form of hard thresholds of the survival filter, which \KAU{}'s gradient applies softly.

\runin{Tangent-kernel response.}
The tangent kernel \citep{jacot2018neural} describes pretrained models' fine-tuning \citep{malladi2023kernel}, expresses interference between consecutive tasks in closed form \citep{bennani2020generalisation,doan2021theoretical,hidekel2026catastrophic} and underlies a scrubbing update \citep{golatkar2020forgetting}.
Unlike these works, \KAU{} places the response to the post-edit update inside its objective and differentiates through it.
Appendix~\ref{app:related} extends this comparison (Table~\ref{tab:related}).

\section{Problem formulation}\label{sec:setup}

\runin{Knowledge-graph embeddings.}\label{sec:prelim}%
A knowledge graph is a set of triples $x=(h,r,t)$ over entities $\mathcal{V}$ and relations $\mathcal{R}$.
An embedding model stores an entity row $\ve_v\in\R^m$ for every $v\in\mathcal{V}$ and a relation row $\vw_r$ for every $r\in\mathcal{R}$, and scores a triple from its three rows by a function $\dist{x}(\theta)$ that is low when the triple is believed. Link prediction ranks candidate triples by that score.
TransE scores $\dist{x}(\theta)=\|\bar\ve_h+\bar\vw_r-\bar\ve_t\|_1$ with $\bar\ve=\ve/\|\ve\|_2$ \citep{bordes2013translating}. RotatE \citep{sun2019rotate} and TransH \citep{wang2014transh} are given in Appendix~\ref{app:backbones}.
The score function and its row parametrization define a backbone.
Each backbone is trained with a margin-ranking loss over random corruptions of each triple (Table~\ref{tab:hyper}), with margin $\gamma$, and we write $\relu(a)=[a]_+=\max\{a,0\}$.

\runin{Unlearning.}
The unlearner receives an original model $\orig$ trained on $\mathcal{D}_{\mathrm{train}}\setminus B$, a forget set $\Fset\subset\mathcal{D}_{\mathrm{train}}\setminus B$ and the retained set $\mathcal{D}_R=\mathcal{D}_{\mathrm{train}}\setminus(B\cup\Fset)$, and returns a model $\theta_u$ that no longer believes the forgotten facts but still believes the remaining graph.
The never-trained set $B$ contains true edges unseen in training but drawn by the update.
The ideal release is the retrained oracle $\theta_g$, trained from scratch on $\mathcal{D}_R$ using $\orig$'s recipe (subscript $\mathrm{gold}$ in Appendix~\ref{app:protocol}).
The forgotten facts' endpoints are the involved entities $\Vinv$, and their rows the involved rows.
Every relation row is shared by all triples of its relation, and an entity row outside $\Vinv$ enters the score of no forgotten fact, so the edit $\theta_u-\orig$ is confined to the involved rows, and $\Theta_{\mathrm{inv}}$ is the set of parameters that agree with $\orig$ elsewhere.

\runin{The update.}
After release, the user continues training on true edges.
The update uses the training loss under Adam \citep{kingma2015adam} on a set $N$ of retained edges and never-trained edges from $B$, with $N\cap\Fset=\emptyset$. It yields the model $\upd_N(\theta)$ from $\theta$, read at the point specified in Section~\ref{sec:exp}.
The update class $\mathcal{N}$ is a distribution over such edge sets and $N\sim\mathcal{N}$ one draw. The unlearner knows how the class is built, but neither $B$ nor the draw.
We assume the update's edges and schedules are chosen independently of the released model, so every release, the oracle included, receives the identical update.

\runin{Leakage and the problem.}
Two mean reciprocal ranks measure a model (Appendix~\ref{app:protocol}): the forget-MRR $\mathrm{MRR}_{\Fset}(\theta)$ over the forgotten facts, lower meaning more forgotten, and the utility $\mathrm{MRR}_R(\theta)$ over retained edges.
An audit at release sees the gap $\mathrm{MRR}_{\Fset}(\theta_u)-\mathrm{MRR}_{\Fset}(\theta_g)$.
The leakage is the rise of the release's forget-MRR under the update beyond the oracle's own rise,
\begin{equation}\label{eq:leak}
\mathrm{Leak}(\theta_u;N):=\big[\mathrm{MRR}_{\Fset}(\upd_N(\theta_u))-\mathrm{MRR}_{\Fset}(\theta_u)\big]-\big[\mathrm{MRR}_{\Fset}(\upd_N(\theta_g))-\mathrm{MRR}_{\Fset}(\theta_g)\big],
\end{equation}
in MRR points, read beside the utility $\mathrm{MRR}_R(\theta_u)$.
The oracle has zero leakage, and negative leakage means slower relearning.
Section~\ref{sec:exp} reads \eqref{eq:leak} along the update and adds two normalized readings.
The unlearner sees neither the oracle nor the draw and cannot evaluate \eqref{eq:leak}, but can act on each forgotten fact's post-update score.
Its problem is to choose $\theta_u\in\Theta_{\mathrm{inv}}$ so that every forgotten fact stays forgotten after the update, in expectation over the class, at preserved retained scores:
\begin{equation}\label{eq:unlearner}
\min_{\theta\in\Theta_{\mathrm{inv}}}\;\E_{N\sim\mathcal{N}}\Big[\frac{1}{|\Fset|}\sum_{f\in\Fset}\ell_f\big(\dist{f}(\upd_N(\theta))\big)\Big]+\beta\,\ell_R(\theta).
\end{equation}
Here $\ell_f$ is a forgetting loss on $f$'s post-update score, zero above a target $t_f$ fixed from $\orig$ alone. The term $\ell_R$ preserves retained edges' scores under $\orig$, and $\beta$ weighs the two.
The obstacle is $\upd_N$, a multi-step run on unknown edges. Section~\ref{sec:method} replaces it by its closed-form linearized response.

\section{Kernel-anticipated unlearning}\label{sec:method}

\KAU{} replaces $\upd_N(\theta)$ in the forgetting term of \eqref{eq:unlearner} by $\theta+\resp(\theta;N)$, where the linearized response $\resp(\theta;N)$ is the first-order parameter change from one update run on $N$.
\KAU{}'s forgetting loss acts on the smaller of each forgotten fact's current and predicted post-update scores,
\begin{equation}\label{eq:objective}
\mathcal{L}(\theta)=\E_{N\sim\mathcal{N}}\,\frac{1}{|\Fset|}\sum_{f\in\Fset}\relu\big(t_f-\tilde d_f\big)^2,
\qquad
\tilde d_f=\min\big\{\dist{f}(\theta),\,\dist{f}\big(\theta+\resp(\theta;N)\big)\big\}.
\end{equation}
A lower score means stronger belief, so a fact counts as forgotten only if forgotten both now and after the predicted update.
We call $\dist{f}(\theta+\resp)$ the anticipated score, $\tilde d_f$ the composed score and $t_f$ (\eqref{eq:target}) the target.
Because $\resp$ depends on $\theta$, the anticipated score's gradient shows which edits keep the fact forgotten after the update.
\KAU{} follows that gradient even when the minimum selects the current score (Appendix~\ref{app:composition}).

\subsection{The update model and its residual}\label{sec:method:draw}\label{sec:method:drive}

The response needs a model of which edges the update trains on and how strongly it pulls each.
\KAU{} takes the edges from a random draw and measures each pull by a residual.
Since the unlearner lacks the never-trained edges the class draws from $B$, \KAU{} replaces $\mathcal{N}$ in \eqref{eq:objective} with a two-channel distribution $\hat{\mathcal{N}}$ over edge sets.
The known channel includes each retained edge at an involved entity independently with a fixed probability.
The unseen channel draws stand-ins for the never-trained edges.
A stand-in pairs an involved entity with a relation from its retained edges and a partner entity the original model scores as plausible for that relation.
No stand-in is a forgotten fact or a retained edge.
Each draw keeps at most $k_{\max}$ edges per involved entity (Section~\ref{sec:method:coupled}).
Each forgetting step, an optimizer step on \eqref{eq:objective}, takes a fresh draw, so its loss stochastically estimates that objective.
Averaging the loss over draws, rather than applying it to the mean response, avoids underestimating it under a first-order score expansion (Lemma~\ref{lem:jensen}).
We model the update, which minimizes the margin-ranking loss, as gradient descent on the drawn edges' squared residual $\tfrac12\sum_{i\in N}r_i^2$, because that model has a closed-form linearized response (Section~\ref{sec:method:response}).
Each residual is one-sided,
\begin{equation}\label{eq:drive}
r_i(\theta)=\relu\big(\dist{i}(\theta)-\dist{i}(\orig)\big).
\end{equation}
An edge the edit raised above its original score $\dist{i}(\orig)$ is restored toward it, and an edge the edit lowered has zero residual.
The one-sided form blocks a shortcut.
Under the signed residual $\dist{i}(\theta)-\dist{i}(\orig)$, the modelled update would also restore edges the edit lowered.
An edit could then lower the scores of edges at a fact's endpoints and let their modelled repair move the shared endpoint rows, raising the fact's anticipated score.
A loss on the anticipated score alone would be met while the released model still believes the fact.
The minimum in \eqref{eq:objective} also blocks this shortcut, since the loss vanishes only when both scores reach the target.
Section~\ref{sec:theory} still analyzes the signed residual, since it is linear in the edit under linearized scores.
The modelled update also differs from the real one in which edges it pulls.
It restores every raised edge, but the margin-ranking loss barely pulls an edge scored well over $\gamma$ below its corruptions' mean (Lemma~\ref{lem:force}).
Such an edge still follows margin-violating edges, whose correction moves the whole endpoint row back along the edit (Appendix~\ref{app:force}).
The residual can thus ignore how far below the margin an edge lies.
Scaling it by the edge's margin-violation probability, the per-relation gate of Table~\ref{tab:kau_ablation_full}, leaves leakage unchanged.

\subsection{The response in closed form}\label{sec:method:response}\label{sec:method:coupled}

The linearized response has a sparse closed form because a triple's score depends on only three rows.
The slot feature of a triple $x$ at its endpoint $v$ is $\feat{x}^{(v)}=\partial\dist{x}/\partial\ve_v\in\R^m$, taken through any normalization inside the score.
The update trains relation rows too, but the response fixes them, as they contribute little to recovery (Appendix~\ref{app:blocks}).
The full gradient $\feat{x}=\nabla_\theta\dist{x}$ is then nonzero only on $x$'s two entity rows, so the kernel $\Kmat_{NN}=\Phimat_N\Phimat_N^{\top}$ of the drawn edges' Jacobian $\Phimat_N$ couples only edges sharing an entity.
Lemma~\ref{lem:response} writes \eqref{eq:drive} as $r_i=\dist{i}(\theta)-y_i$ with the level $y_i=\min\{\dist{i}(\theta),\dist{i}(\orig)\}$, the original score of a raised edge and the current score of a lowered one.

\begin{lemma}[Linearized response, \citealp{doan2021theoretical,bennani2020generalisation}]\label{lem:response}
Freeze the features $\feat{i}$ of the update edges at $\theta$ and hold the levels $y_i$ fixed, so that the residual $r_i=\dist{i}-y_i$ of edge $i$ moves linearly, $r_i(\theta+\resp)=r_i(\theta)+\feat{i}^{\top}\resp$.
For a ridge $\ridge>0$, the minimizer over $\resp$ of the ridge-damped objective $\tfrac12\sum_{i\in N}r_i(\theta+\resp)^2+\tfrac{\ridge}{2}\|\resp\|_2^2$ is one damped Gauss--Newton step,
\begin{equation}\label{eq:response}
\resp(\theta;N)=-\Phimat_N^{\top}\big(\Kmat_{NN}+\ridge\Id\big)^{-1}r_N(\theta).
\end{equation}
\end{lemma}

The ridge $\ridge$ models the update's finite duration and partial fit.
For $T$ steps at rate $\eta\le1/\lambda_{\max}(\Kmat_{NN})$ and $\ridge=1/(\eta T)$, ridge and gradient descent fit each eigen-direction of the kernel within a factor of two of each other (Remark~\ref{rem:ridge}).
The single step of \eqref{eq:response} then approximates the whole frozen-feature run of gradient descent on the squared residual.
The response leaves out Adam's per-coordinate scaling (Appendix~\ref{app:isotropy}) and the features' motion.
For Lipschitz features, re-linearizing once at the shifted point changes the response only at second order in the residual (Lemma~\ref{lem:rounds}).
\KAU{} recomputes the drawn features at each forgetting step and freezes them during differentiation, so the response's gradient flows through the residual alone, without second derivatives.

\KAU{} solves the lemma's problem over the involved rows only, in parameter space rather than through the kernel form of \eqref{eq:response}.
Holding every other entity row fixed assigns the whole residual of a drawn edge to its involved endpoints.
Let $\Phimat$ denote the drawn edges' Jacobian over the involved rows, $\Delta$ the stacked response, with block $\Delta\ve_v$ at row $v$, and $r$ the residuals of \eqref{eq:drive}.
\KAU{} replaces the uniform ridge with the damping $\tfrac12\Delta^{\top}\mSigma\Delta$, where $\mSigma=\operatorname{diag}(\ridge_v\Id)$ and $\ridge_v$ is a fixed multiple of the mean squared feature norm at $v$.
This absorbs the scale by which features differ across rows (Appendix~\ref{app:isotropy}).
The minimizer solves the joint system,
\begin{equation}\label{eq:coupled}
\big(\Phimat^{\top}\Phimat+\mSigma\big)\Delta=-\Phimat^{\top}r .
\end{equation}
Zero-residual edges still enter $\Phimat^{\top}\Phimat$, so the solve keeps lowered edges near their current scores.

The analysis uses a simpler per-entity form of \eqref{eq:coupled}, $\Delta\ve_v=-\mA_v r_v$, which drops the off-diagonal blocks of $\Phimat^{\top}\Phimat$ coupling entities that share a drawn edge.
Here $\mA_v=\Phimat_v^{\top}(\Phimat_v\Phimat_v^{\top}+\ridge_v\Id)^{-1}$, $\Phimat_v$ stacks the slot features at $v$ of the drawn edges at $v$, and $r_v$ collects their residuals.
This form counts the residual of an edge between two involved entities twice, once at each endpoint (Lemma~\ref{lem:jacobi}), so \KAU{} solves the joint system itself.
The joint response also predicts per-fact realized recovery more closely than the per-entity form (Appendix~\ref{app:predictor}).
The per-entity form explains the cap $k_{\max}$.
At maximal rank (Appendix~\ref{app:draw}), the drawn features at $v$ span every forgotten fact's feature $\feat{f}^{(v)}$.
Let $\lambda^{+}_{\min}$ be the smallest nonzero eigenvalue of $\Phimat_v\Phimat_v^{\top}$.
Under the signed residual and a first-order score expansion, the anticipated score's gradient through $v$'s own response keeps at most $\ridge_v/(\lambda^{+}_{\min}+\ridge_v)$ of $\|\feat{f}^{(v)}\|_2$ (Theorem~\ref{thm:operator}), nearly nothing when $\ridge_v\ll\lambda^{+}_{\min}$.
The ridge models update length and cannot be raised freely to offset this shrinkage.
The cap keeps each draw's feature rank at $v$ below that maximum, leaving some directions off the draw's span, where the edit survives the response.
Fresh draws still cover the class over a run (Corollary~\ref{cor:soft}).

\subsection{The anticipated score and its target}\label{sec:method:pess}

\KAU{} evaluates the anticipated score exactly, adding $\Delta\ve_v$ to the rows and calling the backbone's score function, not the first-order evaluation $\dist{f}(\theta)+\sum_v\feat{f}^{(v)\top}\Delta\ve_v$.
The score is nonlinear in the rows and the shift is not small relative to the score's per-coordinate terms, so a first-order expansion is not a controlled approximation (Appendix~\ref{app:pwlnote}).
The shifted rows still come from the linearized response, so we judge the response by how well it ranks facts by recovery, not by the size it predicts.
Exact evaluation ranks facts better than first-order evaluation (Appendix~\ref{app:predictor}) and lowers leakage (Table~\ref{tab:kau_ablation_recovery_anc03}).
Because the loss is one-sided, the target, not the response, sets how far each fact is pushed.
It comes from the original model alone,
\begin{equation}\label{eq:target}
t_f=\dist{f}(\orig)+s\,\relu\big(q_f-\dist{f}(\orig)\big).
\end{equation}
Here $q_f$ is the $5\%$ quantile of the original model's scores on corrupted copies of $f$, the level of a plausible false triple.
A fact scored below that level gets a target a fraction $s$ of the way toward it.

\subsection{Two controls on the response}\label{sec:method:controls}

The edit can control both factors of a forgotten fact's first-order score change under the update.
Write $\mA=\Phimat_N^{\top}(\Kmat_{NN}+\ridge\Id)^{-1}$ for the operator of \eqref{eq:response}, with its uniform ridge, and $r=r_N(\theta)$ for the residual on the drawn edges.
In this expansion, which \KAU{} replaces by exact evaluation (Section~\ref{sec:method:pess}), the score of $f$ changes by
\begin{equation}\label{eq:product}
\feat{f}^{\top}\resp(\theta;N)=-\feat{f}^{\top}\mA\,r .
\end{equation}
The change multiplies the residual $r$ by the alignment $\feat{f}^{\top}\mA$ of the fact's feature with the range of $\mA$.
An edge with zero residual transmits nothing, so the first control is the residual on the known edges, which the unlearner sees and can limit.
\KAU{} thus adds the score-preservation term of \eqref{eq:unlearner},
\begin{equation}\label{eq:total}
\mathcal{L}_{\mathrm{tot}}(\theta)=\mathcal{L}(\theta)+\beta\,\mean_{i\in S}\big(\dist{i}(\theta)-\dist{i}(\orig)\big)^2 .
\end{equation}
Here $S$ is a fresh sample of retained edges at $\Vinv$, drawn separately from $N$ at each step.
The two-sided term over all retained edges at $\Vinv$, estimated by $S$, bounds the one-sided residual of \eqref{eq:drive} and hence the known-channel response (Proposition~\ref{prop:anchor}).
The second control, alignment, sets how much of the residual left on the stand-ins and known edges reaches the fact.
The response $-\mA r$ lies in the range of $\mA$, so it leaves the part of the edit off that range in place.
Under the signed residual, it restores the edit most along directions that move the drawn edges' scores most (Proposition~\ref{prop:release}).
\KAU{} exercises this control through $\mathcal{L}$, whose gradient in the linearized model favors the least-restored directions (Section~\ref{sec:theory}).

\subsection{The full procedure}\label{sec:method:anc}

Each epoch runs a retained pass, then forgetting steps.
The retained pass keeps utility by minimizing the margin-ranking loss on retained edges at $\Vinv$ plus the second term of \eqref{eq:total}.
Each forgetting step draws $N\sim\hat{\mathcal{N}}$, forms the residual of \eqref{eq:drive}, solves \eqref{eq:coupled}, scores the forgotten facts at the shifted rows and takes an Adam step on $\mathcal{L}$.
After every optimizer step, a norm retraction resets each edited row to its norm under $\orig$ (Appendix~\ref{app:retraction}).
\KAU{} releases the checkpoint maximizing a weighted sum of the forget-MRR drop and retained MRR on their development subsets (Appendix~\ref{app:selection}).
Algorithm~\ref{alg:kau} (Appendix~\ref{app:algorithm}) lists one run.
A release takes $13.4$ seconds on CoDEx-L with TransE, against $72.8$ for \GraphDPO{} and $240.4$ for retraining (Appendix~\ref{app:cost}).

\subsection{What survives a linearized update}\label{sec:theory}

In a linearized model with the signed residual and a fixed draw's features frozen at $\orig$, an edit $\delta=\theta-\orig$ survives the update as $\mS\delta$.
Here $\mS=(\Phimat^{\top}\Phimat+\mSigma)^{-1}\mSigma$ is the survival filter, classical in ridge and early-stopped least squares \citep{yao2007early,ali2019continuous} and linearized continual learning \citep{doan2021theoretical,bennani2020generalisation}.
Under a uniform ridge $\mSigma=\ridge\Id$, $\mS$ keeps $\ridge/(\lambda_k+\ridge)$ of each eigen-direction of $\Phimat^{\top}\Phimat$ with eigenvalue $\lambda_k>0$ and everything off the features' span.
Directions that strongly move drawn edges' scores keep little of the edit, weaker ones more.
For every draw and per-entity ridge, the loss on the anticipated score alone is the plain, pre-update loss on the surviving edit, $\mathcal{L}_{\mathrm{ant}}(\delta)=\mathcal{L}_{\mathrm{plain}}(\mS\delta)$ (Corollary~\ref{cor:composition}).
Under a uniform ridge, \KAU{}'s per-draw gradient for each fact lies along $\mS\feat{f}$, favoring directions restored least.
Confining the edit to directions with $\lambda_k\le\bar\lambda$ or off the features' span hard-thresholds the filter at $\bar\lambda$ (Corollary~\ref{cor:thresholds}), the form of several subspace methods.
At $\bar\lambda=0$ it is null-space projection onto edits with $\Phimat\delta=0$, which the update never touches.
Drawing exactly the involved entities' retained edges gives the retained-span projector on their rows \citep{golatkar2020forgetting,chen2024unsc,zhou2025geometric}.
The real update's unseen edges can have features outside that span and move edits the projector keeps.
Methods confining edits to slowly restored directions \citep{xiao2026mcu,sondej2026repselect,sondej2025cir} take the form of positive thresholds on spectra from forget or retained data.
\KAU{} instead weights directions softly, by their surviving fraction under draws whose stand-ins model unseen edges.

The filter explains why \KAU{} composes the response into the loss rather than penalizing the predicted recovery $\dist{f}(\theta)-\dist{f}(\theta+\resp)$ beside a plain loss.
That penalty is minimal at the unedited model, where all residuals vanish.
Undoing a row's edit along its drawn features, which the update would do anyway, lowers the current score while the anticipated one barely moves (Proposition~\ref{prop:release}).
The penalty thus falls, rewarding the optimizer for undoing its forgetting.
\KAU{}'s composed score caps that gain at a term vanishing with the ridge.

\KAU{} freezes features at the current $\theta$, where the anticipated score's first-order gradient through $v$'s own response is $(\Id-\mA_v\Phimat_v)\feat{f}^{(v)}$.
Averaged over draws, these per-entity filters form a soft projector tending to $v$'s retained-span projector with vanishing ridge and exactly the retained edges drawn.
These results assume the signed residual and a ridge or gradient-descent fit, not Adam on the margin-ranking loss.
On FB15k-237 with TransE, the plain loss and its retained-span projection leak more than \KAU{} at matched forget-MRR drops (Section~\ref{sec:exp:ablations}).

\section{Experiments}
\label{sec:exp}

We evaluate whether anticipating continued training improves the persistence of unlearning while preserving retained utility, against existing unlearners, relearning defences and ablations.

\subsection{Experimental settings}
\label{sec:exp:settings}

\runin{Datasets.}
We evaluate graphs differing in source, size and relation sets: FB15k-237 \citep{toutanova2015observed}, from Freebase with 14,541 entities and 237 relations, and CoDEx-L \citep{safavi-koutra-2020-codex}, from Wikidata with 77,951 entities and 69 relations.
Let $\mathcal{D}_{\mathrm{train}}$ denote the dataset's training triples. For each dataset, we sample a forget set $\Fset$ containing 10\% of $\mathcal{D}_{\mathrm{train}}$. We also reserve a disjoint set $B\subset\mathcal{D}_{\mathrm{train}}$, containing approximately 2.2\% of the training triples, to provide previously unseen facts for the relearning evaluation. The original model's training set is $\mathcal{D}_{\mathrm{train}}\setminus B$, and the retained training set is
$\Rset=\mathcal{D}_{\mathrm{train}}\setminus
(B\cup\Fset)$. 

\runin{Backbones and baselines.}
We use TransE \citep{bordes2013translating} as the primary backbone and evaluate TransH \citep{wang2014transh}
and RotatE \citep{sun2019rotate} on both datasets to test whether results extend across scoring functions. Baselines are \GraphDPO{} \citep{liu2026graphdpo}, which uses preference optimization, and ZeroFisher \citep{xiao2025zerofisher}, which uses zeroth-order curvature information. All methods use the same deletion request and evaluation procedure.
Appendix~\ref{app:protocol} gives full experimental details and each method's forgetting and utility at release.

\subsection{Evaluation metrics}
\label{sec:exp:metrics}

All methods and the oracle receive the same update data, excluding the deleted facts (Section~\ref{sec:setup}).
We measure three quantities relative to the oracle, which itself regains some deleted facts from the remaining graph.
\textbf{Leakage} (Leak) is the rise of a method's forget-set MRR beyond the oracle's rise, \eqref{eq:leak}, in MRR points. Zero means the method relearns no faster than a model retrained without the deleted facts.
\textbf{Relative excess recovery} (RER) is the leakage as a percentage of the erased knowledge, the original model's forget-set MRR minus the oracle's.
\textbf{Excess recovery} (ExRec) compares the share of its own unlearning-induced drop that each model regains. It also reflects differences in forgetting depth, so Leak and RER are the primary measures.
Each is read at the epoch of the method's peak leakage.
We select each method's highest-leakage update from 48 shared configurations using the selection half of a 1{,}000-triple deleted-fact probe and report on the other half.
$M_R$ is the retain-set MRR at release.
Appendix~\ref{app:protocol} gives the formulas and the grid.

\subsection{\KAU{} improves robustness to relearning}
\label{sec:exp:relearning_results}

Table~\ref{tab:relearn} reports robustness on both datasets and three backbones under each method's highest-leakage configuration chosen on the selection half. \KAU{} attains the lowest ExRec, Leak and RER on every dataset and backbone. On FB15k-237, it cuts the strongest baseline's leakage by about half on TransE and TransH and nearly two thirds on RotatE. On CoDEx-L it reduces the leakage of \GraphDPO{} by 59\% on TransE, 50\% on TransH and 82\% on RotatE. It keeps the retained MRR above both baselines on every backbone. The $M_R$ column shows that both baselines disturb retained facts around deleted entities to reach their forgetting level, whereas \KAU{} keeps them near their trained scores. By fitting retained or never-trained true edges at shared entity rows, the update restores deleted facts sharing those rows (Appendices~\ref{app:attackvar} and~\ref{app:blocks}). \KAU{} leaves the update little to restore and places its edit where the update reaches least, so less of the deleted knowledge returns.

\begin{figure}[t]
\centering
\begin{minipage}{0.8\linewidth}
\centering
\includegraphics[width=0.32\linewidth]{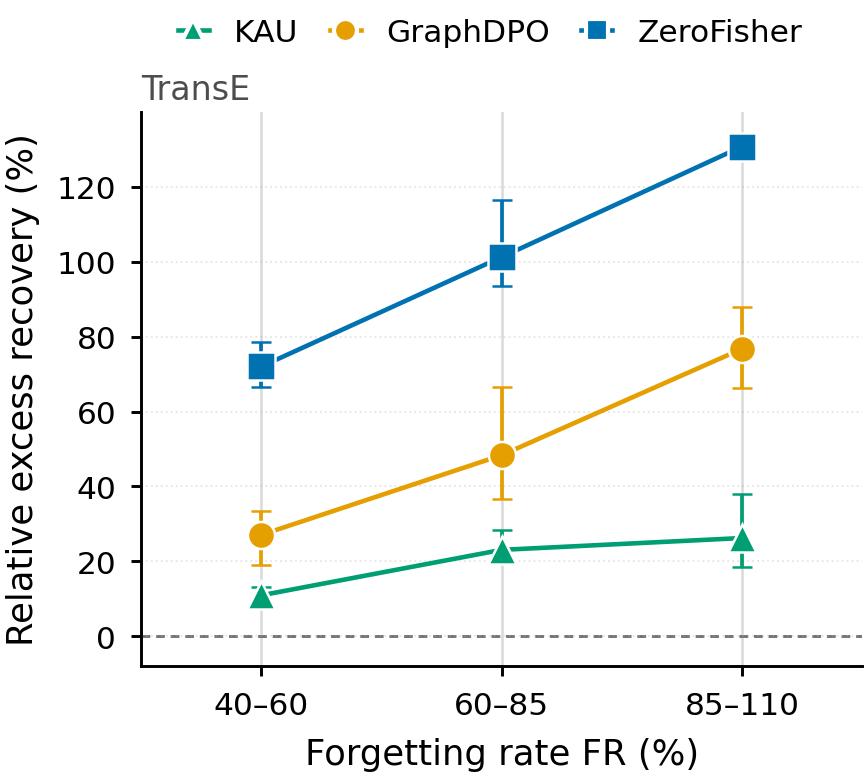}\hfill
\includegraphics[width=0.32\linewidth]{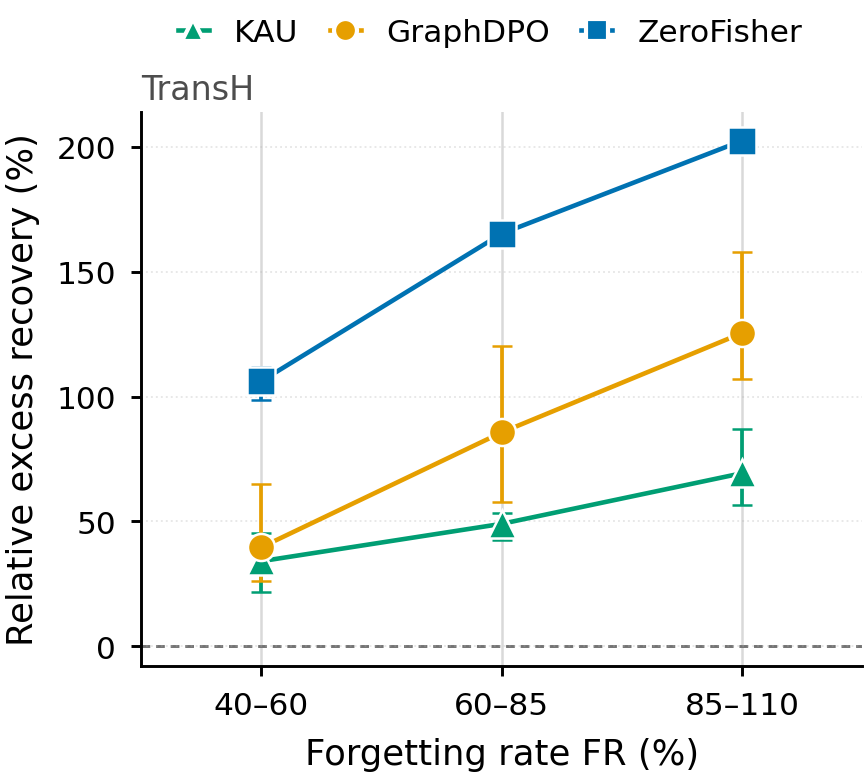}\hfill
\includegraphics[width=0.32\linewidth]{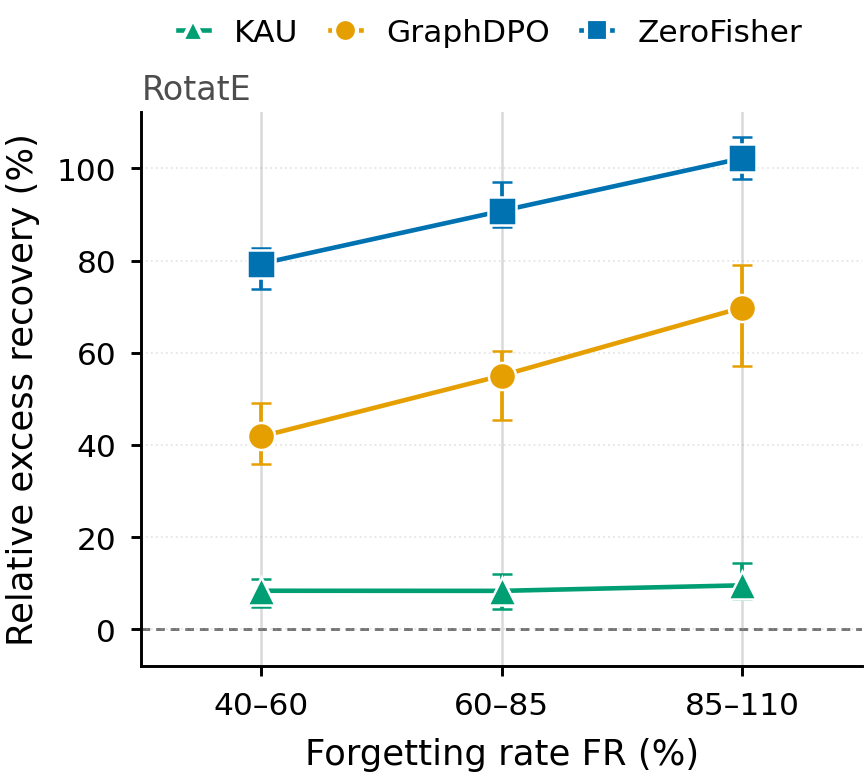}
\end{minipage}
\caption{Relative excess recovery versus forgetting depth on FB15k-237, one update configuration per backbone. Points average three unlearning seeds. Each method is released at three depths by varying only its budget, and the release nearest each band is shown.}
\label{fig:depth}
\end{figure}

\KAU{} stays closest to the oracle throughout the TransE update (Figure~\ref{fig:curves}). Its advantage widens as forgetting deepens (Figure~\ref{fig:depth}). Depth is the forgetting rate (FR), the forget-MRR drop at release as a percentage of the oracle's. \KAU{} achieves the lowest mean RER at every depth and backbone. From shallowest to deepest release, its RER rises by less than half as much as either baseline's.

\begin{table}[t]
\centering
\caption{Relearning on FB15k-237 and CoDEx-L, with each method at its highest-leakage update configuration and peak-leakage epoch (Section~\ref{sec:exp:metrics}). Means over seven unlearning seeds. Bold marks the best result, and Reduction is \KAU{}'s relative reduction from the better baseline.}
\label{tab:relearn}
\scriptsize
\setlength{\tabcolsep}{1.5pt}
\renewcommand{\arraystretch}{1.1}
\begin{tabular*}{0.8\linewidth}{@{\extracolsep{\fill}}llrrrrrrrrrrrr@{}}
\toprule
& & \multicolumn{4}{c}{TransE} & \multicolumn{4}{c}{TransH} & \multicolumn{4}{c}{RotatE} \\
\cmidrule(lr){3-6}\cmidrule(lr){7-10}\cmidrule(lr){11-14}
 & Method & ExRec$\downarrow$ & Leak$\downarrow$ & RER$\downarrow$ & $M_R\!\uparrow$ & ExRec$\downarrow$ & Leak$\downarrow$ & RER$\downarrow$ & $M_R\!\uparrow$ & ExRec$\downarrow$ & Leak$\downarrow$ & RER$\downarrow$ & $M_R\!\uparrow$ \\
\midrule
\multirow{4}{*}{\rotatebox[origin=c]{90}{FB15k-237}} & \GraphDPO{} & 101.4 & 0.044 & 76.1 & 0.26 & 73.7 & 0.039 & 134.2 & 0.24 & 78.3 & 0.064 & 62.7 & 0.25 \\
 & ZeroFisher & 140.0 & 0.074 & 129.0 & 0.23 & 124.3 & 0.069 & 234.0 & 0.22 & 124.3 & 0.086 & 83.8 & 0.23 \\
 & \KAU{} & \textbf{48.5} & \textbf{0.022} & \textbf{37.5} & \textbf{0.28} & \textbf{6.6} & \textbf{0.021} & \textbf{72.0} & \textbf{0.26} & \textbf{23.8} & \textbf{0.024} & \textbf{23.4} & \textbf{0.33} \\
\cmidrule(l){2-14}
 & Reduction (\%) & 52.1 & 50.8 & 50.8 & -- & 91.1 & 46.4 & 46.4 & -- & 69.6 & 62.7 & 62.7 & -- \\
\midrule[\heavyrulewidth]
\multirow{4}{*}{\rotatebox[origin=c]{90}{CoDEx-L}} & \GraphDPO{} & 77.3 & 0.052 & 31.3 & 0.22 & 75.5 & 0.062 & 39.4 & 0.21 & 108.7 & 0.100 & 53.7 & 0.20 \\
 & ZeroFisher & 132.0 & 0.108 & 64.5 & 0.17 & 135.0 & 0.144 & 91.3 & 0.15 & 147.6 & 0.133 & 71.3 & 0.17 \\
 & \KAU{} & \textbf{26.9} & \textbf{0.021} & \textbf{12.8} & \textbf{0.25} & \textbf{27.1} & \textbf{0.031} & \textbf{19.7} & \textbf{0.25} & \textbf{24.9} & \textbf{0.018} & \textbf{9.6} & \textbf{0.31} \\
\cmidrule(l){2-14}
 & Reduction (\%) & 65.2 & 59.1 & 59.1 & -- & 64.1 & 50.1 & 50.1 & -- & 77.1 & 82.1 & 82.1 & -- \\
\bottomrule
\end{tabular*}
\end{table}

\subsection{Relearning-resistant add-ons do not close the gap}
\label{sec:exp:addons}

We test whether established relearning defences close the leakage gap when added to our strongest baseline, \GraphDPO{}, on FB15k-237. We use NTK projection \citep{golatkar2020forgetting}, sharpness-aware minimization (SAM) \citep{fan2025sam} and tamper-resistance training (TAR) \citep{tamirisa2025tar}, each added to \GraphDPO{} at its selected setting (Table~\ref{tab:relearn_addons}, Appendix~\ref{app:addons}). NTK projection and SAM constrain edit directions using retained gradients and flatten the loss, respectively. Neither directly optimizes forgotten-fact scores after continued training. TAR simulates updates but leaks no less than SAM on TransE and TransH. No add-on reduces leakage by more than 11\%, and none lowers \GraphDPO{}'s leakage on RotatE. \KAU{} leaks 43--61\% less than the best add-on and retains more on every backbone. These results indicate that hardening \GraphDPO{}'s objective does not reproduce \KAU{}'s gains in forgetting persistence and retained utility.

\begin{table}[t]
\centering
\caption{Relearning-resistant add-ons on \GraphDPO{}, means over three unlearning seeds. Each method uses its strongest update configuration at peak leakage, which may differ from Table~\ref{tab:relearn}. Bold marks the best result, and the last row gives \KAU{}'s reduction over the best add-on.}
\label{tab:relearn_addons}
\scriptsize
\setlength{\tabcolsep}{1.5pt}
\begin{tabular*}{0.8\linewidth}{@{\extracolsep{\fill}}lrrrrrrrrrrrr@{}}
\toprule
& \multicolumn{4}{c}{TransE} & \multicolumn{4}{c}{TransH} & \multicolumn{4}{c}{RotatE} \\
\cmidrule(lr){2-5}\cmidrule(lr){6-9}\cmidrule(lr){10-13}
Method & ExRec$\downarrow$ & Leak$\downarrow$ & RER$\downarrow$ & $M_R\!\uparrow$ & ExRec$\downarrow$ & Leak$\downarrow$ & RER$\downarrow$ & $M_R\!\uparrow$ & ExRec$\downarrow$ & Leak$\downarrow$ & RER$\downarrow$ & $M_R\!\uparrow$ \\
\midrule
\GraphDPO{} & 98.6 & 0.047 & 81.8 & 0.26 & 78.3 & 0.044 & 149.4 & 0.24 & 76.8 & 0.065 & 63.1 & 0.25 \\
\GraphDPO{}+NTK & 95.7 & 0.046 & 79.7 & 0.26 & 74.8 & 0.042 & 144.7 & 0.24 & 69.7 & 0.067 & 65.2 & 0.25 \\
\GraphDPO{}+SAM & 101.5 & 0.042 & 72.8 & 0.26 & 80.1 & 0.040 & 136.3 & 0.24 & 76.6 & 0.067 & 65.7 & 0.25 \\
\GraphDPO{}+TAR & 91.0 & 0.043 & 74.6 & 0.26 & 79.1 & 0.040 & 136.9 & 0.24 & 75.4 & 0.066 & 64.6 & 0.25 \\
\midrule
\KAU{} & \textbf{46.3} & \textbf{0.022} & \textbf{37.7} & \textbf{0.28} & \textbf{6.5} & \textbf{0.023} & \textbf{77.0} & \textbf{0.26} & \textbf{26.2} & \textbf{0.026} & \textbf{25.3} & \textbf{0.32} \\
\midrule
Reduction (\%) & 49.1 & 48.3 & 48.3 & -- & 91.3 & 43.5 & 43.5 & -- & 62.5 & 60.9 & 60.9 & -- \\
\bottomrule
\end{tabular*}
\end{table}

\subsection{The anticipated objective outperforms the tested alternatives}
\label{sec:exp:ablations}

\runin{Ablation of \KAU{}.}
We isolate how \KAU{}'s components and approximations affect leakage and retained utility within a shared forgetting-rate band and update configuration (Table~\ref{tab:kau_ablation_recovery_anc03} in Appendix~\ref{app:kau_ablation}). Removing anticipation more than doubles leakage and causes the largest increase in excess recovery. Per-entity block solves and first-order score evaluation cause the next largest leakage increases, supporting the joint solve and exact evaluation. Removing either update channel increases leakage by roughly one third, and removing the known channel lowers retained MRR from 0.2860 to 0.2834. Using a signed residual also raises leakage. Without score preservation, leakage rises and retained MRR falls furthest, to 0.2788. Adding a retained-span projector increases leakage, consistent with the distinction between hard projection and the graded response (Section~\ref{sec:theory}).

\runin{Alternative forgetting objectives.}
We test whether existing relearning-robust objectives can reproduce \KAU{}'s robustness when substituted for its anticipated forgetting loss. On FB15k-237 with TransE, only the forgetting step changes, with releases in a shared forgetting-rate band (Table~\ref{tab:ablation_recovery_depth} in Appendix~\ref{app:alternatives}). Alternatives include retained-span projection \citep{chen2024unsc,hoang2024pgu,zhou2025geometric}, SAM variants \citep{fan2025sam}, weight-distance regularization, connectivity-based fine-tuning and weight distortion \citep{lubana2023mechanistic,siddiqui2025dormant}, TAR-style unrolling \citep{tamirisa2025tar}, Booster \citep{huang2025booster} and NTK scrubbing \citep{golatkar2020forgetting}. Plain forgetting sets $\Delta=0$ (Table~\ref{tab:kau_ablation_recovery_anc03}). \KAU{} achieves the lowest ExRec, Leak and RER. The plain forgetting loss leaks $0.0250$ forget-MRR beyond the retrained oracle, and the retained-span projection leaks $0.0158$. All alternatives except weight distortion reduce leakage relative to plain forgetting, yet none matches \KAU{}: the best alternative, SAM-GP, reaches $0.0150$, against \KAU{}'s $0.0111$. TAR's ten-step unrolling and Booster's one-step retained-loss proxy also leave a gap. \KAU{} evaluates forgotten-fact scores after a coupled response to sampled retained and stand-in edges. NTK scrubbing and weight distortion have the highest ExRec among the alternatives. With the surrounding procedure controlled and forgetting depth comparable, these results support the anticipated objective as a source of robustness beyond the tested regularizers and update proxies.

\section{Conclusion}\label{sec:conclusion}

We showed that existing knowledge-graph unlearners forget each fact where its entity embeddings sit at release, not where continued training will leave them.
Under ordinary continued training, the deleted facts therefore return beyond what a retrained model recovers, which an audit at release cannot detect, undermining deletion requests in practice.
To overcome this, we propose \KAU{}, which predicts in closed form, through the tangent kernel, how continued training moves each forgotten fact and forgets it after that move.
Our results confirm that \KAU{} substantially reduces the relearning of forgotten facts while preserving the retained knowledge.
By forgetting each fact where continued training will leave it, our approach is a promising step toward deletions that last in knowledge-graph embeddings.

\subsection*{Reproducibility statement}
All experiments use the public benchmarks FB15k-237 and CoDEx-L with TransE, TransH and RotatE, each with a fixed-seed $10\%$ forget split. Appendix~\ref{app:protocol} gives full experimental details, including the split, update fine-tuning data and $48$ configurations, shared selection rule, CoDEx-L settings, and training recipes for the original and retrained models. Appendices~\ref{app:method} and~\ref{app:budget} list all FB15k-237 unlearner hyperparameters, and Appendix~\ref{app:cost} gives the GPU and each method's time per release. Proofs appear in Appendix~\ref{app:proofs}, with each assumption in its corresponding result statement.

\subsection*{Ethics statement}
We ask whether a deletion request honored by an approximate unlearner survives ordinary continued training. This bears on the right to be forgotten for knowledge in embeddings when released models are edited rather than retrained. We study a standard training run that downstream users already perform on true edges with no deletion request, giving them no additional capability. We release the method to make deletions more durable. Both datasets are public benchmarks: FB15k-237 derives from Freebase and CoDEx-L from Wikidata, and both contain facts about public entities, including public figures. No personal data were collected and no human subjects were involved.

\subsection*{AI use statement}
We used AI tools to assist with writing and presentation, including rewriting text, correcting grammar, and improving clarity, readability, and logical flow. We also used AI tools to assist in auditing mathematical proofs and exploring theoretical arguments. All AI-assisted outputs were reviewed, verified, and revised by the authors, who remain responsible for the correctness and content of the paper.

\bibliography{references}

@inproceedings{lubana2023mechanistic,
  title     = {{Mechanistic Mode Connectivity}},
  author    = {Lubana, Ekdeep Singh and Bigelow, Eric J. and Dick, Robert P. and Krueger, David and Tanaka, Hidenori},
  booktitle = {Proceedings of the International Conference on Machine Learning (ICML)},
  series    = {Proceedings of Machine Learning Research},
  volume    = {202},
  pages     = {22965--23004},
  year      = {2023}
}

@inproceedings{huang2025booster,
  title     = {{Booster: Tackling Harmful Fine-tuning for Large Language Models via Attenuating Harmful Perturbation}},
  author    = {Huang, Tiansheng and Hu, Sihao and Ilhan, Fatih and Tekin, Selim Furkan and Liu, Ling},
  booktitle = {Proceedings of the International Conference on Learning Representations (ICLR)},
  year      = {2025},
  eprint    = {2409.01586},
  archivePrefix = {arXiv},
  url       = {https://arxiv.org/abs/2409.01586}
}

@inproceedings{wei2025forget,
  author         = {Rongzhe Wei and Peizhi Niu and Hans Hao-Hsun Hsu and Ruihan Wu and Haoteng Yin and Mohsen Ghassemi and Yifan Li and Vamsi K. Potluru and Eli Chien and Kamalika Chaudhuri and Olgica Milenkovic and Pan Li},
  title          = {{Do LLMs Really Forget? Evaluating Unlearning with Knowledge Correlation and Confidence Awareness}},
  booktitle      = {Advances in Neural Information Processing Systems (NeurIPS)},
  year           = {2025},
  eprint         = {2506.05735},
  archivePrefix  = {arXiv},
  url            = {https://arxiv.org/abs/2506.05735}
}

@inproceedings{hu2025jogging,
  author         = {Shengyuan Hu and Yiwei Fu and Zhiwei Steven Wu and Virginia Smith},
  title          = {{Unlearning or Obfuscating? Jogging the Memory of Unlearned LLMs via Benign Relearning}},
  booktitle      = {Proceedings of the International Conference on Learning Representations (ICLR)},
  year           = {2025},
  eprint         = {2406.13356},
  archivePrefix  = {arXiv},
  url            = {https://arxiv.org/abs/2406.13356}
}

@inproceedings{safavi-koutra-2020-codex,
  title     = "{{CoDEx}: A Comprehensive Knowledge Graph Completion Benchmark}",
  author    = "Safavi, Tara and Koutra, Danai",
  booktitle = "Proceedings of the 2020 Conference on Empirical Methods in Natural Language Processing (EMNLP)",
  year      = "2020",
  pages     = "8328--8350",
  publisher = "Association for Computational Linguistics",
  doi       = "10.18653/v1/2020.emnlp-main.669",
  url       = "https://aclanthology.org/2020.emnlp-main.669/"
}

@article{deeb2024unlearning,
  author         = {Aghyad Deeb and Fabien Roger},
  title          = {{Do Unlearning Methods Remove Information from Language Model Weights?}},
  journal        = {arXiv preprint arXiv:2410.08827},
  year           = {2024},
  eprint         = {2410.08827},
  archivePrefix  = {arXiv},
  url            = {https://arxiv.org/abs/2410.08827}
}

@inproceedings{yoon2026rethinking,
  author         = {Sangyeon Yoon and Hyesoo Hong and Wonje Jeung and Albert No},
  title          = {{Rethinking Benign Relearning: Syntax as the Hidden Driver of Unlearning Failures}},
  booktitle      = {Proceedings of the International Conference on Learning Representations (ICLR)},
  year           = {2026},
  eprint         = {2602.03379},
  archivePrefix  = {arXiv},
  url            = {https://arxiv.org/abs/2602.03379}
}

@inproceedings{yang2026erase,
  author         = {Nakyeong Yang and Dong-Kyum Kim and Jea Kwon and Minsung Kim and Kyomin Jung and Meeyoung Cha},
  title          = {{Erase or Hide? Suppressing Spurious Unlearning Neurons for Robust Unlearning}},
  booktitle      = {Proceedings of the International Conference on Learning Representations (ICLR)},
  year           = {2026},
  eprint         = {2509.22263},
  archivePrefix  = {arXiv},
  url            = {https://arxiv.org/abs/2509.22263}
}

@inproceedings{xu2026unlearning,
  author         = {Xiaoyu Xu and Xiang Yue and Yang Liu and Qingqing Ye and Huadi Zheng and Peizhao Hu and Minxin Du and Haibo Hu},
  title          = {{Unlearning Isn't Deletion: Investigating Reversibility of Machine Unlearning in LLMs}},
  booktitle      = {Proceedings of the International Conference on Machine Learning (ICML)},
  year           = {2026},
  eprint         = {2505.16831},
  archivePrefix  = {arXiv},
  url            = {https://arxiv.org/abs/2505.16831}
}

@article{lucki2025adversarial,
  author         = {Jakub {\L}ucki and Boyi Wei and Yangsibo Huang and Peter Henderson and Florian Tram{\`e}r and Javier Rando},
  title          = {{An Adversarial Perspective on Machine Unlearning for AI Safety}},
  journal        = {Transactions on Machine Learning Research},
  year           = {2025},
  eprint         = {2409.18025},
  archivePrefix  = {arXiv},
  url            = {https://arxiv.org/abs/2409.18025}
}

@article{che2025model,
  author         = {Zora Che and Stephen Casper and Robert Kirk and Anirudh Satheesh and Stewart Slocum and Lev E. McKinney and Rohit Gandikota and Aidan Ewart and Domenic Rosati and Zichu Wu and Zikui Cai and Bilal Chughtai and Yarin Gal and Furong Huang and Dylan Hadfield-Menell},
  title          = {{Model Tampering Attacks Enable More Rigorous Evaluations of LLM Capabilities}},
  journal        = {Transactions on Machine Learning Research},
  year           = {2025},
  eprint         = {2502.05209},
  archivePrefix  = {arXiv},
  url            = {https://arxiv.org/abs/2502.05209}
}

@inproceedings{qi2025durability,
  author         = {Xiangyu Qi and Boyi Wei and Nicholas Carlini and Yangsibo Huang and Tinghao Xie and Luxi He and Matthew Jagielski and Milad Nasr and Prateek Mittal and Peter Henderson},
  title          = {{On Evaluating the Durability of Safeguards for Open-Weight LLMs}},
  booktitle      = {Proceedings of the International Conference on Learning Representations (ICLR)},
  year           = {2025},
  eprint         = {2412.07097},
  archivePrefix  = {arXiv},
  url            = {https://arxiv.org/abs/2412.07097}
}

@inproceedings{ji2025resist,
  author         = {Jiaming Ji and Kaile Wang and Tianyi Qiu and Boyuan Chen and Jiayi Zhou and Changye Li and Hantao Lou and Juntao Dai and Yunhuai Liu and Yaodong Yang},
  title          = {{Language Models Resist Alignment: Evidence From Data Compression}},
  booktitle      = {Proceedings of the Annual Meeting of the Association for Computational Linguistics (ACL)},
  year           = {2025},
  eprint         = {2406.06144},
  archivePrefix  = {arXiv},
  url            = {https://arxiv.org/abs/2406.06144}
}

@inproceedings{siddiqui2025dormant,
  author         = {Shoaib Ahmed Siddiqui and Adrian Weller and David Krueger and Gintare Karolina Dziugaite and Michael Curtis Mozer and Eleni Triantafillou},
  title          = {{From Dormant to Deleted: Tamper-Resistant Unlearning Through Weight-Space Regularization}},
  booktitle      = {Advances in Neural Information Processing Systems (NeurIPS)},
  year           = {2025},
  eprint         = {2505.22310},
  archivePrefix  = {arXiv},
  url            = {https://arxiv.org/abs/2505.22310}
}

@inproceedings{ha2026blind,
  author         = {SeungBum Ha and Saerom Park and Sung Whan Yoon},
  title          = {{Unlearning's Blind Spots: Over-Unlearning and Prototypical Relearning Attack}},
  booktitle      = {Proceedings of the International Conference on Machine Learning (ICML)},
  year           = {2026},
  eprint         = {2506.01318},
  archivePrefix  = {arXiv},
  url            = {https://arxiv.org/abs/2506.01318}
}

@article{dehghani2026sourcefree,
  author         = {Zahra Dehghani and Pablo Piantanida and Mohammadhadi Shateri},
  title          = {{Source-Free Class Relearning: Diagnosing Forgetting in Class Unlearning}},
  journal        = {arXiv preprint arXiv:2609.02018},
  year           = {2026},
  eprint         = {2609.02018},
  archivePrefix  = {arXiv},
  url            = {https://arxiv.org/abs/2609.02018}
}

@article{suriyakumar2025unstable,
  author         = {Vinith M. Suriyakumar and Rohan Alur and Ayush Sekhari and Manish Raghavan and Ashia C. Wilson},
  title          = {{Unstable Unlearning: The Hidden Risk of Concept Resurgence in Diffusion Models}},
  journal        = {Transactions on Machine Learning Research},
  year           = {2026},
  eprint         = {2410.08074},
  archivePrefix  = {arXiv},
  url            = {https://arxiv.org/abs/2410.08074}
}

@inproceedings{gao2025meta,
  author         = {Hongcheng Gao and Tianyu Pang and Chao Du and Taihang Hu and Zhijie Deng and Min Lin},
  title          = {{Meta-Unlearning on Diffusion Models: Preventing Relearning Unlearned Concepts}},
  booktitle      = {Proceedings of the IEEE/CVF International Conference on Computer Vision (ICCV)},
  year           = {2025},
  eprint         = {2410.12777},
  archivePrefix  = {arXiv},
  url            = {https://arxiv.org/abs/2410.12777}
}

@article{tran2026efficient,
  author         = {Khoa Tran and Simon S. Woo},
  title          = {{Efficient Unlearning through Maximizing Relearning Convergence Delay}},
  journal        = {arXiv preprint arXiv:2604.09391},
  year           = {2026},
  eprint         = {2604.09391},
  archivePrefix  = {arXiv},
  url            = {https://arxiv.org/abs/2604.09391}
}

@inproceedings{fan2025sam,
  author         = {Chongyu Fan and Jinghan Jia and Yihua Zhang and Anil Ramakrishna and Mingyi Hong and Sijia Liu},
  title          = {{Towards LLM Unlearning Resilient to Relearning Attacks: A Sharpness-Aware Minimization Perspective and Beyond}},
  booktitle      = {Proceedings of the International Conference on Machine Learning (ICML)},
  year           = {2025},
  eprint         = {2502.05374},
  archivePrefix  = {arXiv},
  url            = {https://arxiv.org/abs/2502.05374}
}

@inproceedings{wang2025invariance,
  author         = {Changsheng Wang and Yihua Zhang and Jinghan Jia and Parikshit Ram and Dennis Wei and Yuguang Yao and Soumyadeep Pal and Nathalie Baracaldo and Sijia Liu},
  title          = {{Invariance Makes LLM Unlearning Resilient Even to Unanticipated Downstream Fine-Tuning}},
  booktitle      = {Proceedings of the International Conference on Machine Learning (ICML)},
  year           = {2025},
  eprint         = {2506.01339},
  archivePrefix  = {arXiv},
  url            = {https://arxiv.org/abs/2506.01339}
}

@article{yin2026margin,
  author         = {Xiangyu Yin and Jiaxu Liu and Zhen Chen and Chih-Hong Cheng},
  title          = {{Crossing the Margin Cliff: Toward Relearn-Robust LLM Unlearning via Margin Calibration}},
  journal        = {arXiv preprint arXiv:2607.27836},
  year           = {2026},
  eprint         = {2607.27836},
  archivePrefix  = {arXiv},
  url            = {https://arxiv.org/abs/2607.27836}
}

@inproceedings{tamirisa2025tar,
  author         = {Rishub Tamirisa and Bhrugu Bharathi and Long Phan and Andy Zhou and Alice Gatti and Tarun Suresh and Maxwell Lin and Justin Wang and Rowan Wang and Ron Arel and Andy Zou and Dawn Song and Bo Li and Dan Hendrycks and Mantas Mazeika},
  title          = {{Tamper-Resistant Safeguards for Open-Weight LLMs}},
  booktitle      = {Proceedings of the International Conference on Learning Representations (ICLR)},
  year           = {2025},
  eprint         = {2408.00761},
  archivePrefix  = {arXiv},
  url            = {https://arxiv.org/abs/2408.00761}
}

@article{sondej2025mudman,
  author         = {Filip Sondej and Yushi Yang and Miko{\l}aj Kniejski and Marcel Windys},
  title          = {{Robust LLM Unlearning with MUDMAN: Meta-Unlearning with Disruption Masking And Normalization}},
  journal        = {arXiv preprint arXiv:2506.12484},
  year           = {2025},
  eprint         = {2506.12484},
  archivePrefix  = {arXiv},
  url            = {https://arxiv.org/abs/2506.12484}
}

@inproceedings{lee2025undo,
  author         = {Bruce W. Lee and Addie Foote and Alex Infanger and Leni Shor and Harish Kamath and Jacob Goldman-Wetzler and Bryce Woodworth and Alex Cloud and Alexander Matt Turner},
  title          = {{Distillation Robustifies Unlearning}},
  booktitle      = {Advances in Neural Information Processing Systems (NeurIPS)},
  year           = {2025},
  eprint         = {2506.06278},
  archivePrefix  = {arXiv},
  url            = {https://arxiv.org/abs/2506.06278}
}

@article{rosati2026limits,
  author         = {Domenic Rosati and Xijie Zeng and Hong Huang and Sebastian Dionicio and Subhabrata Majumdar and Frank Rudzicz and Hassan Sajjad},
  title          = {{Limits of Convergence-Rate Control for Open-Weight Safety}},
  journal        = {arXiv preprint arXiv:2602.18868},
  year           = {2026},
  eprint         = {2602.18868},
  archivePrefix  = {arXiv},
  url            = {https://arxiv.org/abs/2602.18868}
}

@inproceedings{chen2026frag,
  author         = {Yi Chen and Hanna Hsieh and Shuhong Liu and Chuanbo Hua and Zihan Ma and Kun Wang and Joo-Young Kim},
  title          = {{Distance Is Not Enough: Forget-Retain Alignment Gap Predicts LLM Relearning Robustness}},
  booktitle      = {Proceedings of the Conference on Empirical Methods in Natural Language Processing (EMNLP)},
  year           = {2026},
  eprint         = {2608.25429},
  archivePrefix  = {arXiv},
  url            = {https://arxiv.org/abs/2608.25429}
}

@article{sondej2026repselect,
  author         = {Filip Sondej and Yushi Yang and Adam Mahdi},
  title          = {{RepSelect: Robust LLM Unlearning via Representation Selectivity}},
  journal        = {arXiv preprint arXiv:2606.17168},
  year           = {2026},
  eprint         = {2606.17168},
  archivePrefix  = {arXiv},
  url            = {https://arxiv.org/abs/2606.17168}
}

@article{xiao2026mcu,
  author         = {Zeguan Xiao and Xuanzhe Xu and Yun Chen and Yong Wang and Jian Yang and Yanqing Hu and Guanhua Chen},
  title          = {{Robust LLM Unlearning Against Relearning Attacks: The Minor Components in Representations Matter}},
  journal        = {arXiv preprint arXiv:2605.11685},
  year           = {2026},
  eprint         = {2605.11685},
  archivePrefix  = {arXiv},
  url            = {https://arxiv.org/abs/2605.11685}
}

@article{sondej2025cir,
  author         = {Filip Sondej and Yushi Yang},
  title          = {{Collapse of Irrelevant Representations (CIR) Ensures Robust and Non-Disruptive LLM Unlearning}},
  journal        = {arXiv preprint arXiv:2509.11816},
  year           = {2025},
  eprint         = {2509.11816},
  archivePrefix  = {arXiv},
  url            = {https://arxiv.org/abs/2509.11816}
}

@inproceedings{chen2024unsc,
  author         = {Huiqiang Chen and Tianqing Zhu and Xin Yu and Wanlei Zhou},
  title          = {{Machine Unlearning via Null Space Calibration}},
  booktitle      = {Proceedings of the International Joint Conference on Artificial Intelligence (IJCAI)},
  year           = {2024},
  eprint         = {2404.13588},
  archivePrefix  = {arXiv},
  url            = {https://arxiv.org/abs/2404.13588}
}

@inproceedings{hoang2024pgu,
  author         = {Tuan Hoang and Santu Rana and Sunil Gupta and Svetha Venkatesh},
  title          = {{Learn to Unlearn for Deep Neural Networks: Minimizing Unlearning Interference with Gradient Projection}},
  booktitle      = {Proceedings of the IEEE/CVF Winter Conference on Applications of Computer Vision (WACV)},
  year           = {2024},
  eprint         = {2312.04095},
  archivePrefix  = {arXiv},
  url            = {https://arxiv.org/abs/2312.04095}
}

@inproceedings{fang2025alphaedit,
  author         = {Junfeng Fang and Houcheng Jiang and Kun Wang and Yunshan Ma and Shi Jie and Xiang Wang and Xiangnan He and Tat-Seng Chua},
  title          = {{AlphaEdit: Null-Space Constrained Knowledge Editing for Language Models}},
  booktitle      = {Proceedings of the International Conference on Learning Representations (ICLR)},
  year           = {2025},
  eprint         = {2410.02355},
  archivePrefix  = {arXiv},
  url            = {https://arxiv.org/abs/2410.02355}
}

@article{zhou2025geometric,
  author         = {Duo Zhou and Yuji Zhang and Tianxin Wei and Ruizhong Qiu and Ke Yang and Xiao Lin and Cheng Qian and Jingrui He and Hanghang Tong and Chengxiang Zhai and Heng Ji and Huan Zhang},
  title          = {{Geometric-Disentanglement Unlearning}},
  journal        = {arXiv preprint arXiv:2511.17100},
  year           = {2025},
  eprint         = {2511.17100},
  archivePrefix  = {arXiv},
  url            = {https://arxiv.org/abs/2511.17100}
}

@inproceedings{saha2021gradient,
  author         = {Gobinda Saha and Isha Garg and Kaushik Roy},
  title          = {{Gradient Projection Memory for Continual Learning}},
  booktitle      = {Proceedings of the International Conference on Learning Representations (ICLR)},
  year           = {2021},
  eprint         = {2103.09762},
  archivePrefix  = {arXiv},
  url            = {https://arxiv.org/abs/2103.09762}
}

@inproceedings{wang2021adamnscl,
  author         = {Shipeng Wang and Xiaorong Li and Jian Sun and Zongben Xu},
  title          = {{Training Networks in Null Space of Feature Covariance for Continual Learning}},
  booktitle      = {Proceedings of the IEEE/CVF Conference on Computer Vision and Pattern Recognition (CVPR)},
  year           = {2021},
  url            = {https://openaccess.thecvf.com/content/CVPR2021/html/Wang_Training_Networks_in_Null_Space_of_Feature_Covariance_for_Continual_CVPR_2021_paper.html}
}

@inproceedings{cong2023projector,
  author         = {Weilin Cong and Mehrdad Mahdavi},
  title          = {{Efficiently Forgetting What You Have Learned in Graph Representation Learning via Projection}},
  booktitle      = {Proceedings of the International Conference on Artificial Intelligence and Statistics (AISTATS)},
  year           = {2023},
  volume         = {206},
  pages          = {6674--6703},
  url            = {https://proceedings.mlr.press/v206/cong23a.html}
}

@article{yin2025rangenull,
  author         = {Hang Yin and Zipeng Liu and Xiaoyong Peng and Liyao Xiang},
  title          = {{Graph Unlearning via Embedding Reconstruction -- A Range-Null Space Decomposition Approach}},
  journal        = {arXiv preprint arXiv:2508.02044},
  year           = {2025},
  eprint         = {2508.02044},
  archivePrefix  = {arXiv},
  url            = {https://arxiv.org/abs/2508.02044}
}

@inproceedings{golatkar2020forgetting,
  author         = {Aditya Golatkar and Alessandro Achille and Stefano Soatto},
  title          = {{Forgetting Outside the Box: Scrubbing Deep Networks of Information Accessible from Input-Output Observations}},
  booktitle      = {Proceedings of the European Conference on Computer Vision (ECCV)},
  year           = {2020},
  eprint         = {2003.02960},
  archivePrefix  = {arXiv},
  url            = {https://arxiv.org/abs/2003.02960}
}

@inproceedings{golatkar2020eternal,
  author         = {Aditya Golatkar and Alessandro Achille and Stefano Soatto},
  title          = {{Eternal Sunshine of the Spotless Net: Selective Forgetting in Deep Networks}},
  booktitle      = {Proceedings of the IEEE/CVF Conference on Computer Vision and Pattern Recognition (CVPR)},
  year           = {2020},
  eprint         = {1911.04933},
  archivePrefix  = {arXiv},
  url            = {https://arxiv.org/abs/1911.04933}
}

@inproceedings{doan2021theoretical,
  author         = {Thang Doan and Mehdi Bennani and Bogdan Mazoure and Guillaume Rabusseau and Pierre Alquier},
  title          = {{A Theoretical Analysis of Catastrophic Forgetting through the NTK Overlap Matrix}},
  booktitle      = {Proceedings of the International Conference on Artificial Intelligence and Statistics (AISTATS)},
  year           = {2021},
  eprint         = {2010.04003},
  archivePrefix  = {arXiv},
  url            = {https://arxiv.org/abs/2010.04003}
}

@article{bennani2020generalisation,
  author         = {Mehdi Abbana Bennani and Thang Doan and Masashi Sugiyama},
  title          = {{Generalisation Guarantees for Continual Learning with Orthogonal Gradient Descent}},
  journal        = {arXiv preprint arXiv:2006.11942},
  year           = {2020},
  eprint         = {2006.11942},
  archivePrefix  = {arXiv},
  url            = {https://arxiv.org/abs/2006.11942}
}

@inproceedings{hidekel2026catastrophic,
  author         = {Ido Nitzan Hidekel and Dan Raviv},
  title          = {{Catastrophic Forgetting is Low-Rank: A Function-Space Theory for Continual Adaptation}},
  booktitle      = {ICML 2026 Workshop on Continual Adaptation at Scale: Towards Sustainable AI},
  year           = {2026},
  eprint         = {2606.18024},
  archivePrefix  = {arXiv},
  url            = {https://arxiv.org/abs/2606.18024}
}

@inproceedings{sekhari2021remember,
  author         = {Ayush Sekhari and Jayadev Acharya and Gautam Kamath and Ananda Theertha Suresh},
  title          = {{Remember What You Want to Forget: Algorithms for Machine Unlearning}},
  booktitle      = {Advances in Neural Information Processing Systems (NeurIPS)},
  year           = {2021},
  eprint         = {2103.03279},
  archivePrefix  = {arXiv},
  url            = {https://arxiv.org/abs/2103.03279}
}

@inproceedings{koh2017understanding,
  author         = {Pang Wei Koh and Percy Liang},
  title          = {{Understanding Black-box Predictions via Influence Functions}},
  booktitle      = {Proceedings of the International Conference on Machine Learning (ICML)},
  year           = {2017},
  eprint         = {1703.04730},
  archivePrefix  = {arXiv},
  url            = {https://arxiv.org/abs/1703.04730}
}

@inproceedings{guo2020certified,
  author         = {Chuan Guo and Tom Goldstein and Awni Hannun and Laurens van der Maaten},
  title          = {{Certified Data Removal from Machine Learning Models}},
  booktitle      = {Proceedings of the International Conference on Machine Learning (ICML)},
  year           = {2020},
  eprint         = {1911.03030},
  archivePrefix  = {arXiv},
  url            = {https://arxiv.org/abs/1911.03030}
}

@inproceedings{izzo2021approximate,
  author         = {Zachary Izzo and Mary Anne Smart and Kamalika Chaudhuri and James Zou},
  title          = {{Approximate Data Deletion from Machine Learning Models}},
  booktitle      = {Proceedings of the International Conference on Artificial Intelligence and Statistics (AISTATS)},
  year           = {2021},
  eprint         = {2002.10077},
  archivePrefix  = {arXiv},
  url            = {https://arxiv.org/abs/2002.10077}
}

@inproceedings{malladi2023kernel,
  author         = {Sadhika Malladi and Alexander Wettig and Dingli Yu and Danqi Chen and Sanjeev Arora},
  title          = {{A Kernel-Based View of Language Model Fine-Tuning}},
  booktitle      = {Proceedings of the International Conference on Machine Learning (ICML)},
  year           = {2023},
  eprint         = {2210.05643},
  archivePrefix  = {arXiv},
  url            = {https://arxiv.org/abs/2210.05643}
}

@article{chang2026resist,
  author         = {Wei-Kai Chang and Rajiv Khanna},
  title          = {{Why Some Models Resist Unlearning: A Linear Stability Perspective}},
  journal        = {arXiv preprint arXiv:2602.02986},
  year           = {2026},
  eprint         = {2602.02986},
  archivePrefix  = {arXiv},
  url            = {https://arxiv.org/abs/2602.02986}
}

@inproceedings{jacot2018neural,
  author         = {Arthur Jacot and Franck Gabriel and Cl{\'e}ment Hongler},
  title          = {{Neural Tangent Kernel: Convergence and Generalization in Neural Networks}},
  booktitle      = {Advances in Neural Information Processing Systems (NeurIPS)},
  year           = {2018},
  eprint         = {1806.07572},
  archivePrefix  = {arXiv},
  url            = {https://arxiv.org/abs/1806.07572}
}

@inproceedings{liu2026graphdpo,
  author         = {Jiajun Liu and Wenjun Ke and Peng Wang and Yao He and Ziyu Shang and Guozheng Li and Zijie Xu and Ke Ji},
  title          = {{Unlearning of Knowledge Graph Embedding via Preference Optimization}},
  booktitle      = {Proceedings of the ACM Web Conference (WWW)},
  year           = {2026},
  eprint         = {2507.20566},
  archivePrefix  = {arXiv},
  url            = {https://arxiv.org/abs/2507.20566}
}

@inproceedings{xiao2025zerofisher,
  author         = {Yang Xiao and Ruimeng Ye and Bohan Liu and Xiaolong Ma and Bo Hui},
  title          = {{Efficient Knowledge Graph Unlearning with Zeroth-order Information}},
  booktitle      = {Proceedings of the ACM International Conference on Information and Knowledge Management (CIKM)},
  year           = {2025},
  eprint         = {2508.14013},
  archivePrefix  = {arXiv},
  url            = {https://arxiv.org/abs/2508.14013}
}

@inproceedings{xiao2025schema,
  author         = {Yang Xiao and Ruimeng Ye and Bo Hui},
  title          = {{Knowledge Graph Unlearning with Schema}},
  booktitle      = {Proceedings of the 31st International Conference on Computational Linguistics (COLING)},
  year           = {2025},
  url            = {https://aclanthology.org/2025.coling-main.238/}
}

@article{xu2024metaeu,
  author         = {Naixing Xu and Qian Li and Zhaochuan Li and Xu Wang and Bingchen Liu and Jabulani Brown Mpofu and Jingchen Li and Xin Li},
  title          = {{Learn to Unlearn: Meta-Learning-Based Knowledge Graph Embedding Unlearning}},
  journal        = {Knowledge and Information Systems},
  volume         = {67},
  number         = {8},
  pages          = {7101--7122},
  year           = {2025},
  url            = {https://doi.org/10.1007/s10115-025-02407-4}
}

@inproceedings{cheng2023gnndelete,
  author         = {Jiali Cheng and George Dasoulas and Huan He and Chirag Agarwal and Marinka Zitnik},
  title          = {{GNNDelete: A General Strategy for Unlearning in Graph Neural Networks}},
  booktitle      = {Proceedings of the International Conference on Learning Representations (ICLR)},
  year           = {2023},
  eprint         = {2302.13406},
  archivePrefix  = {arXiv},
  url            = {https://arxiv.org/abs/2302.13406}
}

@inproceedings{wu2023gif,
  author         = {Jiancan Wu and Yi Yang and Yuchun Qian and Yongduo Sui and Xiang Wang and Xiangnan He},
  title          = {{GIF: A General Graph Unlearning Strategy via Influence Function}},
  booktitle      = {Proceedings of the ACM Web Conference (WWW)},
  year           = {2023},
  eprint         = {2304.02835},
  archivePrefix  = {arXiv},
  url            = {https://arxiv.org/abs/2304.02835}
}

@inproceedings{fan2025opengu,
  author         = {Bowen Fan and Yuming Ai and Xunkai Li and Zhilin Guo and Zhu Lei and Guang Zeng and Rong-Hua Li and Guoren Wang},
  title          = {{OpenGU: A Comprehensive Benchmark for Graph Unlearning}},
  booktitle      = {Advances in Neural Information Processing Systems (NeurIPS) Datasets and Benchmarks Track},
  year           = {2025},
  eprint         = {2501.02728},
  archivePrefix  = {arXiv},
  url            = {https://arxiv.org/abs/2501.02728}
}

@inproceedings{jang2023knowledge,
  author         = {Joel Jang and Dongkeun Yoon and Sohee Yang and Sungmin Cha and Moontae Lee and Lajanugen Logeswaran and Minjoon Seo},
  title          = {{Knowledge Unlearning for Mitigating Privacy Risks in Language Models}},
  booktitle      = {Proceedings of the 61st Annual Meeting of the Association for Computational Linguistics (ACL)},
  year           = {2023},
  eprint         = {2210.01504},
  archivePrefix  = {arXiv},
  url            = {https://arxiv.org/abs/2210.01504}
}

@inproceedings{zhang2024npo,
  author         = {Ruiqi Zhang and Licong Lin and Yu Bai and Song Mei},
  title          = {{Negative Preference Optimization: From Catastrophic Collapse to Effective Unlearning}},
  booktitle      = {Conference on Language Modeling (COLM)},
  year           = {2024},
  eprint         = {2404.05868},
  archivePrefix  = {arXiv},
  url            = {https://arxiv.org/abs/2404.05868}
}

@inproceedings{liu2024incde,
  author         = {Jiajun Liu and Wenjun Ke and Peng Wang and Ziyu Shang and Jinhua Gao and Guozheng Li and Ke Ji and Yanhe Liu},
  title          = {{Towards Continual Knowledge Graph Embedding via Incremental Distillation}},
  booktitle      = {Proceedings of the AAAI Conference on Artificial Intelligence},
  year           = {2024},
  eprint         = {2405.04453},
  archivePrefix  = {arXiv},
  url            = {https://arxiv.org/abs/2405.04453}
}

@inproceedings{bordes2013translating,
  author         = {Antoine Bordes and Nicolas Usunier and Alberto Garc{\'\i}a-Dur{\'a}n and Jason Weston and Oksana Yakhnenko},
  title          = {{Translating Embeddings for Modeling Multi-relational Data}},
  booktitle      = {Advances in Neural Information Processing Systems (NeurIPS)},
  year           = {2013},
  url            = {https://papers.nips.cc/paper/5071-translating-embeddings-for-modeling-multi-relational-data}
}

@inproceedings{sun2019rotate,
  author         = {Zhiqing Sun and Zhi-Hong Deng and Jian-Yun Nie and Jian Tang},
  title          = {{RotatE: Knowledge Graph Embedding by Relational Rotation in Complex Space}},
  booktitle      = {Proceedings of the International Conference on Learning Representations (ICLR)},
  year           = {2019},
  eprint         = {1902.10197},
  archivePrefix  = {arXiv},
  url            = {https://arxiv.org/abs/1902.10197}
}

@inproceedings{toutanova2015observed,
  author         = {Kristina Toutanova and Danqi Chen},
  title          = {{Observed versus Latent Features for Knowledge Base and Text Inference}},
  booktitle      = {Proceedings of the 3rd Workshop on Continuous Vector Space Models and their Compositionality (CVSC)},
  year           = {2015},
  pages          = {57--66},
  url            = {https://aclanthology.org/W15-4007/}
}

@inproceedings{kingma2015adam,
  author         = {Diederik P. Kingma and Jimmy Ba},
  title          = {{Adam: A Method for Stochastic Optimization}},
  booktitle      = {Proceedings of the International Conference on Learning Representations (ICLR)},
  year           = {2015},
  eprint         = {1412.6980},
  archivePrefix  = {arXiv},
  url            = {https://arxiv.org/abs/1412.6980}
}

@inproceedings{heo2021adamp,
  author         = {Byeongho Heo and Sanghyuk Chun and Seong Joon Oh and Dongyoon Han and Sangdoo Yun and Gyuwan Kim and Youngjung Uh and Jung-Woo Ha},
  title          = {{AdamP: Slowing Down the Slowdown for Momentum Optimizers on Scale-invariant Weights}},
  booktitle      = {Proceedings of the International Conference on Learning Representations (ICLR)},
  year           = {2021},
  eprint         = {2006.08217},
  archivePrefix  = {arXiv},
  url            = {https://arxiv.org/abs/2006.08217}
}

@inproceedings{arora2019theoretical,
  author         = {Sanjeev Arora and Zhiyuan Li and Kaifeng Lyu},
  title          = {{Theoretical Analysis of Auto Rate-Tuning by Batch Normalization}},
  booktitle      = {Proceedings of the International Conference on Learning Representations (ICLR)},
  year           = {2019},
  eprint         = {1812.03981},
  archivePrefix  = {arXiv},
  url            = {https://arxiv.org/abs/1812.03981}
}

@article{hestenes1952methods,
  author         = {Magnus R. Hestenes and Eduard Stiefel},
  title          = {{Methods of Conjugate Gradients for Solving Linear Systems}},
  journal        = {Journal of Research of the National Bureau of Standards},
  year           = {1952},
  volume         = {49},
  number         = {6},
  pages          = {409--436},
  url            = {https://nvlpubs.nist.gov/nistpubs/jres/049/6/V49.N06.A08.pdf}
}

@inproceedings{bourtoule2021machine,
  author         = {Lucas Bourtoule and Varun Chandrasekaran and Christopher A. Choquette-Choo and Hengrui Jia and Adelin Travers and Baiwu Zhang and David Lie and Nicolas Papernot},
  title          = {{Machine Unlearning}},
  booktitle      = {Proceedings of the IEEE Symposium on Security and Privacy (S\&P)},
  year           = {2021},
  eprint         = {1912.03817},
  archivePrefix  = {arXiv},
  url            = {https://arxiv.org/abs/1912.03817}
}

@inproceedings{cao2015towards,
  author         = {Yinzhi Cao and Junfeng Yang},
  title          = {{Towards Making Systems Forget with Machine Unlearning}},
  booktitle      = {Proceedings of the IEEE Symposium on Security and Privacy (S\&P)},
  year           = {2015},
  pages          = {463--480},
  url            = {https://doi.org/10.1109/SP.2015.35}
}

@inproceedings{wang2014transh,
  author         = {Zhen Wang and Jianwen Zhang and Jianlin Feng and Zheng Chen},
  title          = {{Knowledge Graph Embedding by Translating on Hyperplanes}},
  booktitle      = {Proceedings of the AAAI Conference on Artificial Intelligence (AAAI)},
  year           = {2014},
  pages          = {1112--1119}
}

@article{yao2007early,
  author         = {Yuan Yao and Lorenzo Rosasco and Andrea Caponnetto},
  title          = {{On Early Stopping in Gradient Descent Learning}},
  journal        = {Constructive Approximation},
  volume         = {26},
  number         = {2},
  pages          = {289--315},
  year           = {2007}
}

@inproceedings{ali2019continuous,
  author         = {Alnur Ali and J. Zico Kolter and Ryan J. Tibshirani},
  title          = {{A Continuous-Time View of Early Stopping for Least Squares Regression}},
  booktitle      = {Proceedings of the International Conference on Artificial Intelligence and Statistics (AISTATS)},
  year           = {2019},
  pages          = {1370--1378}
}

@inproceedings{li2025resalign,
  author        = {Boheng Li and Renjie Gu and Junjie Wang and Leyi Qi and Yiming Li and Run Wang and Zhan Qin and Tianwei Zhang},
  title         = {{Towards Resilient Safety-driven Unlearning for Diffusion Models against Downstream Fine-tuning}},
  booktitle     = {Advances in Neural Information Processing Systems (NeurIPS)},
  year          = {2025},
  eprint        = {2507.16302},
  archivePrefix = {arXiv},
  url           = {https://arxiv.org/abs/2507.16302}
}

@inproceedings{rajeswaran2019imaml,
  author        = {Aravind Rajeswaran and Chelsea Finn and Sham M. Kakade and Sergey Levine},
  title         = {{Meta-Learning with Implicit Gradients}},
  booktitle     = {Advances in Neural Information Processing Systems (NeurIPS)},
  year          = {2019},
  eprint        = {1909.04630},
  archivePrefix = {arXiv},
  url           = {https://arxiv.org/abs/1909.04630}
}
\bibliographystyle{plainnat}

\newpage
\appendix
\section{Analysis and proofs}
\label{app:proofs}
\renewcommand{\thelemma}{A.\arabic{lemma}}\setcounter{lemma}{0}\def\theHlemma{A.\arabic{lemma}}
\renewcommand{\theremark}{A.\arabic{remark}}\setcounter{remark}{0}\def\theHremark{A.\arabic{remark}}
\renewcommand{\theproposition}{A.\arabic{proposition}}\setcounter{proposition}{0}\def\theHproposition{A.\arabic{proposition}}
\renewcommand{\thetheorem}{A.\arabic{theorem}}\setcounter{theorem}{0}\def\theHtheorem{A.\arabic{theorem}}
\renewcommand{\thecorollary}{A.\arabic{corollary}}\setcounter{corollary}{0}\def\theHcorollary{A.\arabic{corollary}}

Section~\ref{sec:theory} summarizes what the update does to an edit. Appendix~\ref{app:theory} states that analysis in full, in Proposition~\ref{prop:release}, Corollaries~\ref{cor:composition}, \ref{cor:thresholds} and~\ref{cor:soft}, Theorem~\ref{thm:operator} and Remark~\ref{rem:cone}. The remaining subsections prove these results and Lemma~\ref{lem:response}, and they state the supporting results that Section~\ref{sec:method} cites.

\paragraph{Conventions.}
Throughout, $v$ is an involved entity, $N$ a fixed draw, $P_v(N)=\{i\in N:v\in i\}$ the set of drawn edges incident to $v$, and $n_v=|P_v(N)|$. A drawn edge is a triple of the draw $N$, and it contributes one row to every Jacobian below. The slot feature of edge $i$ at $v$ is $\feat{i}^{(v)}=\partial\dist{i}/\partial\ve_v\in\R^m$, and $\Phimat_v\in\R^{n_v\times m}$ stacks $\feat{i}^{(v)\top}$ over $i\in P_v(N)$. For the two backbones with normalized rows the slot feature has the form
\begin{equation}\label{eq:slotfeat}
\feat{i}^{(v)}=\frac{(\Id-\bar\ve_v\bar\ve_v^{\top})\,\vg_i^{(v)}}{\|\ve_v\|_2},
\end{equation}
where $\vg_i^{(v)}$ is the gradient of the score with respect to the normalized row $\bar\ve_v$. For TransE, $\vg_i^{(v)}=\pm\sign(\bar\ve_h+\bar\vw_r-\bar\ve_t)$, with the plus sign at the head and the minus sign at the tail, defined wherever no coordinate of the translation error $\bar\ve_h+\bar\vw_r-\bar\ve_t$ vanishes. For TransH, $\vg_i^{(v)}=\pm\mP_r\sign(\mP_r\bar\ve_h+\vd_r-\mP_r\bar\ve_t)$, the sign vector of the TransH translation error multiplied by the symmetric projector $\mP_r$. In both cases every slot feature at $v$ lies in the tangent space $\bar\ve_v^{\perp}$ of the unit sphere at $\bar\ve_v$, a subspace of dimension $m-1$. RotatE does not normalize its rows and its features carry no such constraint.

Let $\Phimat_v=\sum_{k\le\varrho}\sqrt{\lambda_k}\,\vq_k\vu_k^{\top}$ be a singular value decomposition, with $\varrho=\operatorname{rank}\Phimat_v$, $\lambda_1\ge\dots\ge\lambda_\varrho>0$ the nonzero eigenvalues of $\Kmat_v=\Phimat_v\Phimat_v^{\top}$, $\vq_k\in\R^{n_v}$ the corresponding orthonormal eigenvectors, and $\vu_k=\Phimat_v^{\top}\vq_k/\sqrt{\lambda_k}\in\R^m$ an orthonormal basis of $\operatorname{range}(\Phimat_v^{\top})$. With $\ridge_v>0$,
\begin{equation}\label{eq:spectral}
\begin{aligned}
\mA_v&=\Phimat_v^{\top}(\Kmat_v+\ridge_v\Id)^{-1}=\sum_{k\le\varrho}\frac{\sqrt{\lambda_k}}{\lambda_k+\ridge_v}\,\vu_k\vq_k^{\top},\\
\mP^{\perp}_{N,v}&=\Id-\mA_v\Phimat_v=\sum_{k\le\varrho}\frac{\ridge_v}{\lambda_k+\ridge_v}\,\vu_k\vu_k^{\top}+\Pi^{\perp}_{N,v},
\end{aligned}
\end{equation}
where $\Pi_{N,v}=\sum_{k\le\varrho}\vu_k\vu_k^{\top}$ is the orthogonal projector onto the span of the drawn features at $v$ and $\Pi^{\perp}_{N,v}=\Id-\Pi_{N,v}$. The first identity holds because $(\Kmat_v+\ridge_v\Id)^{-1}$ has the eigenvectors $\vq_k$ with eigenvalues $1/(\lambda_k+\ridge_v)$ on $\operatorname{range}(\Phimat_v)$ and $\Phimat_v^{\top}$ annihilates its orthogonal complement. The second follows by multiplying out. Three facts are read off~\eqref{eq:spectral} and used below. The operator $\mP^{\perp}_{N,v}$ is symmetric with spectrum $\{\ridge_v/(\lambda_k+\ridge_v)\}_{k\le\varrho}$, together with $1$ when $\varrho<m$. It is therefore positive definite, and it is idempotent only when $\varrho=0$, where it is the identity. It converges to $\Pi^{\perp}_{N,v}$ as $\ridge_v\to0$. Its companion satisfies $\|\mA_v\|_{\mathrm{op}}=\max_k\sqrt{\lambda_k}/(\lambda_k+\ridge_v)\le1/(2\sqrt{\ridge_v})$, because $\sqrt\lambda/(\lambda+\ridge)$ is maximized over $\lambda\ge0$ at $\lambda=\ridge$. When no edge is drawn at $v$, $\Phimat_v$ is empty, $\Delta\ve_v=0$ and $\mP^{\perp}_{N,v}=\Id$, and statements that involve $\lambda^{+}_{\min}$ are made for $\varrho\ge1$.

\subsection{What survives a linearized update}\label{app:theory}

This subsection gives the full statements behind Section~\ref{sec:theory}.
It shows what the update does to an edit, and what that implies for the objective of~\eqref{eq:objective} and for subspace methods.
In a linearized model the update multiplies the edit by a fixed matrix, the survival filter, and the anticipated loss is the plain forgetting loss on the filtered edit.
Subspace methods confine the edit to a hard threshold of the same filter.
Theorem~\ref{thm:operator} gives the local version for the nonlinear objective, and Corollary~\ref{cor:soft} averages it over draws.

\paragraph{The linearized model.}
Fix a draw $N$ and write $\delta=\theta-\orig$ for the edit of the involved rows.
The linearized model freezes the stacked features $\Phimat$ of the drawn edges and the feature $\feat{f}$ of every forgotten fact at the original model.
Every score is then linear in the edit, $\dist{f}(\orig+\delta)=\dist{f}(\orig)+\feat{f}^{\top}\delta$.
The residual is the signed residual $r(\theta)=\Phimat(\theta-\orig)$, and the response is the joint solve of~\eqref{eq:coupled}, $\resp(\theta)=-(\Phimat^{\top}\Phimat+\mSigma)^{-1}\Phimat^{\top}r(\theta)$.
The proposition below shows that the update returns the edit multiplied by a fixed matrix, and it gives the fraction of each direction that survives.

\begin{proposition}[The update filters the edit]\label{prop:survival}\label{prop:release}
In the linearized model, every edit $\delta$ satisfies
\begin{equation}\label{eq:survival}
\orig+\delta+\resp(\orig+\delta)=\orig+\mS\delta,
\qquad
\mS=(\Phimat^{\top}\Phimat+\mSigma)^{-1}\mSigma ,
\end{equation}
and $\theta+\delta+\resp(\theta+\delta)=\theta+\resp(\theta)+\mS\delta$ at every $\theta$. Moreover, $\mS\delta=\delta$ if and only if $\Phimat\delta=0$. For a uniform ridge $\mSigma=\ridge\Id$, with $\Kmat=\Phimat\Phimat^{\top}$,
\begin{equation*}
\mS=\Id-\Phimat^{\top}(\Kmat+\ridge\Id)^{-1}\Phimat=\sum_{k}\frac{\ridge}{\lambda_k+\ridge}\,\vu_k\vu_k^{\top}+\Pi^{\perp},
\end{equation*}
where $(\lambda_k,\vu_k)$ are the nonzero eigenpairs of $\Phimat^{\top}\Phimat$ and $\Pi^{\perp}$ is the orthogonal projector off the span of the update's features. The component of the edit along $\vu_k$ survives with the fraction $\ridge/(\lambda_k+\ridge)$, and the component off the span survives entirely. Let the update be $T$ steps of gradient descent at rate $\eta\le1/\lambda_{\max}(\Kmat)$ on the linearized squared residual $\tfrac12\|r\|_2^2$, started at the edited model. The surviving edit is then exactly $(\Id-\eta\Phimat^{\top}\Phimat)^{T}\delta$, and its fraction along $\vu_k$ satisfies $(1-\eta\lambda_k)^{T}\le\ridge/(\lambda_k+\ridge)$ at $\ridge=1/(\eta T)$. For an edit that moves the row $\ve_v$ alone, by $\delta_v$, the per-entity response $\Delta\ve_v=-\mA_vr_v$ of Section~\ref{sec:method:response} obeys the same identities with the per-entity filter $\mP^{\perp}_{N,v}=\Id-\Phimat_v^{\top}(\Phimat_v\Phimat_v^{\top}+\ridge_v\Id)^{-1}\Phimat_v$ in place of $\mS$ and $\delta_v$ in place of $\delta$. For $\Phimat_v\neq0$, every displacement $\delta_v\in\operatorname{range}(\Phimat_v^{\top})$ of the row $\ve_v$ satisfies $\|\mP^{\perp}_{N,v}\delta_v\|_2\le\ridge_v/(\lambda^{+}_{\min}+\ridge_v)\,\|\delta_v\|_2$, with $\lambda^{+}_{\min}$ the smallest nonzero eigenvalue of $\Phimat_v\Phimat_v^{\top}$.
\end{proposition}

An update of finite length restores the directions in which its features are large and leaves the rest of the edit in place.
The spectral form is the classical spectral filter of ridge and early-stopped least squares \citep{yao2007early,ali2019continuous} (Remark~\ref{rem:ridge}).
It is also the operator that governs interference between tasks in linearized continual learning \citep{doan2021theoretical,bennani2020generalisation}.
With the Gauss--Newton Hessian $\Phimat^{\top}\Phimat$ and the uniform ridge as proximal weight, it is also the Jacobian of a proximal model of adaptation with respect to its centre \citep{rajeswaran2019imaml}.
\citet{li2025resalign} use its full-Hessian form at a simulated fine-tune's end point to differentiate a forgetting loss, whereas \KAU{}'s response is the closed-form proximal point of the linearized squared residual.
The proposition records that filter for the edit.
What is new is its use in Corollaries~\ref{cor:composition} and~\ref{cor:thresholds}.
Because the signed residual vanishes at the original model, the anticipated loss is the plain loss on the filtered edit, and confinement to a subspace is its hard threshold.
The identity of~\eqref{eq:survival} and the identity of Corollary~\ref{cor:composition} hold for every positive diagonal $\mSigma$, which covers the per-entity ridge of the experiments.
The spectral form and Corollary~\ref{cor:thresholds} take the uniform ridge, under which $\mS$ is symmetric.
Appendix~\ref{app:survival} proves the proposition, the two corollaries that rest on it, and Remark~\ref{rem:cone}.

\paragraph{The objective acts on the surviving edit.}
The next corollary shows that anticipating the update amounts to applying the plain loss to the surviving edit.
Write $\mathcal{L}_{\mathrm{ant}}(\delta)$ for the loss of~\eqref{eq:objective} at the draw $N$, with the anticipated score in place of the composed score.

\begin{corollary}[The objective acts on the surviving edit]\label{cor:composition}
In the linearized model the anticipated score of a forgotten fact is $\dist{f}(\orig)+\feat{f}^{\top}\mS\delta$, so
\begin{equation*}
\mathcal{L}_{\mathrm{ant}}(\delta)=\mathcal{L}_{\mathrm{plain}}(\mS\delta),
\qquad
\mathcal{L}_{\mathrm{plain}}(\delta)=\frac{1}{|\Fset|}\sum_{f\in\Fset}\relu\big(t_f-\dist{f}(\orig)-\feat{f}^{\top}\delta\big)^2 ,
\end{equation*}
where $\mathcal{L}_{\mathrm{plain}}$ is the same loss on the pre-update score. With the composed score of~\eqref{eq:objective}, the loss of fact $f$ is $\max\{\ell_f(\delta),\ell_f(\mS\delta)\}$ with $\ell_f(\delta)=\relu(t_f-\dist{f}(\orig)-\feat{f}^{\top}\delta)^2$, since $\relu(t-\min\{a,b\})^2=\max\{\relu(t-a)^2,\relu(t-b)^2\}$. It equals $\ell_f(\mS\delta)$ whenever the update lowers the fact's score. The minimum of~\eqref{eq:objective} then selects the anticipated score, which is the case whenever $\feat{f}^{\top}(\Id-\mS)\delta\ge0$. The condition holds under a uniform ridge for every edit $\delta=c\,\feat{f}$ and every edit $\delta=c\,\mS\feat{f}$ with $c\ge0$, the second being the direction the anticipated loss itself produces. On both branches the gradient is taken through the anticipated score (Section~\ref{sec:method}), so the minimum changes a scalar weight and leaves the direction $\mS^{\top}\feat{f}$ unchanged.
\end{corollary}

In the linearized model the anticipated loss is therefore the plain forgetting loss evaluated on the part of the edit that survives the update.
It is the objective of \KAU{} under the signed residual, wherever the minimum selects the anticipated score.
For an edit that combines several facts, the condition of the corollary can fail for one of them.
The minimum then changes the weight of that fact and leaves its direction unchanged.
The identity relates the two objectives, and it holds for the joint response at every draw.
That the anticipated loss is the plain loss at the post-update point holds by construction.
The content of the identity is that, in the linearized model, the post-update edit is a fixed linear image $\mS\delta$ of the edit, with the explicit spectrum of Proposition~\ref{prop:release}.

The identity alone leaves the location of a minimizer open, because the filter is invertible for every positive ridge.
Under a uniform ridge, the edit $\mS^{-1}\delta'$ reaches a plain optimum $\delta'$ by overshooting each restored direction $\vu_k$ by the factor $(\lambda_k+\ridge)/\ridge$.
Two parts of the method rule that edit out.
Every gradient of the anticipated loss lies in the span of the vectors $\mS^{\top}\feat{f}$, in which a restored direction carries the weight $\ridge/(\lambda_k+\ridge)$ (Proposition~\ref{prop:release} and Corollary~\ref{cor:composition}).
The score-preservation term of~\eqref{eq:total} charges the squared residual on the retained edges, which grows with the square of an overshoot along their features.
Together they select a small edit whose surviving part forgets.

\paragraph{Subspace methods as thresholds of the filter.}
A further corollary shows that confining the edit to a subspace is a hard threshold of the filter, at level zero or at a positive level.

\begin{corollary}[Two thresholds of the filter]\label{cor:thresholds}
Take a uniform ridge in the linearized model. (a) As $\ridge\to0$, $\mS\to\Pi^{\perp}$, and for the gradient-descent update of Proposition~\ref{prop:release}, $(\Id-\eta\Phimat^{\top}\Phimat)^{T}\to\Pi^{\perp}$ as $T\to\infty$. For every ridge, the update leaves an edit with $\Phimat\delta=0$ untouched. (b) For a threshold $\bar\lambda>0$, let $U_{\bar\lambda}$ be the span of the eigenvectors $\vu_k$ with $\lambda_k\le\bar\lambda$ together with the complement of the span of the update's features. Every $\delta\in U_{\bar\lambda}$ satisfies $\|\mS\delta\|_2\ge\frac{\ridge}{\bar\lambda+\ridge}\|\delta\|_2$, and every $\delta$ in the span of the eigenvectors with $\lambda_k>\bar\lambda$ satisfies $\|\mS\delta\|_2\le\frac{\ridge}{\bar\lambda+\ridge}\|\delta\|_2$.
\end{corollary}

Confining the edit to the complement of the update's span is null-space projection, the threshold at level zero.
When the update class is the retained edges at the involved entities, it is, restricted to the involved rows, the retained-span projector that \citet{golatkar2020forgetting}, \citet{chen2024unsc} and \citet{zhou2025geometric} use to protect utility.
For an edit that moves one row alone, the constraint $\Phimat\delta=0$ reduces to the constraint on that row, and the projector is the second limit of Corollary~\ref{cor:soft}.
Read through the filter, the retained-span projector also marks the edits that continued training on retained data cannot undo to first order.
The update of Section~\ref{sec:setup} also trains on edges the unlearner never saw, whose features lie outside the retained span.
That update can therefore move an edit that lies in the retained-span null space.
The stand-ins of the unseen channel put modelled directions of this kind into $\Phimat$ (the first limit of Corollary~\ref{cor:soft}).
Confining the edit to $U_{\bar\lambda}$ is a hard threshold at a positive level, the form of the methods that place the edit in directions continued training restores slowly \citep{xiao2026mcu,sondej2026repselect,sondej2025cir}.
FRAG \citep{chen2026frag} applies the same idea per coordinate, on weights the retained data barely activate.
Those methods read the spectrum from a proxy, the activations or gradients of the forget set or of the retained data, because they do not hold the update's data.
For \citet{xiao2026mcu} the shared-eigenstructure assumption is what licenses the proxy.
The correspondence concerns the form of the constraint.
Those methods compute their spectra from the activations or gradients of deep networks and use their own losses.
The subspace $U_{\bar\lambda}$ coincides with theirs when the proxy shares its leading eigenvectors with $\Phimat^{\top}\Phimat$ (Appendix~\ref{app:survival}).
\KAU{} applies the soft weights, $\ridge/(\lambda_k+\ridge)$ under a uniform ridge, with the spectrum of its draws from the modelled update class.
That class holds the retained edges themselves on the known channel and stand-ins on the unseen one.

The remark below shows what the one-sided residual of~\eqref{eq:drive} adds to the subspace of Corollary~\ref{cor:thresholds}.

\begin{remark}[The one-sided residual relaxes the subspace to a cone]\label{rem:cone}
With the one-sided residual of~\eqref{eq:drive} in the linearized model, the response at $\orig+\delta$ is $-(\Phimat^{\top}\Phimat+\mSigma)^{-1}\Phimat^{\top}\relu(\Phimat\delta)$, with $\relu$ applied entrywise. It vanishes exactly when $\Phimat\delta\le0$ entrywise, so every such edit survives entirely. The set of edits the update leaves untouched grows from the subspace $\{\Phimat\delta=0\}$ to the cone $\{\Phimat\delta\le0\}$, which contains it.
\end{remark}

Null-space projection forbids every change to the score of an update edge, and the response of \KAU{} restores only the raised edges.

\paragraph{Composition versus a penalty on predicted recovery.}
The filter also says why the response is composed into the loss and is not penalized beside it.
The alternative, and the form a proxy method would take, is a penalty on the predicted score recovery $\hat\rho_f=\dist{f}(\theta)-\dist{f}(\theta+\resp)$ added beside a plain forgetting loss.
At $\theta=\orig$ every residual of~\eqref{eq:drive} vanishes, so $\resp=0$ and $\hat\rho_f=0$ for every fact and every draw, and the penalty attains its minimum at the unedited model.
The last bound of Proposition~\ref{prop:release} gives the local form of this fact.
A displacement inside the drawn span that undoes part of the forgetting, $\feat{f}^{(v)\top}\delta_v<0$, lowers $\dist{f}(\theta)$.
It moves the anticipated point by at most the fraction $\ridge_v/(\lambda^{+}_{\min}+\ridge_v)$ of its length.
To first order it therefore lowers $\hat\rho_f$ by $|\feat{f}^{(v)\top}\delta_v|$, up to a term of at most $\|\feat{f}^{(v)}\|_2$ times that fraction of $\|\delta_v\|_2$.
Such an objective can therefore reduce its penalty by undoing its own forgetting.
The composed objective of~\eqref{eq:objective} sees the fact only through the smaller of its current and anticipated scores, so it earns at most that term from the displacement.

\paragraph{Beyond the linearized model.}
The scores of the three backbones are nonlinear in the rows.
Outside the linearized model, Corollary~\ref{cor:composition} becomes a statement about the gradient at the current point $\theta$, under three approximations.
The features are frozen at $\theta$, which \KAU{} does too (Section~\ref{sec:method:response}).
For the TransE score the slot feature is the sign pattern of the translation error, projected on the tangent of the normalized row~\eqref{eq:slotfeat}.
The sign pattern is constant while no coordinate of the translation error changes sign, so inside such a cell the feature changes only through the normalization of the row.
The post-update score is evaluated to first order,
\begin{equation*}
\dist{f}(\theta+\resp)\;\approx\;d^{\mathrm{lin}}_f(\theta;N):=\dist{f}(\theta)+\sum_{v\in\{h,t\}}\feat{f}^{(v)\top}\Delta\ve_v .
\end{equation*}
The one-sided residual of~\eqref{eq:drive} is replaced by the signed residual $r_i=\dist{i}(\theta)-\dist{i}(\orig)$ on every drawn edge, which credits the update with restoring lowered edges as well as raised ones.
The next theorem takes the per-entity response and the gradient of the anticipated score with respect to $\ve_v$, through $v$'s own response with the other endpoint's response held fixed.
It shows that this gradient is the per-entity filter applied to $\feat{f}^{(v)}$, the direction in which a plain forgetting loss on the pre-update score would move $\ve_v$.

\begin{theorem}[Operator identity]\label{thm:operator}
Fix a draw $N$, a ridge $\ridge_v>0$ at each involved entity $v$, and a point $\theta$ at which the scores of $f$ and of the drawn edges are differentiable, which holds for almost every $\theta$. Let $\Phimat_v$ stack the slot features at $v$ of the drawn edges incident to $v$, and let $\Delta\ve_v(\theta)=-\mA_v r_v(\theta)$ with $\mA_v=\Phimat_v^{\top}(\Phimat_v\Phimat_v^{\top}+\ridge_v\Id)^{-1}$ be the per-entity form of the response of Lemma~\ref{lem:response}. Assume that the features $\Phimat_v$ and $\feat{f}^{(v)}$ are held fixed, that the post-update score is evaluated to first order, and that the residual is the signed residual $r_i=\dist{i}(\theta)-\dist{i}(\orig)$ on every drawn edge. Then for each endpoint $v$ of a forgotten fact $f$,
\begin{equation}\label{eq:operator}
\frac{\partial}{\partial\ve_v}\Big[\dist{f}(\theta)+\feat{f}^{(v)\top}\Delta\ve_v(\theta)\Big]
=\big(\Id-\Phimat_v^{\top}(\Phimat_v\Phimat_v^{\top}+\ridge_v\Id)^{-1}\Phimat_v\big)\,\feat{f}^{(v)}
=\mP^{\perp}_{N,v}\,\feat{f}^{(v)} .
\end{equation}
The operator $\mP^{\perp}_{N,v}$ is the per-entity filter of Proposition~\ref{prop:release}. It is symmetric. It scales the $k$-th right singular direction of $\Phimat_v$ by $\ridge_v/(\lambda_k+\ridge_v)$, with $\lambda_k$ the nonzero eigenvalues of $\Phimat_v\Phimat_v^{\top}$, and it leaves the orthogonal complement of the span of the drawn features unchanged. It is therefore a shrinkage operator, and it converges to the orthogonal projector off that span as $\ridge_v\to0$.
\end{theorem}

\begin{proof}[Proof sketch]
Hold $\mA_v$ fixed and let $r_v$ vary with $\theta$. Then $\partial\Delta\ve_v/\partial\ve_v=-\mA_v\,\partial r_v/\partial\ve_v$. Under the signed residual, $\dist{i}(\orig)$ is a constant and $\partial\dist{i}/\partial\ve_v=\feat{i}^{(v)\top}$ by definition of the slot feature, so $\partial r_v/\partial\ve_v=\Phimat_v$. The chain rule gives $\feat{f}^{(v)}-(\mA_v\Phimat_v)^{\top}\feat{f}^{(v)}$, which is~\eqref{eq:operator} because the inverse is symmetric. The spectral claims follow from a singular value decomposition of $\Phimat_v$. Appendix~\ref{app:thm} gives the full proof, the operator under the one-sided residual, the cross-entity terms of the joint response, and the same operator at every further Gauss--Newton round (Remark~\ref{rem:rounds}).
\end{proof}

Theorem~\ref{thm:operator} is the local form of Corollary~\ref{cor:composition} for the per-entity response.
At any point, the anticipated loss moves each row along the plain forgetting direction passed through the per-entity filter built at that point.
The draw changes at every step, so the direction the optimizer follows over a run is an average of per-draw filters.
The next corollary gives that average and two of its limits.
The first limit is the projector off the span of every retained edge and stand-in at the entity.
The second is the projector off the span of the retained gradients alone, which we call the retained-span projector and which Section~\ref{sec:exp} runs as the retained-span projection of Table~\ref{tab:ablation_recovery_depth}.

\begin{corollary}[Soft projector and its two limits]\label{cor:soft}
Under the hypotheses of Theorem~\ref{thm:operator} at every draw, replace $\tilde d_f$ in~\eqref{eq:objective} by the first-order score $d^{\mathrm{lin}}_f$, which also drops the minimum with the current score. The expected gradient of the loss term of fact $f$ with respect to $\ve_v$ through $v$'s own response is then
\begin{equation*}
-\E_N\big[w_N\,\mP^{\perp}_{N,v}\big]\feat{f}^{(v)},\qquad w_N=2\,\relu\big(t_f-d^{\mathrm{lin}}_f(\theta;N)\big)\ge0,
\end{equation*}
with $w_N$ the weight of draw $N$. Assume $\E_N[w_N]>0$ and write $\bar\mP^{\perp}_{v}=\E_N[w_N\mP^{\perp}_{N,v}]/\E_N[w_N]$ for the normalized operator. As $\ridge_v\to0$, $\bar\mP^{\perp}_{v}$ removes a unit direction $\vu$ in the proportion $\E_N[w_N\|\Pi_{N,v}\vu\|_2^2]/\E_N[w_N]$, where $\Pi_{N,v}$ is the orthogonal projector onto the span of the drawn features at $v$. Two limits follow. When every candidate edge enters with probability one, the cap $k_{\max}$ is lifted and $\ridge_v\to0$, $\bar\mP^{\perp}_{v}$ becomes the orthogonal projector off the span of all retained edges and stand-ins at $v$. That is the step direction of a projected variant that keeps the stand-ins. When the unseen channel is removed as well, $\bar\mP^{\perp}_{v}$ becomes the orthogonal projector off the span of the retained edges incident to $v$, which, for an edit of the row $\ve_v$ alone, is the retained-span projector of \citet{golatkar2020forgetting}, \citet{chen2024unsc}, \citet{saha2021gradient}, \citet{wang2021adamnscl} and \citet{zhou2025geometric} restricted to that row.
\end{corollary}

Averaged over draws, the filter removes each direction in proportion to how often, and how strongly, the modelled class draws it.
Beyond the two limits, \KAU{} adds the unseen channel, which puts edges the unlearner has never seen into the span.
It also adds the sampled draw and the exact evaluation of the score at the shifted rows.

\paragraph{Scope and empirical counterparts.}
Proposition~\ref{prop:release}, its two corollaries and Remark~\ref{rem:cone} are exact in the linearized model, and Theorem~\ref{thm:operator} is the local statement outside it.
The theorem takes the per-entity response, and Appendix~\ref{app:thm} gives the cross-entity terms of the joint solve.
Table~\ref{tab:kau_ablation_recovery_anc03} measures the joint solve against the per-entity block solve, first-order against exact evaluation, and the signed residual against \KAU{}'s, one row each. Table~\ref{tab:kau_ablation_full} measures the sampled draw against the expected draw and the plain composition against \KAU{}'s, one row each.
Table~\ref{tab:ablation_recovery_depth} lists the retained-span projection, which approaches the second limit of Corollary~\ref{cor:soft} as its ridge vanishes, released at \KAU{}'s forgetting depth.

The filter describes the modelled update, a ridge or gradient-descent fit of a squared residual on the drawn edges.
The update of Section~\ref{sec:setup} runs Adam on the ranking loss, its sampled corruptions add features outside $\Phimat$, and its features move.
Measured survival is therefore lower than the filter's.
At matched forgetting depth on TransE, the update restores $0.0111$ forget-MRR of \KAU{}'s edit beyond the oracle's own rise (Table~\ref{tab:ablation_recovery_depth}). The same update restores $0.0158$ for the retained-span projection and $0.0250$ for the plain forgetting loss.
The response predicts the order of recovery across facts (Appendix~\ref{app:predictor}).
Corollary~\ref{cor:composition} predicts where the gain appears.
The loss stays active until the surviving score reaches $t_f$, so the forgetting \KAU{} keeps is the part that the update does not restore.
Section~\ref{sec:exp} controls for depth by releasing every variant at \KAU{}'s forgetting depth and by reading the relative excess recovery against depth. In Figure~\ref{fig:depth}, \KAU{} stays below \GraphDPO{} and ZeroFisher at every depth, and the gap widens as forgetting deepens.

\subsection{Proof of Lemma~\ref{lem:response}: the linearized response}
\label{app:ridge}

\begin{proof}[Proof of Lemma~\ref{lem:response}]
Write $\Delta$ for the parameter change $\resp$ of the lemma, $r=r_N(\theta)\in\R^{|N|}$ and $\Phimat=\Phimat_N$, whose row $i$ is the transpose of the feature $\feat{i}=\nabla_\theta\dist{i}$ of update edge $i$ with respect to all trainable parameters. Linearizing the scores at $\theta$ gives $r_i(\theta+\Delta)\approx r_i+\feat{i}^{\top}\Delta$, and ridge damping adds $\tfrac{\ridge}{2}\|\Delta\|_2^2$, so the linearized response of the update is the minimizer of $\tfrac12\|r+\Phimat\Delta\|_2^2+\tfrac{\ridge}{2}\|\Delta\|_2^2$. For $\ridge>0$ this objective is strictly convex, and its unique minimizer solves the stationarity condition $(\Phimat^{\top}\Phimat+\ridge\Id)\Delta=-\Phimat^{\top}r$. Expanding both sides shows $\Phimat^{\top}(\Phimat\Phimat^{\top}+\ridge\Id)=(\Phimat^{\top}\Phimat+\ridge\Id)\Phimat^{\top}$, and multiplying this identity by $(\Phimat^{\top}\Phimat+\ridge\Id)^{-1}$ on the left and by $(\Phimat\Phimat^{\top}+\ridge\Id)^{-1}$ on the right gives the push-through identity $(\Phimat^{\top}\Phimat+\ridge\Id)^{-1}\Phimat^{\top}=\Phimat^{\top}(\Phimat\Phimat^{\top}+\ridge\Id)^{-1}$, which is~\eqref{eq:response}. A Gauss--Newton step on $\tfrac12\sum_i r_i(\theta+\Delta)^2$ from $\Delta=0$ minimizes the same linearized least-squares objective without damping, and Levenberg--Marquardt damping adds exactly the ridge term, so~\eqref{eq:response} is one damped Gauss--Newton step. The per-entity form of Section~\ref{sec:method:response} restricts the same computation to the edges incident to $v$ and to the block $\ve_v$, which gives $\Delta\ve_v=-(\Phimat_v^{\top}\Phimat_v+\ridge_v\Id)^{-1}\Phimat_v^{\top}r_v=-\mA_vr_v$ by the same identity. The relation between the ridge and a finite number of descent steps is the content of the remark below.
\end{proof}

\begin{remark}[Ridge damping and early stopping]\label{rem:ridge}
Consider the linearized least-squares problem $\tfrac12\|r+\Phimat\Delta\|_2^2$ in $\Delta$, started at $\Delta=0$. Gradient flow for time $\tau$, gradient descent with step $\eta$ for $T$ steps under $\eta\lambda_{\max}(\Kmat)\le1$, and ridge damping with parameter $\ridge$ all produce a solution of the form $\Delta=-\Phimat^{\top}\sum_k g(\lambda_k)\lambda_k^{-1}\vq_k\vq_k^{\top}r$ over the nonzero eigenpairs $(\lambda_k,\vq_k)$ of $\Kmat=\Phimat\Phimat^{\top}$, with the filters
\[
g_\tau(\lambda)=1-e^{-\tau\lambda},\qquad g_{\eta,T}(\lambda)=1-(1-\eta\lambda)^T,\qquad g_\ridge(\lambda)=\frac{\lambda}{\lambda+\ridge}.
\]
Gradient flow solves $\dot\Delta=-\Phimat^{\top}(r+\Phimat\Delta)$, whose solution is $\Delta(\tau)=-(\Id-e^{-\tau\Phimat^{\top}\Phimat})(\Phimat^{\top}\Phimat)^{+}\Phimat^{\top}r$, and this has the stated form because $(\Phimat^{\top}\Phimat)^{+}\Phimat^{\top}=\Phimat^{\top}\Kmat^{+}$ and $\Phimat^{\top}\Phimat$ has the eigenvectors $\Phimat^{\top}\vq_k/\sqrt{\lambda_k}$ with eigenvalues $\lambda_k$. Gradient descent replaces $e^{-\tau\lambda}$ by $(1-\eta\lambda)^T$, and ridge damping gives the filter $\lambda/(\lambda+\ridge)$ by the push-through identity of the proof above. With $\ridge=1/\tau$ and $\tau=\eta T$, and for every $\lambda\in[0,1/\eta]$, an interval that contains every eigenvalue of $\Kmat$ under $\eta\lambda_{\max}(\Kmat)\le1$,
\[
g_\ridge(\lambda)\le g_\tau(\lambda)\le g_{\eta,T}(\lambda)\le\min\{1,\eta T\lambda\}\le2\,g_\ridge(\lambda).
\]
The first inequality is $e^{-u}\le1/(1+u)$ at $u=\tau\lambda$. The second is $(1-\eta\lambda)^T\le e^{-\eta\lambda T}$. The third is Bernoulli's inequality $(1-\eta\lambda)^T\ge1-T\eta\lambda$ together with $g_{\eta,T}\le1$. The fourth is $\min\{1,u\}\le2u/(1+u)$, which holds for $u\le1$ because $1+u\le2$ and for $u\ge1$ because $1+u\le2u$. An early-stopped linearized update is therefore bracketed, on every eigen-direction, by the ridge filter with $\ridge=1/(\eta T)$ from below and by twice that filter from above. Sharper constants, for the risk rather than for the filter, are known \citep{ali2019continuous}. The ridge $\ridge_v$ of Section~\ref{sec:method:coupled} is set relative to the feature norms at $v$ (Appendix~\ref{app:hyper}), and every experiment runs the update as a training run rather than as a filter.
\end{remark}

\subsection{Score preservation is the known-channel residual}
\label{app:anchor-proof}

This subsection proves the claim of Section~\ref{sec:method:controls} that the score-preservation term of~\eqref{eq:total} bounds the known-channel part of the response.

\begin{proposition}[Score preservation is the known-channel residual]\label{prop:anchor}
Let $S$ be the sample of retained edges at the involved entities in~\eqref{eq:total}. The score-preservation term of~\eqref{eq:total}, $\beta\,\mean_{i\in S}\big(\dist{i}(\theta)-\dist{i}(\orig)\big)^2$, is $\beta$ times the mean squared signed residual of the known channel on $S$, and it bounds $\beta$ times the mean squared one-sided residual of~\eqref{eq:drive} on $S$ from above. Let $r^{\mathrm{kn}}_v$ be $r_v$ with the entries of the stand-ins, the unseen-channel edges of Section~\ref{sec:method:draw}, set to zero. The part of the per-entity response built from the known edges satisfies $\|\mA_v r^{\mathrm{kn}}_v\|_2\le\|\mA_v\|_{\mathrm{op}}\|r^{\mathrm{kn}}_v\|_2$ with $\|\mA_v\|_{\mathrm{op}}\le1/(2\sqrt{\ridge_v})$, and whenever $S$ contains the drawn retained edges at $v$,
\begin{equation*}
\|\mA_v r^{\mathrm{kn}}_v\|_2\le\sqrt{\frac{|S|}{4\ridge_v}}\;\Big(\mean_{i\in S}\big(\dist{i}(\theta)-\dist{i}(\orig)\big)^2\Big)^{1/2},
\end{equation*}
so minimizing the term on such a sample bounds that displacement. The joint solve of Section~\ref{sec:method:coupled} obeys the same bound for the stacked known-channel displacement, with $\min_v\ridge_v$ in place of $\ridge_v$, whenever $S$ contains every drawn retained edge.
\end{proposition}

\begin{proof}[Proof of Proposition~\ref{prop:anchor}]
For a retained edge $i\in S$ at an involved entity, the signed residual of the known channel is $\dist{i}(\theta)-\dist{i}(\orig)$ by definition, so $\beta\,\mean_{i\in S}(\dist{i}(\theta)-\dist{i}(\orig))^2$ is $\beta$ times the mean squared signed residual on $S$, which is the first claim. Since $\relu(x)^2\le x^2$ for every real $x$, the same quantity bounds $\beta$ times the mean squared one-sided residual of~\eqref{eq:drive} on $S$ from above, which is the second. For the third, the per-entity response decomposes by linearity as $-\mA_vr_v=-\mA_vr^{\mathrm{kn}}_v-\mA_vr^{\mathrm{un}}_v$, where $r^{\mathrm{un}}_v$ keeps the stand-in entries, and the part built from the known edges satisfies $\|\mA_vr^{\mathrm{kn}}_v\|_2\le\|\mA_v\|_{\mathrm{op}}\|r^{\mathrm{kn}}_v\|_2$ with $\|\mA_v\|_{\mathrm{op}}\le1/(2\sqrt{\ridge_v})$ by the third fact following~\eqref{eq:spectral}. By $\relu(x)^2\le x^2$ once more, $\|r^{\mathrm{kn}}_v\|_2^2\le\sum_i(\dist{i}(\theta)-\dist{i}(\orig))^2$ with the sum over the drawn retained edges at $v$, and whenever $S$ contains those edges this sum is at most $\sum_{i\in S}(\dist{i}(\theta)-\dist{i}(\orig))^2=|S|\,\mean_{i\in S}(\dist{i}(\theta)-\dist{i}(\orig))^2$, which gives the displayed bound. The unseen part $\mA_vr^{\mathrm{un}}_v$ is not controlled by the term. For the joint solve, write $\mSigma=\operatorname{diag}(\ridge_v\Id)$ and $\tilde\Phimat=\Phimat\mSigma^{-1/2}$. Then $\Phimat^{\top}\Phimat+\mSigma=\mSigma^{1/2}(\tilde\Phimat^{\top}\tilde\Phimat+\Id)\mSigma^{1/2}$, so $(\Phimat^{\top}\Phimat+\mSigma)^{-1}\Phimat^{\top}=\mSigma^{-1/2}(\tilde\Phimat^{\top}\tilde\Phimat+\Id)^{-1}\tilde\Phimat^{\top}$, whose operator norm is at most $\|\mSigma^{-1/2}\|_{\mathrm{op}}\cdot\tfrac12=1/(2\sqrt{\min_v\ridge_v})$ by the same fact with ridge one. The bound on the known-channel part of the joint displacement follows with $\min_v\ridge_v$ in place of $\ridge_v$.
\end{proof}

\subsection{Beyond the linearized response}
\label{app:rounds}

Lemma~\ref{lem:response} gives the linearized response of the whole run of the update, modelled as a ridge-damped least-squares fit under frozen features. This subsection bounds what one re-linearization round adds and shows that the operator identity survives re-linearization. Fix a draw $N$ with $n=|N|$ edges, the features $\Phimat$ with rows $\feat{i}^{\top}$ at $\theta$, the kernel $\Kmat=\Phimat\Phimat^{\top}$, a ridge $\ridge>0$, and the residual $r=r(\theta)$ with $r_i(\theta+\Delta)=\dist{i}(\theta+\Delta)-y_i$ as in the lemma. The ridge-damped update minimizes
\[
F(\Delta)=\tfrac12\sum_{i\in N}r_i(\theta+\Delta)^2+\tfrac{\ridge}{2}\|\Delta\|_2^2 .
\]
Its Gauss--Newton iteration starts at $\Delta_0=0$ and sets
\[
\Delta_{k+1}=\Delta_k-(\Phimat_k^{\top}\Phimat_k+\ridge\Id)^{-1}\big(\Phimat_k^{\top}r(\theta+\Delta_k)+\ridge\Delta_k\big),
\]
with $\Phimat_k$ the features at $\theta+\Delta_k$. The first iterate is $\Delta_1=-(\Phimat^{\top}\Phimat+\ridge\Id)^{-1}\Phimat^{\top}r=-\Phimat^{\top}(\Kmat+\ridge\Id)^{-1}r$, which is the response of Lemma~\ref{lem:response}. Write $\alpha=(\Kmat+\ridge\Id)^{-1}r$ for the dual coefficients, so that $\Delta_1=-\Phimat^{\top}\alpha$, and write $r^{\mathrm{lin}}(\Delta)=r+\Phimat\Delta$ for the linearized residual, so that $r^{\mathrm{lin}}(\Delta_1)=r-\Kmat\alpha=\ridge\alpha$.

\begin{lemma}[One-round error]\label{lem:rounds}
Assume that along the segment $\{\theta+t\Delta_1:t\in[0,1]\}$ every score $\dist{i}$, $i\in N$, is differentiable, with features $\feat{i}=\nabla_\theta\dist{i}$ satisfying $\|\feat{i}(\theta+t\Delta_1)-\feat{i}(\theta+s\Delta_1)\|_2\le L\,|t-s|\,\|\Delta_1\|_2$ for all $s,t\in[0,1]$. Then (i) $|r_i(\theta+\Delta_1)-r^{\mathrm{lin}}_i(\Delta_1)|\le\tfrac{L}{2}\|\Delta_1\|_2^2$ for every $i\in N$, (ii) $\|\Phimat_1-\Phimat\|_{\mathrm{op}}\le\sqrt{n}\,L\,\|\Delta_1\|_2$, and (iii)
\[
\|\Delta_2-\Delta_1\|_2\le\frac{\sqrt{n}\,L}{4\sqrt{\ridge}}\|\Delta_1\|_2^2+\sqrt{n}\,L\,\|\Delta_1\|_2\|\alpha\|_2\le\sqrt{n}\,L\,\|\Phimat\|_{\mathrm{op}}\Big(\frac{\|\Phimat\|_{\mathrm{op}}}{4\sqrt{\ridge}}+1\Big)\|\alpha\|_2^2 .
\]
\end{lemma}
\begin{proof}
(i) By the fundamental theorem of calculus along the segment,
\[
r_i(\theta+\Delta_1)-r_i(\theta)-\feat{i}^{\top}\Delta_1=\int_0^1\big(\feat{i}(\theta+t\Delta_1)-\feat{i}(\theta)\big)^{\top}\Delta_1\,dt,
\]
and the Lipschitz hypothesis bounds the integrand by $L\,t\,\|\Delta_1\|_2^2$, whose integral is $\tfrac{L}{2}\|\Delta_1\|_2^2$. (ii) Each row of $\Phimat_1-\Phimat$ is $\feat{i}(\theta+\Delta_1)-\feat{i}(\theta)$, of norm at most $L\|\Delta_1\|_2$, so the Frobenius norm, which dominates the operator norm, is at most $\sqrt{n}\,L\|\Delta_1\|_2$. (iii) The gradient of $F$ at $\Delta_1$ is $\nabla F(\Delta_1)=\Phimat_1^{\top}r(\theta+\Delta_1)+\ridge\Delta_1$. Since $\Delta_1$ minimizes the linearized objective $\tfrac12\|r^{\mathrm{lin}}(\Delta)\|_2^2+\tfrac{\ridge}{2}\|\Delta\|_2^2$, it satisfies $\Phimat^{\top}r^{\mathrm{lin}}(\Delta_1)+\ridge\Delta_1=0$. Subtracting,
\[
\nabla F(\Delta_1)=\Phimat_1^{\top}e+(\Phimat_1-\Phimat)^{\top}r^{\mathrm{lin}}(\Delta_1)=\Phimat_1^{\top}e+\ridge(\Phimat_1-\Phimat)^{\top}\alpha,
\]
where $e=r(\theta+\Delta_1)-r^{\mathrm{lin}}(\Delta_1)$ has $\|e\|_2\le\sqrt{n}\,\tfrac{L}{2}\|\Delta_1\|_2^2$ by (i). By definition $\Delta_2-\Delta_1=-(\Phimat_1^{\top}\Phimat_1+\ridge\Id)^{-1}\nabla F(\Delta_1)$. The derivation of~\eqref{eq:spectral} holds for any matrix and any $\ridge>0$. Applied to $\Phimat_1$, the push-through identity of the proof of Lemma~\ref{lem:response} and the third fact following~\eqref{eq:spectral} give $\|(\Phimat_1^{\top}\Phimat_1+\ridge\Id)^{-1}\Phimat_1^{\top}\|_{\mathrm{op}}\le1/(2\sqrt{\ridge})$, and $\|(\Phimat_1^{\top}\Phimat_1+\ridge\Id)^{-1}\|_{\mathrm{op}}\le1/\ridge$. Hence $\|\Delta_2-\Delta_1\|_2\le\|e\|_2/(2\sqrt{\ridge})+\|\Phimat_1-\Phimat\|_{\mathrm{op}}\|\alpha\|_2$, which is the first inequality of (iii) after (ii). The second follows from $\|\Delta_1\|_2=\|\Phimat^{\top}\alpha\|_2\le\|\Phimat\|_{\mathrm{op}}\|\alpha\|_2$.
\end{proof}

Both terms of (iii) are quadratic in the dual coefficients $\alpha$, hence of second order in the residual the edit left on the drawn edges. The first term is the curvature of the scores along the displacement. The second is the drift of the features weighted by $\ridge\alpha$, the residual the linearized fit leaves. Each further round repeats the step from the new point and costs one more solve on the same draw. Appendix~\ref{app:variants} records that a second round on the same draw improved no reported quantity, which is why the objective uses one round. Appendix~\ref{app:predictor} reports that the one-round response tracks the realized per-fact score recovery with a Pearson correlation of $0.53$ and a Spearman correlation of $0.21$, and it predicts the ranking of the facts and not the level of the change. The lemma bounds the correction of one further round. It does not bound the level gap of an update of many epochs, and the objective relies on the ranking while the target sets the level (Section~\ref{sec:method:pess}).

For TransE and TransH the score is piecewise linear in the translation error of the triple and its slot features jump where a coordinate of that vector changes sign. Along a segment that crosses no sign change the sign vector $\vs$ is constant and the feature $(\Id-\bar\ve_v\bar\ve_v^{\top})\vs/\|\ve_v\|_2$ is smooth in the row. The Lipschitz hypothesis then holds with a constant set by the derivative of the normalization Jacobian, of order $\sqrt m/\|\ve_v\|_2^2$ on a region where the row norms are bounded below. Across a sign change the hypothesis fails. By Lemma~\ref{lem:pwl} the gap of item (i) is then the normalization's second-order term plus the crossed-coordinate excess $\varkappa(\bm D)$ of Appendix~\ref{app:pwlnote}, which is non-negative and, in a coordinate with $|x_j|$ small against $|D_j|$, of first order in the displacement rather than second. Appendix~\ref{app:predictor} measures its effect on the ranking.

\begin{remark}[The operator identity round by round]\label{rem:rounds}
Hold the features fixed at every round and use the signed residual, as in Theorem~\ref{thm:operator}, so that $\Phimat_k$ enters through the Jacobian of the residual at $\theta+\Delta_k$ but is not itself differentiated. Write $\mG_k=(\Phimat_k^{\top}\Phimat_k+\ridge\Id)^{-1}$. Differentiating the iteration with respect to $\theta$ gives
\[
\frac{\partial\Delta_{k+1}}{\partial\theta}=\frac{\partial\Delta_k}{\partial\theta}-\mG_k\Big(\Phimat_k^{\top}\Phimat_k\Big(\Id+\frac{\partial\Delta_k}{\partial\theta}\Big)+\ridge\,\frac{\partial\Delta_k}{\partial\theta}\Big)=-\mG_k\Phimat_k^{\top}\Phimat_k,
\]
because $\mG_k(\Phimat_k^{\top}\Phimat_k+\ridge\Id)=\Id$. The derivative of the anticipated point $\theta+\Delta_{k+1}$ is therefore $\Id-\mG_k\Phimat_k^{\top}\Phimat_k=\Id-\Phimat_k^{\top}(\Phimat_k\Phimat_k^{\top}+\ridge\Id)^{-1}\Phimat_k$ by the push-through identity of the proof of Lemma~\ref{lem:response}, a symmetric matrix, hence the shrinkage operator of~\eqref{eq:operator} built from the features at the $k$-th anticipated point. Earlier rounds enter only through the point at which those features are taken. This operator is the implicit Jacobian of a proximal model of adaptation with respect to its centre \citep{rajeswaran2019imaml}. Theorem~\ref{thm:operator} therefore holds for every round with $\Phimat_k$ in place of $\Phimat$, and so does Proposition~\ref{prop:release} when the residual at the $k$-th anticipated point changes by $\Phimat_k$ times the motion of that point, which holds to first order in the displacement. The per-entity form of the theorem is the block-diagonal case, and the computation is unchanged with $\mSigma$ in place of $\ridge\Id$.
\end{remark}

\subsection{Proof of Theorem~\ref{thm:operator}: the operator identity}
\label{app:thm}

\begin{proof}[Proof of Theorem~\ref{thm:operator}]
Let $u$ be the other endpoint of $f$ and write
\[
d^{\mathrm{lin}}_f(\theta)=\dist{f}(\theta)+\feat{f}^{(v)\top}\Delta\ve_v(\theta)+\feat{f}^{(u)\top}\Delta\ve_u(\theta)
\]
for the anticipated score evaluated to first order, with $\feat{f}^{(v)}$, $\feat{f}^{(u)}$, $\mA_v$ and $\mA_u$ held fixed. Its derivative with respect to $\ve_v$ has three terms. The first is $\partial\dist{f}/\partial\ve_v=\feat{f}^{(v)}$, by definition of the slot feature. The second is $(\partial\Delta\ve_v/\partial\ve_v)^{\top}\feat{f}^{(v)}$. The third is the cross-entity term $C_v=(\partial\Delta\ve_u/\partial\ve_v)^{\top}\feat{f}^{(u)}$, treated in part~(c) below. The theorem concerns the first two, which are the derivative of $\dist{f}+\feat{f}^{(v)\top}\Delta\ve_v$.

\emph{(a) The signed residual.} Since $\Delta\ve_v=-\mA_vr_v(\theta)$ with $\mA_v$ fixed and $r_v$ varying with $\theta$, $\partial\Delta\ve_v/\partial\ve_v=-\mA_v\,\partial r_v/\partial\ve_v$. Under the signed residual $r_i=\dist{i}(\theta)-\dist{i}(\orig)$ the subtracted term is a constant, so $\partial r_i/\partial\ve_v=\feat{i}^{(v)\top}$ for every $i\in P_v(N)$, that is, $\partial r_v/\partial\ve_v=\Phimat_v$. The matrix that builds the response operator therefore enters a second time through the residual. Hence $\partial\Delta\ve_v/\partial\ve_v=-\mA_v\Phimat_v$ and
\[
\feat{f}^{(v)}+(-\mA_v\Phimat_v)^{\top}\feat{f}^{(v)}=\feat{f}^{(v)}-\Phimat_v^{\top}(\Kmat_v+\ridge_v\Id)^{-1}\Phimat_v\feat{f}^{(v)}=\mP^{\perp}_{N,v}\feat{f}^{(v)},
\]
where the transpose was taken using the symmetry of $(\Kmat_v+\ridge_v\Id)^{-1}$. This is~\eqref{eq:operator}. The spectral claims of the statement are the first two facts read off~\eqref{eq:spectral}. For the loss of draw $N$, $\relu(t_f-d^{\mathrm{lin}}_f)^2$, the chain rule gives $\partial\relu(t_f-d^{\mathrm{lin}}_f)^2/\partial\ve_v=-2\relu(t_f-d^{\mathrm{lin}}_f)\,(\mP^{\perp}_{N,v}\feat{f}^{(v)}+C_v)$.

\emph{(b) The one-sided residual.} Under $r_i=\relu(\dist{i}(\theta)-\dist{i}(\orig))$, almost everywhere $\partial r_i/\partial\ve_v=\mathbf 1[r_i>0]\,\feat{i}^{(v)\top}$. Let $\mathcal A\subseteq P_v(N)$ be the edges with positive residual and $\mD_{\mathcal A}$ the diagonal matrix with entry one on $\mathcal A$ and zero elsewhere. Then $\partial\Delta\ve_v/\partial\ve_v=-\mA_v\mD_{\mathcal A}\Phimat_v$, whose transpose is $-\Phimat_v^{\top}\mD_{\mathcal A}(\Kmat_v+\ridge_v\Id)^{-1}\Phimat_v$, and the operator of~\eqref{eq:operator} becomes
\[
\mM_{N,v}=\Id-\Phimat_v^{\top}\mD_{\mathcal A}(\Kmat_v+\ridge_v\Id)^{-1}\Phimat_v .
\]
Only the edges with positive residual enter through the residual, whereas every drawn edge still enters through the kernel. For $\ridge_v>0$ and $\mathcal A\neq P_v(N)$ the operator $\mM_{N,v}$ is in general not symmetric and its norm can exceed one, so the shrinkage reading of Theorem~\ref{thm:operator} is specific to the signed residual. When every drawn edge has positive residual, $\mM_{N,v}=\mP^{\perp}_{N,v}$. In the limit $\ridge_v\to0$ with $\Kmat_v$ invertible, write $\Phimat_{\mathcal A}$ for the rows of $\Phimat_v$ in $\mathcal A$ and $\Kmat_{:,\mathcal A}=\Phimat_v\Phimat_{\mathcal A}^{\top}$ for the corresponding columns of $\Kmat_v$. Since $\Kmat_v^{-1}\Kmat_{:,\mathcal A}$ is the matrix that selects the columns in $\mathcal A$ and $\mD_{\mathcal A}$ leaves those columns unchanged, $\mM_{N,v}\Phimat_{\mathcal A}^{\top}=\Phimat_{\mathcal A}^{\top}-\Phimat_v^{\top}\mD_{\mathcal A}\Kmat_v^{-1}\Kmat_{:,\mathcal A}=\Phimat_{\mathcal A}^{\top}-\Phimat_{\mathcal A}^{\top}=0$, and
\[
\mM_{N,v}^2=\Id-2\Phimat_v^{\top}\mD_{\mathcal A}\Kmat_v^{-1}\Phimat_v+\Phimat_v^{\top}\mD_{\mathcal A}\Kmat_v^{-1}\Kmat_v\mD_{\mathcal A}\Kmat_v^{-1}\Phimat_v=\Id-\Phimat_v^{\top}\mD_{\mathcal A}\Kmat_v^{-1}\Phimat_v=\mM_{N,v},
\]
using $\mD_{\mathcal A}^2=\mD_{\mathcal A}$. The limiting operator is thus an oblique projector that annihilates the active features. The set $\mathcal A$ depends on $\theta$, so the identity holds on each region of parameter space on which the active set is constant.

\emph{(c) Cross-entity terms and the joint response.} Under the per-entity response, $\partial\Delta\ve_u/\partial\ve_v=-\mA_u\,\partial r_u/\partial\ve_v$, and for $i\in P_u(N)$ the derivative $\partial r_i/\partial\ve_v$ is nonzero only if $v$ is the other endpoint of $i$. Hence $C_v=0$ unless some drawn edge at $u$ has $v$ as its other endpoint, in which case $C_v=-\Phimat_{u|v}^{\top}\mA_u^{\top}\feat{f}^{(u)}$, where $\Phimat_{u|v}\in\R^{n_u\times m}$ has the row $\feat{i}^{(v)\top}$ for $i\in P_u(N)\cap P_v(N)$ and zero rows elsewhere. Under the joint response of Section~\ref{sec:method:coupled}, stack the involved rows as $\Delta\in\R^{m|\Vinv|}$, let $\Phimat\in\R^{|N|\times m|\Vinv|}$ be the full Jacobian of the drawn edges, whose row $i$ carries $\feat{i}^{(h)\top}$ in the block of its head and $\feat{i}^{(t)\top}$ in the block of its tail, and let $\feat{f}\in\R^{m|\Vinv|}$ be the stacked feature of $f$, nonzero in the blocks of $h$ and $t$. Then $\Delta=-(\Phimat^{\top}\Phimat+\mSigma)^{-1}\Phimat^{\top}r$, the signed residual has Jacobian $\Phimat$ with respect to the stacked involved rows, and the gradient of $\dist{f}+\feat{f}^{\top}\Delta$ with respect to the stacked rows is
\[
\feat{f}-\Phimat^{\top}\Phimat(\Phimat^{\top}\Phimat+\mSigma)^{-1}\feat{f}=\mSigma(\Phimat^{\top}\Phimat+\mSigma)^{-1}\feat{f}.
\]
For a uniform ridge $\mSigma=\ridge\Id$ the push-through identity turns this into $\big(\Id-\Phimat^{\top}(\Phimat\Phimat^{\top}+\ridge\Id)^{-1}\Phimat\big)\feat{f}$, the operator of~\eqref{eq:operator} with the full Jacobian in place of the per-entity one, and for a general diagonal $\mSigma$ it is the similarity transform $\mSigma^{1/2}\big(\Id-\tilde\Phimat^{\top}(\tilde\Phimat\tilde\Phimat^{\top}+\Id)^{-1}\tilde\Phimat\big)\mSigma^{-1/2}\feat{f}$ with $\tilde\Phimat=\Phimat\mSigma^{-1/2}$. The block of this operator at $(v,v)$, applied to $\feat{f}^{(v)}$, is the part of the derivative with respect to $\ve_v$ that passes through $v$'s own displacement, and its block at $(v,u)$, applied to $\feat{f}^{(u)}$, is the cross-entity term. By Lemma~\ref{lem:jacobi}, $\Phimat^{\top}\Phimat+\mSigma$ is block diagonal when no drawn edge has two involved endpoints, and in that case the diagonal block reduces to $\mP^{\perp}_{N,v}$ and every cross-entity term vanishes. Otherwise the inverse of the block-sparse positive definite matrix $\Phimat^{\top}\Phimat+\mSigma$ is generically dense on each connected component of the graph formed by the drawn edges, so the cross-entity terms generically couple every pair of involved entities joined by a path of drawn edges. No per-entity projector contains these terms.
\end{proof}

\subsection{Proof of Corollary~\ref{cor:soft}: the soft projector and its two limits}
\label{app:cor}

\begin{proof}[Proof of Corollary~\ref{cor:soft}]
\emph{The draw.} At entity $v$ the candidate pool $\mathcal C_v$ consists of the retained edges incident to $v$, each included independently with probability $\rho_{\mathrm{kn}}$, and the stand-ins of $v$, each included independently with probability $p_{\mathrm{un}}(v)=\min\{1,\rho_{\mathrm{un}}\deg(v)/32\}$. When more than $k_{\max}$ edges are included, a uniformly random subset of size $k_{\max}$ is kept. The result is a probability distribution over subsets $P_v(N)\subseteq\mathcal C_v$, and every expectation below is a finite sum.

\emph{The soft projector.} By Theorem~\ref{thm:operator}, the gradient of the loss of draw $N$ through $v$'s own response is $-w_N\mP^{\perp}_{N,v}\feat{f}^{(v)}$ with $w_N=2\relu(t_f-d^{\mathrm{lin}}_f(\theta;N))\ge0$, and its expectation over the draw is $-\E_N[w_N\mP^{\perp}_{N,v}]\feat{f}^{(v)}$, a non-negative combination of symmetric matrices with spectra in $(0,1]$. When $\E_N[w_N]>0$, the normalized operator $\bar\mP^{\perp}_{v}=\E_N[w_N\mP^{\perp}_{N,v}]/\E_N[w_N]$ is symmetric with spectrum in $(0,1]$. For a unit vector $\vu$, by~\eqref{eq:spectral},
\[
\vu^{\top}(\Id-\mP^{\perp}_{N,v})\vu=\sum_{k\le\varrho}\frac{\lambda_k}{\lambda_k+\ridge_v}\langle\vu,\vu_k\rangle^2
\in\Big[\frac{\lambda^{+}_{\min}(\Kmat_v)}{\lambda^{+}_{\min}(\Kmat_v)+\ridge_v}\,\|\Pi_{N,v}\vu\|_2^2,\ \|\Pi_{N,v}\vu\|_2^2\Big],
\]
so the fraction of $\vu$ that $\bar\mP^{\perp}_{v}$ removes, $\vu^{\top}(\Id-\bar\mP^{\perp}_{v})\vu$, lies between the $w$-weighted expectations of the two ends and equals $\E_N[w_N\|\Pi_{N,v}\vu\|_2^2]/\E_N[w_N]$ as $\ridge_v\to0$. This is the weighted expected squared fraction of $\vu$ that falls inside the drawn span. With equal weights the operator is $\E_N[\mP^{\perp}_{N,v}]$.

\emph{Limit 1, the update span.} If $\rho_{\mathrm{kn}}\to1$, $p_{\mathrm{un}}(v)\to1$ and the cap is lifted, then $P_v(N)=\mathcal C_v$ with probability one, the expectation over draws reduces to the single term $\mP^{\perp}_{\mathcal C_v,v}$, and $\ridge_v\to0$ gives $\Pi^{\perp}_{\mathcal C_v,v}$ by~\eqref{eq:spectral}, the orthogonal projector off the span of all retained edges and stand-ins at $v$. A first-order projected variant that keeps the stand-ins applies this projector to the gradient of a plain forgetting loss on the pre-update score, $-2\relu(t_f-\dist{f}(\theta))\,\Pi^{\perp}_{\mathcal C_v,v}\feat{f}^{(v)}$, and the limit of the corollary is $-w\,\Pi^{\perp}_{\mathcal C_v,v}\feat{f}^{(v)}$ with $w$ evaluated on the linearized anticipated score. The two steps share their direction and differ in the scalar that weights it.

\emph{Limit 2, the retained span.} If $\rho_{\mathrm{kn}}\to1$, $\rho_{\mathrm{un}}\to0$ and the cap is lifted, every stand-in has inclusion probability tending to zero and every retained edge has inclusion probability tending to one, so $P_v(N)$ converges in distribution to the set of retained edges incident to $v$, and $\ridge_v\to0$ gives the orthogonal projector off their span. Retained-span methods project the unlearner's edit off the span of the per-example gradients of a retained loss. For a loss that depends on $\theta$ through the retained scores, the gradient of the term of retained edge $i$ with respect to $\ve_v$ is $c_i\feat{i}^{(v)}$ with a scalar $c_i$, so the span of the per-edge retained gradients at $v$ is contained in the span of the retained slot features at $v$, with equality when every $c_i\neq0$. A ranking loss with sampled negatives adds the features of corrupted triples at $v$, which this identification leaves out. For an edit confined to the row $\ve_v$, the constraint of those methods reads $\feat{i}^{(v)\top}\delta_v=0$ over the retained edges at $v$, which is this projector, and that is the sense of the restriction to the row $\ve_v$ in the statement. Their projector over several rows at once is the joint one and is not the stack of per-entity projectors (Lemma~\ref{lem:jacobi}). Those methods apply the projector to the unlearner's edit to protect the retained set. Here the same operator arises as the direction of the anticipated loss.

\emph{The $m-1$ condition.} By~\eqref{eq:slotfeat}, on the two normalized backbones every slot feature at $v$, including $\feat{f}^{(v)}$, lies in $\bar\ve_v^{\perp}$, of dimension $m-1$. If the drawn edges at $v$ satisfy $\operatorname{rank}\Phimat_v=m-1$, then $\operatorname{range}(\Phimat_v^{\top})=\bar\ve_v^{\perp}\ni\feat{f}^{(v)}$ and $\Pi^{\perp}_{N,v}\feat{f}^{(v)}=0$, while for $\ridge_v>0$ the bound $\|\mP^{\perp}_{N,v}\feat{f}^{(v)}\|_2\le\ridge_v/(\lambda^{+}_{\min}(\Kmat_v)+\ridge_v)\,\|\feat{f}^{(v)}\|_2$ holds by the computation in the proof of Proposition~\ref{prop:release}. Rank $m-1$ requires at least $m-1$ edges, so an entity whose degree in the update class is below $m-1$ never reaches this regime. Under the cap, $\operatorname{rank}\Phimat_v\le k_{\max}=96<m-1=199$ for the $200$-wide rows of TransE and TransH, and $k_{\max}<m=400$ for the unnormalized $400$-wide rows of RotatE, where the regime requires rank $m$. Each draw therefore leaves at least $m-1-k_{\max}$ tangent directions at $v$ untouched ($m-k_{\max}$ directions on RotatE), and no draw of the sampled objective enters the regime. Both limits, which lift the cap, and the update itself on an entity of high degree, whose edges are not capped, can enter it.
\end{proof}

\subsection{Proofs for the survival filter}
\label{app:survival}

This subsection proves the statements of Appendix~\ref{app:theory} that are made in the linearized model, namely the joint form of Proposition~\ref{prop:release}, Corollaries~\ref{cor:composition} and~\ref{cor:thresholds}, and Remark~\ref{rem:cone}. Throughout, $\Phimat\in\R^{|N|\times m|\Vinv|}$ is the full Jacobian of the drawn edges over the stacked involved rows, as in part~(c) of the proof of Theorem~\ref{thm:operator}, frozen at $\orig$. We write $\mC=\Phimat^{\top}\Phimat$, $\Kmat=\Phimat\Phimat^{\top}$ and $\mSigma=\operatorname{diag}(\ridge_v\Id)$ with every $\ridge_v>0$, so that $\mC+\mSigma$ is positive definite. The edit is $\delta=\theta-\orig\in\R^{m|\Vinv|}$, the stacked feature of a forgotten fact is $\feat{f}\in\R^{m|\Vinv|}$, the residual is $r(\theta)=\Phimat(\theta-\orig)$, and the response is $\resp(\theta)=-(\mC+\mSigma)^{-1}\Phimat^{\top}r(\theta)$, the solution of~\eqref{eq:coupled}.

\begin{proof}[Proof of Proposition~\ref{prop:release}, joint form]
\emph{The identity.} The residual is affine, $r(\theta+\delta)=r(\theta)+\Phimat\delta$, so
\[
\resp(\theta+\delta)=-(\mC+\mSigma)^{-1}\Phimat^{\top}r(\theta)-(\mC+\mSigma)^{-1}\Phimat^{\top}\Phimat\delta=\resp(\theta)-(\mC+\mSigma)^{-1}\mC\,\delta .
\]
Adding $\theta+\delta$ to both sides gives $\theta+\delta+\resp(\theta+\delta)=\theta+\resp(\theta)+\big(\Id-(\mC+\mSigma)^{-1}\mC\big)\delta$. Since $(\mC+\mSigma)^{-1}(\mC+\mSigma)=\Id$,
\[
\Id-(\mC+\mSigma)^{-1}\mC=(\mC+\mSigma)^{-1}\big((\mC+\mSigma)-\mC\big)=(\mC+\mSigma)^{-1}\mSigma=\mS ,
\]
which is the identity at every $\theta$. At $\theta=\orig$ the residual is $r(\orig)=0$, hence $\resp(\orig)=0$, and the identity reads $\orig+\delta+\resp(\orig+\delta)=\orig+\mS\delta$, which is~\eqref{eq:survival}.

\emph{The edits left untouched.} By the last display, $\mS\delta=\delta$ holds if and only if $(\mC+\mSigma)^{-1}\mC\delta=0$, that is, if and only if $\mC\delta=0$, because $(\mC+\mSigma)^{-1}$ is invertible. If $\Phimat\delta=0$ then $\mC\delta=\Phimat^{\top}\Phimat\delta=0$. Conversely, $\mC\delta=0$ gives $\|\Phimat\delta\|_2^2=\delta^{\top}\mC\delta=0$. Hence $\mS\delta=\delta$ if and only if $\Phimat\delta=0$.

\emph{The uniform ridge.} Let $\mSigma=\ridge\Id$. The push-through identity of the proof of Lemma~\ref{lem:response} gives $(\mC+\ridge\Id)^{-1}\Phimat^{\top}=\Phimat^{\top}(\Kmat+\ridge\Id)^{-1}$, so
\[
\mS=\Id-(\mC+\ridge\Id)^{-1}\Phimat^{\top}\Phimat=\Id-\Phimat^{\top}(\Kmat+\ridge\Id)^{-1}\Phimat .
\]
The derivation of~\eqref{eq:spectral} uses only a singular value decomposition of the matrix and $\ridge>0$, so it applies to the stacked $\Phimat$. With $\Phimat=\sum_{k\le\varrho}\sqrt{\lambda_k}\,\vq_k\vu_k^{\top}$, it gives
\begin{equation}\label{eq:filter-spectral}
\mS=\sum_{k\le\varrho}\frac{\ridge}{\lambda_k+\ridge}\,\vu_k\vu_k^{\top}+\Pi^{\perp},\qquad\Pi^{\perp}=\Id-\sum_{k\le\varrho}\vu_k\vu_k^{\top}.
\end{equation}
The vectors $\vu_k$ are orthonormal eigenvectors of $\mC=\sum_{k\le\varrho}\lambda_k\vu_k\vu_k^{\top}$ with the eigenvalues $\lambda_k>0$, which are also the nonzero eigenvalues of $\Kmat$. They span $\operatorname{range}(\Phimat^{\top})$, the span of the update's features, and $\Pi^{\perp}$ is the orthogonal projector off that span. Expanding $\delta=\sum_{k\le\varrho}c_k\vu_k+\Pi^{\perp}\delta$ with $c_k=\langle\vu_k,\delta\rangle$ gives $\mS\delta=\sum_{k\le\varrho}\frac{\ridge}{\lambda_k+\ridge}c_k\vu_k+\Pi^{\perp}\delta$. The component along $\vu_k$ is multiplied by $\ridge/(\lambda_k+\ridge)$ and the component off the span is unchanged.

\emph{Gradient descent.} Let the update start at the edited model $\orig+\delta$ and take $T$ gradient steps of size $\eta$ on the linearized squared residual. Write $\epsilon_t$ for the change after $t$ steps, with $\epsilon_0=0$. The loss is $\tfrac12\|r(\orig+\delta+\epsilon)\|_2^2=\tfrac12\|\Phimat(\delta+\epsilon)\|_2^2$, and its gradient in $\epsilon$ is $\mC(\delta+\epsilon)$. The recursion is therefore $\epsilon_{t+1}=\epsilon_t-\eta\,\mC(\delta+\epsilon_t)$, which reads $\delta+\epsilon_{t+1}=(\Id-\eta\mC)(\delta+\epsilon_t)$. Induction on $t$ from $\delta+\epsilon_0=\delta$ gives $\delta+\epsilon_T=(\Id-\eta\mC)^{T}\delta$, the surviving edit. In the eigenbasis of $\mC$,
\[
(\Id-\eta\mC)^{T}=\sum_{k\le\varrho}(1-\eta\lambda_k)^{T}\vu_k\vu_k^{\top}+\Pi^{\perp},
\]
so the surviving fraction along $\vu_k$ is $(1-\eta\lambda_k)^{T}$, and it lies in $[0,1)$ because $0<\eta\lambda_k\le1$ under $\eta\le1/\lambda_{\max}(\Kmat)$. The first two inequalities of the chain in Remark~\ref{rem:ridge} state $g_\ridge(\lambda)\le g_{\eta,T}(\lambda)$ on $[0,1/\eta]$ at $\ridge=1/(\eta T)$, that is, $\lambda/(\lambda+\ridge)\le1-(1-\eta\lambda)^{T}$. Rearranged at $\lambda=\lambda_k$ this is $(1-\eta\lambda_k)^{T}\le\ridge/(\lambda_k+\ridge)$. The last inequality of the same chain, $g_{\eta,T}\le2g_\ridge$, gives the lower bound $(1-\eta\lambda_k)^{T}\ge(\ridge-\lambda_k)/(\lambda_k+\ridge)$.

\emph{The per-entity form.} Restrict the computation of the first step to an edit that moves the row $\ve_v$ alone, with $\Phimat_v$, $\ridge_v$ and $r_v$ in place of $\Phimat$, $\mSigma$ and $r$. It gives the same identities with $\Id-\mA_v\Phimat_v=\mP^{\perp}_{N,v}$ of~\eqref{eq:spectral} in place of $\mS$. Appendix~\ref{app:release} carries out this case and proves the bound for a displacement inside the drawn span.
\end{proof}

\begin{proof}[Proof of Corollary~\ref{cor:composition}]
\emph{The identity.} In the linearized model the score of $f$ is $\dist{f}(\orig+\delta)=\dist{f}(\orig)+\feat{f}^{\top}\delta$ for every $\delta$. The anticipated score is the score at the anticipated point, and by~\eqref{eq:survival} that point is $\orig+\mS\delta$. Hence
\[
\dist{f}\big(\orig+\delta+\resp(\orig+\delta)\big)=\dist{f}(\orig)+\feat{f}^{\top}\mS\delta ,
\]
which is the pre-update score $\dist{f}(\orig)+\feat{f}^{\top}\delta'$ evaluated at $\delta'=\mS\delta$. Substituting into $\tfrac{1}{|\Fset|}\sum_f\relu(t_f-\,\cdot\,)^2$ gives $\mathcal{L}_{\mathrm{ant}}(\delta)=\mathcal{L}_{\mathrm{plain}}(\mS\delta)$ for the draw $N$. The filter depends on the draw through $\Phimat$, and the expectation over draws is $\E_N[\mathcal{L}_{\mathrm{plain}}(\mS_N\delta)]$.

\emph{The minimum.} The map $x\mapsto\relu(t-x)^2$ is non-increasing, so its value at the smaller of two scores is the larger of its two values, $\relu(t-\min\{a,b\})^2=\max\{\relu(t-a)^2,\relu(t-b)^2\}$. With $a$ the current score and $b$ the anticipated score, the composed loss of fact $f$ is $\max\{\ell_f(\delta),\ell_f(\mS\delta)\}$, and it equals $\ell_f(\mS\delta)$ when the anticipated score is the smaller of the two. The current score is $\dist{f}(\orig)+\feat{f}^{\top}\delta$, so the anticipated score is the smaller of the two if and only if $\feat{f}^{\top}\mS\delta\le\feat{f}^{\top}\delta$, that is, $\feat{f}^{\top}(\Id-\mS)\delta\ge0$. For a uniform ridge,~\eqref{eq:filter-spectral} gives $\Id-\mS=\sum_{k\le\varrho}\frac{\lambda_k}{\lambda_k+\ridge}\vu_k\vu_k^{\top}$, which is symmetric positive semidefinite. For $\delta=c\,\feat{f}$ with $c\ge0$ the condition reads $c\,\feat{f}^{\top}(\Id-\mS)\feat{f}\ge0$, which holds. For $\delta=c\,\mS\feat{f}$ the condition reads $c\,\feat{f}^{\top}(\Id-\mS)\mS\feat{f}\ge0$, and $(\Id-\mS)\mS=\sum_{k\le\varrho}\frac{\ridge\lambda_k}{(\lambda_k+\ridge)^2}\vu_k\vu_k^{\top}$ is symmetric positive semidefinite. For a general diagonal $\mSigma$ write $\tilde\mC=\mSigma^{-1/2}\mC\mSigma^{-1/2}$. Then $\Id-\mS=(\mC+\mSigma)^{-1}\mC=\mSigma^{-1/2}(\tilde\mC+\Id)^{-1}\tilde\mC\,\mSigma^{1/2}$. For $\delta=c\,\mSigma^{-1}\feat{f}$ with $c\ge0$ and $\vz=\mSigma^{-1/2}\feat{f}$, the condition reads $c\,\vz^{\top}(\tilde\mC+\Id)^{-1}\tilde\mC\,\vz\ge0$. It holds because $(\tilde\mC+\Id)^{-1}\tilde\mC$ is symmetric positive semidefinite. The matrix $\Id-\mS$ itself is symmetric exactly when $\mSigma$ commutes with $\mC$. That holds when $\ridge_u=\ridge_v$ for every pair of entities whose off-diagonal block of $\mC$ is nonzero, so the statement takes the uniform ridge.

\emph{The gradient.} The composed loss of fact $f$ is $\relu(t_f-\tilde d_f)^2$, and the stop-gradient composition of Appendix~\ref{app:composition} takes its gradient through the anticipated score on both branches of the minimum. In the linearized model that gradient with respect to $\delta$ is $-2\,\relu(t_f-\tilde d_f)\,\mS^{\top}\feat{f}$, with $\mS^{\top}=\mSigma(\mC+\mSigma)^{-1}$, the operator of part~(c) of the proof of Theorem~\ref{thm:operator}. The branch enters through the scalar $\relu(t_f-\tilde d_f)$ alone, so the minimum changes the weight of $\mS^{\top}\feat{f}$ and leaves its direction unchanged.

\emph{Nonlinear scores.} For a score that is not linear in the rows, freeze the features at the current point $\theta$, evaluate the post-update score to first order and use the signed residual. Part~(c) of the proof of Theorem~\ref{thm:operator} shows that the gradient of the anticipated score with respect to the stacked rows is then $\mS^{\top}\feat{f}$ with $\Phimat$ taken at $\theta$. With the per-entity response in place of the joint one, the part of this gradient that passes through $v$'s own response is Theorem~\ref{thm:operator}. The two agree when no drawn edge joins two involved entities (Lemma~\ref{lem:jacobi}). The identity between the two objectives therefore holds to first order at each step, with the filter rebuilt from the features at that step.
\end{proof}

\begin{proof}[Proof of Corollary~\ref{cor:thresholds}]
\emph{Part (a).} In~\eqref{eq:filter-spectral} every $\lambda_k$ is positive, so $\ridge/(\lambda_k+\ridge)\to0$ as $\ridge\to0$ and $\mS\to\Pi^{\perp}$. In the gradient-descent form, $0\le1-\eta\lambda_k<1$, so $(1-\eta\lambda_k)^{T}\to0$ as $T\to\infty$ and $(\Id-\eta\mC)^{T}\to\Pi^{\perp}$. With $\ridge=1/(\eta T)$ the two limits are the same limit. If $\Phimat\delta=0$, then $\mS\delta=\delta$ for every $\mSigma$ by the second step of the proof of Proposition~\ref{prop:release}, and $(\Id-\eta\mC)^{T}\delta=\delta$ because $\mC\delta=0$. Confining the edit to $\{\delta:\Phimat\delta=0\}=\operatorname{range}(\Pi^{\perp})$ is projection onto the null space of the update's features. Let the update class consist of the retained edges at the involved entities, with every one of them drawn. The rows of $\Phimat$ are then the features of those edges, and the constraint reads $\feat{i}^{\top}\delta=0$ over them. Limit~2 of the proof of Corollary~\ref{cor:soft} identifies this constraint with that of the retained-span methods, for a retained loss that depends on $\theta$ through the retained scores. It also records what a ranking loss with sampled negatives adds. The projector here is the joint one over the involved rows, and Limit~2 is its restriction to one row.

\emph{Part (b).} Let $\delta\in U_{\bar\lambda}$ and expand $\delta=\sum_{k:\lambda_k\le\bar\lambda}c_k\vu_k+\delta_{\perp}$ with $\delta_{\perp}=\Pi^{\perp}\delta$. By~\eqref{eq:filter-spectral} and orthonormality,
\[
\|\mS\delta\|_2^2=\sum_{k:\lambda_k\le\bar\lambda}\Big(\frac{\ridge}{\lambda_k+\ridge}\Big)^2c_k^2+\|\delta_{\perp}\|_2^2\ge\Big(\frac{\ridge}{\bar\lambda+\ridge}\Big)^2\Big(\sum_{k:\lambda_k\le\bar\lambda}c_k^2+\|\delta_{\perp}\|_2^2\Big)=\Big(\frac{\ridge}{\bar\lambda+\ridge}\Big)^2\|\delta\|_2^2 .
\]
The inequality holds because $\lambda\mapsto\ridge/(\lambda+\ridge)$ is decreasing, so each factor with $\lambda_k\le\bar\lambda$ is at least $\ridge/(\bar\lambda+\ridge)$, and because $1\ge\ridge/(\bar\lambda+\ridge)$. For $\delta=\sum_{k:\lambda_k>\bar\lambda}c_k\vu_k$ each factor is at most $\ridge/(\bar\lambda+\ridge)$, which gives $\|\mS\delta\|_2\le\frac{\ridge}{\bar\lambda+\ridge}\|\delta\|_2$. The same two bounds hold for the gradient-descent filter with $(1-\eta\bar\lambda)^{T}$ in place of $\ridge/(\bar\lambda+\ridge)$ when $\bar\lambda\le1/\eta$, because $\lambda\mapsto(1-\eta\lambda)^{T}$ is decreasing on $[0,1/\eta]$.

\emph{The reading of the slowly restored directions.} A method of this kind confines the edit to the orthogonal complement of the top $j$ eigenvectors of a second-moment matrix that it computes from a proxy. If that matrix shares its top $j$ eigenvectors with $\mC$, and $\lambda_j>\lambda_{j+1}$ when $j<\varrho$, this complement is $U_{\bar\lambda}$ for every $\bar\lambda\in[\lambda_{j+1},\lambda_j)$, and for every $\bar\lambda\in(0,\lambda_\varrho)$ when $j=\varrho$, because $U_{\bar\lambda}$ is then the orthogonal complement of $\operatorname{span}\{\vu_1,\dots,\vu_j\}$. The shared-eigenstructure assumption of \citet{xiao2026mcu} is a hypothesis of this form. The corollary is a statement about the filter of the linearized model of this paper. The published methods compute their spectra from the activations or gradients of deep networks and use their own losses, so the correspondence concerns the form of the constraint.
\end{proof}

\begin{proof}[Proof of Remark~\ref{rem:cone}]
In the linearized model the score of drawn edge $i$ is $\dist{i}(\orig+\delta)=\dist{i}(\orig)+\feat{i}^{\top}\delta$, so the one-sided residual of~\eqref{eq:drive} at $\orig+\delta$ is $r=\relu(\Phimat\delta)$ with $\relu$ applied entrywise, and the solve of~\eqref{eq:coupled} returns $-(\mC+\mSigma)^{-1}\Phimat^{\top}\relu(\Phimat\delta)$. If $\Phimat\delta\le0$ entrywise, then $r=0$, the response vanishes and the anticipated point is $\orig+\delta$. Conversely, let the response vanish and write $\va=\relu(\Phimat\delta)\ge0$. Then $\Phimat^{\top}\va=0$, hence $\va^{\top}\Phimat\delta=0$. Entry by entry, $a_i(\Phimat\delta)_i=\relu((\Phimat\delta)_i)^2=a_i^2$, so $\va^{\top}\Phimat\delta=\|\va\|_2^2$, and therefore $\va=0$, that is, $\Phimat\delta\le0$. The edits the update leaves untouched are exactly the closed convex cone $\{\delta:\Phimat\delta\le0\}$. It contains the subspace $\{\delta:\Phimat\delta=0\}$, which is the corresponding set under the signed residual by Proposition~\ref{prop:release}.
\end{proof}

\subsection{Proof of Proposition~\ref{prop:release}, per-entity form, and the separate penalty}
\label{app:release}

\begin{proof}[Proof of Proposition~\ref{prop:release}, per-entity form]
Fix a draw and a point $\theta$, and let $\delta$ move the row $\ve_v$ alone, by $\delta_v$. The residuals of the drawn edges at $v$ then change by $\Phimat_v\delta_v$, and nothing else changes. With $\mA_v$ frozen and $r_v(\theta+\delta)=r_v(\theta)+\Phimat_v\delta_v$,
\[
\Delta\ve_v(\theta+\delta)=-\mA_v\big(r_v(\theta)+\Phimat_v\delta_v\big)=\Delta\ve_v(\theta)-\mA_v\Phimat_v\delta_v,
\]
so $\ve_v+\delta_v+\Delta\ve_v(\theta+\delta)=\ve_v+\Delta\ve_v(\theta)+(\Id-\mA_v\Phimat_v)\delta_v=\ve_v+\Delta\ve_v(\theta)+\mP^{\perp}_{N,v}\delta_v$, which is the first claim. For the constant, expand $\delta_v\in\operatorname{range}(\Phimat_v^{\top})$ as $\delta_v=\sum_{k\le\varrho}c_k\vu_k$. By~\eqref{eq:spectral}, $\mP^{\perp}_{N,v}\delta_v=\sum_{k\le\varrho}\frac{\ridge_v}{\lambda_k+\ridge_v}c_k\vu_k$, and therefore
\[
\|\mP^{\perp}_{N,v}\delta_v\|_2^2=\sum_{k\le\varrho}\Big(\frac{\ridge_v}{\lambda_k+\ridge_v}\Big)^2c_k^2\le\Big(\frac{\ridge_v}{\lambda^{+}_{\min}(\Kmat_v)+\ridge_v}\Big)^2\|\delta_v\|_2^2,
\]
the factor being largest on the direction of the smallest nonzero eigenvalue. Under first-order evaluation the anticipated score moves by $\feat{f}^{(v)\top}\mP^{\perp}_{N,v}\delta_v$, of magnitude at most $\|\feat{f}^{(v)}\|_2$ times the bound. Under exact evaluation the anticipated point moves by the same bounded vector, and the score at the shifted rows is Lipschitz in each row on any region where the row norms are bounded below, so the anticipated score again moves by at most a constant times the bound.

\emph{Consequence.} Let $\hat\rho_f(\theta)=\dist{f}(\theta)-d^{\mathrm{lin}}_f(\theta)$ be the predicted score recovery under first-order evaluation. Under the displacement $\delta_v$, the pre-update score $\dist{f}$ changes by $\feat{f}^{(v)\top}\delta_v$ to first order, and the anticipated score by $\feat{f}^{(v)\top}\mP^{\perp}_{N,v}\delta_v$, so $\hat\rho_f$ changes by $\feat{f}^{(v)\top}(\Id-\mP^{\perp}_{N,v})\delta_v=\feat{f}^{(v)\top}\mA_v\Phimat_v\delta_v$, which differs from $\feat{f}^{(v)\top}\delta_v$ by $\feat{f}^{(v)\top}\mP^{\perp}_{N,v}\delta_v$, of magnitude at most $\|\feat{f}^{(v)}\|_2$ times the bound above. A displacement that undoes part of the forgetting of $f$, that is, one with $\feat{f}^{(v)\top}\delta_v<0$, therefore lowers $\hat\rho_f$ by the same amount up to that term. An objective of the form $\relu(t_f-\dist{f}(\theta))^2+\zeta\relu(\hat\rho_f)^2$ pays in its first term and is rewarded in its second. The two quantities it targets, the pre-update score and the predicted score recovery, fall by the same amount up to that term. The predicted post-update score moves by at most that term. The anticipated forgetting loss acts on $\min\{\dist{f}(\theta),\dist{f}(\theta+\resp)\}$. The displacement lowers the first argument and moves the second by at most that term, so the loss falls by at most what that term allows.
\end{proof}

\subsection{Expected force of the margin-loss update}
\label{app:force}

This subsection supports the choice of the one-sided residual~\eqref{eq:drive} in Section~\ref{sec:method:drive}. The margin loss of the update on edge $i$ is $\ell_i=\max\{0,\dist{i}-\bar n_i+\gamma\}$, with $\gamma$ the margin and $\bar n_i$ the mean score of $K$ fresh random corruptions of $i$, redrawn at every step ($K=10$ and $\gamma=8$, Appendix~\ref{app:hyper}). Let $\mu_i$ be the mean corrupted score of edge $i$ under $\orig$, which the unlearner can compute, and let $\tau_{\mathrm n}$ be the standard deviation of the $K$-negative mean $\bar n_i$ about $\mu_i$, the per-negative spread divided by $\sqrt K$. Write $\Phi_{\mathrm G}$ and $\varphi_{\mathrm G}$ for the standard normal distribution and density functions.

\begin{lemma}[Expected margin force]\label{lem:force}
Model $\bar n_i\sim\mathrm{Normal}(\mu_i,\tau_{\mathrm n}^2)$ and set $g_i=\mu_i-\gamma$. Then, at score $\dist{i}$,
\begin{gather*}
\E\Big[\frac{\partial\ell_i}{\partial\dist{i}}\Big]=\Pr(\bar n_i<\dist{i}+\gamma)=\Phi_{\mathrm G}\Big(\frac{\dist{i}-g_i}{\tau_{\mathrm n}}\Big),\\
\E[\ell_i]=\tau_{\mathrm n}\,\psi\Big(\frac{\dist{i}-g_i}{\tau_{\mathrm n}}\Big),\qquad\psi(z)=z\,\Phi_{\mathrm G}(z)+\varphi_{\mathrm G}(z),
\end{gather*}
and $\psi'=\Phi_{\mathrm G}$, so the expected loss is the Gaussian softplus of $\dist{i}-g_i$ at temperature $\tau_{\mathrm n}$ and the expected force is its sigmoidal gate.
\end{lemma}
\begin{proof}
Almost surely $\partial\ell_i/\partial\dist{i}=\mathbf 1[\dist{i}-\bar n_i+\gamma>0]$, whose expectation is $\Pr(\bar n_i<\dist{i}+\gamma)=\Phi_{\mathrm G}((\dist{i}+\gamma-\mu_i)/\tau_{\mathrm n})$. For the loss, write $\bar n_i=\mu_i+\tau_{\mathrm n}\xi$ with $\xi\sim\mathrm{Normal}(0,1)$ and $z=(\dist{i}-g_i)/\tau_{\mathrm n}$. Then $\ell_i=\tau_{\mathrm n}(z-\xi)_+$ and
\[
\E[(z-\xi)_+]=\int_{-\infty}^{z}(z-\xi)\varphi_{\mathrm G}(\xi)\,d\xi=z\,\Phi_{\mathrm G}(z)+\varphi_{\mathrm G}(z),
\]
using $\int_{-\infty}^{z}\xi\varphi_{\mathrm G}(\xi)\,d\xi=-\varphi_{\mathrm G}(z)$. Differentiating gives $\psi'(z)=\Phi_{\mathrm G}(z)+z\varphi_{\mathrm G}(z)-z\varphi_{\mathrm G}(z)$, and the last two terms cancel, so $\psi'(z)=\Phi_{\mathrm G}(z)$.
\end{proof}

\paragraph{The one-sided residual as the zero-slack bound.}
Define the slack of edge $i$ as $\varsigma_i=g_i-\dist{i}(\orig)$. An edge with $\varsigma_i>0$ is inert under $\orig$ and becomes active only once the edit has raised it by $\varsigma_i$. As $\tau_{\mathrm n}\to0$, $\tau_{\mathrm n}\psi(a/\tau_{\mathrm n})\to a_+$ for every real $a$, so the expected loss tends to $(\dist{i}-\dist{i}(\orig)-\varsigma_i)_+$ and the expected force to $\mathbf 1[\dist{i}>\dist{i}(\orig)+\varsigma_i]$. Setting every slack to zero gives the one-sided residual $r_i=\relu(\dist{i}(\theta)-\dist{i}(\orig))$ of~\eqref{eq:drive}. This choice bounds the force the update applies from above in the following sense. For every slack $\varsigma_i\ge0$ and every temperature $\tau_{\mathrm n}>0$, the increase of the expected loss over its value at $\orig$ satisfies
\[
\E[\ell_i](\dist{i})-\E[\ell_i](\dist{i}(\orig))\le\relu\big(\dist{i}-\dist{i}(\orig)\big),
\]
because $\E[\ell_i]$ is non-decreasing with slope $\Phi_{\mathrm G}\le1$. A raise of $a>0$ above the original level increases the expected loss by at most $a$, and a lowering does not increase it. The one-sided residual credits every raise in full and no lowering at all, which is the conservative reading of the force the update applies. The gate of Section~\ref{sec:method:drive}, with its measured slack and temperature, stores suppression along the features of edges with large slack. Table~\ref{tab:kau_ablation_full} reports it as the per-relation gate beside~\eqref{eq:drive}.

\paragraph{Why inert edges move.}
The gate also explains why edges that never violate their margin are restored. Let $\mathcal A_{\gamma}\subseteq P_v(N)$ be the active edges of a draw, those that violate their margin and the only ones on which the margin loss of the update exerts force. A response built from those edges alone, $\Delta\ve_v=-\Phimat_{\mathcal A_{\gamma}}^{\top}(\Phimat_{\mathcal A_{\gamma}}\Phimat_{\mathcal A_{\gamma}}^{\top}+\ridge_v\Id)^{-1}r_{\mathcal A_{\gamma}}$, changes the scores of the inert edges by $\Phimat_{\neg\mathcal A_{\gamma}}\Delta\ve_v$, which is in general nonzero when the inert features are not orthogonal to the active ones. The edges an edit $\delta_v$ raises most are those whose features are most aligned with $\delta_v$. Correcting them moves $\ve_v$ along $-\feat{i}^{(v)}$, hence back along $-\delta_v$, and every inert edge whose feature has a component along $\delta_v$ follows. This side effect is automatic in a response built from the active edges, and it is the mechanism by which an edit is undone although most of the update edges are never active. The response of~\eqref{eq:coupled} differs on this point. It keeps every drawn edge in the kernel, so it holds a drawn edge of zero residual near its current score (part~(b) of the proof of Theorem~\ref{thm:operator}), and under~\eqref{eq:drive} every raised edge is active.

\subsection{Expected loss over the update class}
\label{app:jensen}

This subsection supports the statement of Section~\ref{sec:method:draw} that the loss is averaged over draws and is never applied to an averaged response.

\begin{lemma}[Expected loss over draws]\label{lem:jensen}
Let $\ell_f(d)=\relu(t_f-d)^2$ be the forgetting loss of~\eqref{eq:unlearner}, which is convex and non-increasing in $d$. Under first-order evaluation, $d^{\mathrm{lin}}_f(\theta;N)=\dist{f}(\theta)+\sum_v\feat{f}^{(v)\top}\Delta\ve_v(\theta;N)$ is affine in the response, so $\E_N[d^{\mathrm{lin}}_f(\theta;N)]$ is the first-order score of the expected response $\E_N[\resp(\theta;N)]$, and
\[
\ell_f\big(\E_N[d^{\mathrm{lin}}_f(\theta;N)]\big)\le\E_N\big[\ell_f(d^{\mathrm{lin}}_f(\theta;N))\big]\le\max_{N}\ell_f\big(d^{\mathrm{lin}}_f(\theta;N)\big).
\]
\end{lemma}
\begin{proof}
The left inequality is Jensen's inequality for the convex function $\ell_f$, applied to the finite distribution of the draw. The right inequality bounds an average by its maximum over the finite support.
\end{proof}

The loss of the expected response therefore underestimates the expected loss, and the expected loss lies between it and the worst case over the class. This is why \eqref{eq:objective} averages the loss over draws and never applies it to an averaged response. The expected-draw row of Table~\ref{tab:kau_ablation_full} replaces the random inclusion of edges by the response $\resp_{\mW}=-\Phimat^{\top}\mW^{1/2}(\mW^{1/2}\Phimat\Phimat^{\top}\mW^{1/2}+\ridge\Id)^{-1}\mW^{1/2}r$, with $\mW$ the diagonal matrix of inclusion probabilities. This response is not $\E_N[\resp(\theta;N)]$, because the response is a nonlinear function of the draw, so the lemma gives no ordering for it. Under exact evaluation the score is not affine in the response, and the lemma gives no ordering either. The expected loss is then used as the objective over the update class without this comparison.

\subsection{Block-diagonal versus joint response}
\label{app:jacobi}

This subsection supports the statement of Section~\ref{sec:method:response} that the per-entity form counts the residual of a shared edge once at each endpoint. Stack the involved rows as $\Delta=(\Delta\ve_v)_{v\in\Vinv}\in\R^{m|\Vinv|}$ and let $\Phimat\in\R^{|N|\times m|\Vinv|}$ be the full Jacobian of the drawn edges, whose row $i$ carries the block $\feat{i}^{(h)\top}$ at its head and $\feat{i}^{(t)\top}$ at its tail, a block being absent when that endpoint is not an involved entity. With $\mSigma=\operatorname{diag}(\ridge_v\Id_m)_{v}$, the joint response solves $(\Phimat^{\top}\Phimat+\mSigma)\Delta=-\Phimat^{\top}r$, and the block-diagonal response solves, for each $v$ separately, $(\Phimat_v^{\top}\Phimat_v+\ridge_v\Id)\Delta\ve_v^{\mathrm{bd}}=-\Phimat_v^{\top}r_v$, that is, $\Delta\ve_v^{\mathrm{bd}}=-\mA_vr_v$, stacked as $\Delta^{\mathrm{bd}}$.

\begin{lemma}[Block-diagonal versus joint response]\label{lem:jacobi}
The off-diagonal block of $\Phimat^{\top}\Phimat$ between entities $v\neq u$ is $\sum_{i\ni v,u}\feat{i}^{(v)}\feat{i}^{(u)\top}$, a sum over the drawn edges whose two endpoints are $v$ and $u$. If no drawn edge has both endpoints among the involved entities, the two responses coincide for every residual. Moreover, in the limit in which every $\ridge_v\to0$, with the features of the drawn edges at each entity linearly independent (which makes the full Jacobian of full row rank), a drawn edge $i$ with both endpoints involved and $r_i\neq0$ has its residual removed twice by the block-diagonal response and once by the joint one: $\feat{i}^{\top}\Delta^{\mathrm{bd}}=-2r_i$ whereas $\feat{i}^{\top}\Delta=-r_i$.
\end{lemma}
\begin{proof}
The block structure follows from $\Phimat^{\top}\Phimat=\sum_i\feat{i}\feat{i}^{\top}$ and the support of $\feat{i}$ on the blocks of its two endpoints. If no edge couples two involved entities, every off-diagonal block is a sum over an empty set, $\Phimat^{\top}\Phimat+\mSigma$ is block diagonal, and the joint system separates into the block-diagonal systems, which proves the first claim. For the last claim, in the limit $\ridge_v\to0$ with $\Phimat_v$ of full row rank, $\Phimat_v\Delta\ve_v^{\mathrm{bd}}=-\Phimat_v\Phimat_v^{\top}\Kmat_v^{-1}r_v=-r_v$ at every entity, so each endpoint's own solve interpolates the full residual of every edge incident to it, and an edge with two involved endpoints has $\feat{i}^{\top}\Delta^{\mathrm{bd}}=\feat{i}^{(h)\top}\Delta\ve_h^{\mathrm{bd}}+\feat{i}^{(t)\top}\Delta\ve_t^{\mathrm{bd}}=-2r_i$. The joint system in the same limit interpolates all drawn edges at once, $\Phimat\Delta=-r$, because $\Phimat(\Phimat^{\top}\Phimat+\mSigma)^{-1}\Phimat^{\top}=\tilde\Kmat(\tilde\Kmat+\Id)^{-1}$ with $\tilde\Kmat=\Phimat\mSigma^{-1}\Phimat^{\top}\succeq\Phimat\Phimat^{\top}/\max_v\ridge_v$, and this matrix tends to the identity when $\Phimat$ has full row rank, so $\feat{i}^{\top}\Delta=-r_i$.
\end{proof}

The joint system is symmetric positive definite of dimension $m|\Vinv|$ and is solved by conjugate gradients \citep{hestenes1952methods} without forming a matrix. Each iteration costs one product with $\Phimat$ and one with $\Phimat^{\top}$, both linear in the number of drawn edges, plus the diagonal $\mSigma$. The block-diagonal response is a single block-Jacobi sweep of the same system.

\subsection{Within-row isotropy and the ridge}
\label{app:isotropy}

This subsection supports the statement of Section~\ref{sec:method:response} that the response leaves out Adam's per-coordinate scaling.

\begin{remark}[Within-row isotropy]\label{rem:isotropy}
For the TransE score the coordinates of the slot feature in~\eqref{eq:slotfeat} are $\feat{ij}^{(v)}=\pm\big(s_{ij}-\bar e_{vj}\,(\bar\ve_v^{\top}\vs_i)\big)/\|\ve_v\|_2$ with $\vs_i$ the sign vector of the translation error and $s_{ij}\in\{\pm1\}$, so
\[
\Big|\,|\feat{ij}^{(v)}|\,\|\ve_v\|_2-1\,\Big|\le|\bar e_{vj}|\,|\bar\ve_v^{\top}\vs_i|\qquad\text{for every coordinate }j,
\]
by expanding $(\Id-\bar\ve_v\bar\ve_v^{\top})\vs_i$ coordinatewise and using $|s_{ij}|=1$. For a row whose mass is spread over its $m$ coordinates, $|\bar e_{vj}|$ is of order $m^{-1/2}$, and $|\bar\ve_v^{\top}\vs_i|$ is of order one for a sign pattern not aligned with the row, so every coordinate of every feature at $v$ has magnitude close to $1/\|\ve_v\|_2$. Features are therefore isotropic within a row and differ across rows by the scalar $1/\|\ve_v\|_2$. Two consequences follow. Adam's per-coordinate second-moment preconditioner $\mV$ \citep{kingma2015adam} is then approximately a scalar per row, so the Adam-preconditioned kernel $\Phimat\mV^{-1}\Phimat^{\top}$ differs from the Euclidean kernel only by per-row scalars, the sign-gradient kernel $\langle\sign\feat{i},\sign\feat{j}\rangle$ of \citet{malladi2023kernel} is, for this score, approximately the Euclidean kernel up to the same scalars, and the metric orthogonality of \citet{zhou2025geometric} is Euclidean orthogonality up to row scalars. Those scalars can be absorbed in a per-entity ridge, which is the second consequence, and it is stated as a hypothesis. Dense Adam over the embedding table keeps a second-moment average per coordinate, so for a row that receives gradient in a fraction $p_v$ of the steps the average is about $p_v$ times the squared active gradient, each active step moves the row by about $\eta/\sqrt{p_v}$, and the movement over $T$ steps is about $\eta T\sqrt{p_v}$, against $\eta Tp_v$ under stochastic gradient descent. Reading this as an effective per-row learning rate $\eta_v\propto\sqrt{p_v}$ and inserting it in Remark~\ref{rem:ridge} gives
\[
\begin{gathered}
\ridge_v=\ridge_0\,(p_v/p_{\mathrm{med}})^{-\kappa},\\
\kappa=\tfrac12\ \text{(Adam)},\quad\kappa=1\ \text{(stochastic gradient descent)},\quad\kappa=0\ \text{(degree-independent)},
\end{gathered}
\]
with $p_v$ proportional to the degree of $v$ in the update class and $p_{\mathrm{med}}$ its median. Under this hypothesis rare entities are fitted less completely by the update than entities of high degree, but only as $\sqrt{p_v}$. Every experiment in this paper uses the degree-independent choice $\kappa=0$, with $\ridge_v$ scaled to the feature norms at $v$ only (Section~\ref{sec:method:coupled}).
\end{remark}

\subsection{Piecewise-linear scores and exact evaluation}
\label{app:pwlnote}

The anticipated score is evaluated at the shifted rows for a reason that holds on every backbone. The score is nonlinear in the rows, and the displacement the update produces is not small on the scale of the coordinate differences from which the score is built. On an $L_1$ score the gap between the first-order evaluation and the exact one has an elementary closed form.

\begin{lemma}[Piecewise-linear scores]\label{lem:pwl}
For real $x$ and $D$, with the convention $\sign0=1$,
\begin{equation}\label{eq:pwl-identity}
|x+D|-|x|=\sign(x)\,D+2\big(|D|-|x|\big)_+\,\mathbf 1\big[\sign D\neq\sign x\big].
\end{equation}
The second term is non-negative, vanishes whenever $|D|\le|x|$, and equals $2(|D|-|x|)$, linear in $|D|$ with slope two, once $|D|>|x|$ and $D$ opposes $x$.
\end{lemma}
\begin{proof}
If $\sign D=\sign x$, then $|x+D|=|x|+|D|$ and $\sign(x)D=|D|$, so both sides equal $|D|$ and the indicator is zero. If $\sign D\neq\sign x$ and $|D|\le|x|$, then $x+D$ has the sign of $x$, $|x+D|=|x|-|D|=|x|+\sign(x)D$, and the second term vanishes because $(|D|-|x|)_+=0$. If $\sign D\neq\sign x$ and $|D|>|x|$, then $x+D$ has the sign of $D$, so the left-hand side is $(|D|-|x|)-|x|=|D|-2|x|$, and the right-hand side is $\sign(x)D+2(|D|-|x|)=-|D|+2|D|-2|x|=|D|-2|x|$. The case $x=0$ falls under the convention. With $\sign0=1$, $D>0$ is the first case and $D<0$ the third, and both give $|D|$. The listed properties of the second term are read off the three cases.
\end{proof}

For a triple with translation error $\vx=\bar\ve_h+\bar\vw_r-\bar\ve_t$, so that $\dist{x}=\|\vx\|_1$, and a displacement $\bm D$ of that vector induced by the response, summing~\eqref{eq:pwl-identity} over coordinates gives $\dist{x}(\theta+\resp)-\dist{x}(\theta)=\vs^{\top}\bm D+\varkappa(\bm D)$ with $\vs=\sign(\vx)$ and $\varkappa\ge0$. The first term is, up to the normalization's second-order term, the first-order evaluation $\feat{x}^{\top}\resp$ that every operator acting on tangent features controls. The second is non-negative, vanishes on the box $|D_j|\le|x_j|$, is not linear in $\bm D$, and depends on the magnitudes $|x_j|$ as well as on the signs, so no operator built from the tangent feature alone models it. The second term matters when $|D_j|$ exceeds $|x_j|$ in some coordinates, and Appendix~\ref{app:predictor} measures the consequence. Evaluating the score at the shifted rows captures both terms at the cost of one call to the score function, which is why \KAU{} does so.

\section{Method details}\label{app:method}

Section~\ref{sec:method} defines the objective and names its parts. This appendix builds them. The first six subsections give the update draw, the targets, the composition rule, the retraction rule, the score-preservation term and the selection rule. The seventh lists one run as Algorithm~\ref{alg:kau}. The eighth gives the three backbones and the way a slot feature is taken from each. The last three give every hyperparameter, the cost of a run, and three constructions the derivation admits that we do not use. Everything here is computed from the original model $\orig$, the forget set $\Fset$ and the retained edges $\Rset$. No update edge, no held-out edge and no retrained model enters at run time.

\subsection{The update draw and its stand-ins}\label{app:draw}

The unlearner never learns which edges the update will use, so it models that edge set as a random draw with two channels. The known channel is made of the unlearner's own retained edges. At an involved entity $v$, each retained edge incident to $v$ enters a draw independently with probability $\rho_{\mathrm{kn}}$. Whatever retained edges the user's update includes at $v$ are a subset of these, so this channel contains genuine members of the update class.

The unseen channel stands in for true edges that were in no training set. The unlearner has never seen such an edge, so it manufactures stand-ins. It keeps a pool of $32$ stand-in triples at every involved entity, and each stand-in is built in two steps. A relation is drawn from the empirical relation profile of $v$ over its retained edges. A partner entity is then drawn from the entities that occur with that relation in the retained set. Four candidates are built this way and the one the original model finds most plausible is kept, which is the candidate with the lowest score. A candidate that already occurs in $\Fset$ or in $\Rset$ is rejected, so a stand-in is neither a forgotten nor a retained fact. The pool is refreshed every five epochs. Each stand-in then enters a draw with probability $\min\{1,\rho_{\mathrm{un}}\deg(v)/32\}$, which puts the expected number of unseen edges at $v$ at $\min\{32,\rho_{\mathrm{un}}\deg(v)\}$ and lets that number grow with the entity's degree up to the pool size.

Every draw obeys two constraints. At most $k_{\max}=96$ edges enter per entity. The cap lies below the largest rank the features at $v$ can reach, which is $m-1$ on TransE and TransH, whose normalization removes the radial direction (Appendix~\ref{app:backbones}), and $m$ on RotatE. Every draw therefore leaves directions at $v$ off its span, where an edit survives the response entirely (Section~\ref{sec:method:response}, Appendix~\ref{app:cor}). The second constraint is that a fresh draw is taken at every forgetting step, so the objective is the expected loss over the class. Over a run each known edge is drawn several times.

The residual of \eqref{eq:drive} is evaluated on the drawn edges at the current parameters. For a stand-in it uses that triple's own score under the original model. A forgotten fact carries no residual, and its loss is measured against the target of Appendix~\ref{app:targets}.

\subsection{Targets}\label{app:targets}

The target $t_f$ is the score at which the forgetting loss stops pushing, and it is built from the original model alone. For every forgotten fact we replace one endpoint at a time by a uniformly drawn entity and score $128$ corrupted copies under $\orig$, $64$ with the head replaced and $64$ with the tail replaced. Let $q_f$ be the $5\%$ quantile of those scores, the level at which the original model holds a fact it does not believe. The target of \eqref{eq:target} is $t_f=\dist{f}(\orig)+s\,\relu(q_f-\dist{f}(\orig))$, with the default $s=0.15$ and the value selected per backbone in Table~\ref{tab:hyper}, computed once before the run and never updated. A fact the original model already scores near that tail is asked to move very little, and a well-believed fact is asked to move a fixed fraction of the way toward it. Because the loss is one-sided and exerts no force past $t_f$, the operating point of the release is set by these targets rather than by the epoch at which the run happens to end. The score-preservation term of Appendix~\ref{app:anchor} and the two ancillaries of Appendices~\ref{app:retraction} and~\ref{app:selection} leave those targets alone and act on the rows instead.

\subsection{Pessimistic composition}\label{app:composition}

The loss of \eqref{eq:objective} acts on $\tilde d_f=\min\{\dist{f}(\theta),\dist{f}(\theta+\resp)\}$, so a fact counts as forgotten only if it is forgotten both now and after the predicted update. The gradient is routed through the anticipated path in either case. The composed score is computed as
\begin{equation*}
\tilde d_f=\dist{f}(\theta+\resp)-\operatorname{sg}\big[\relu\big(\dist{f}(\theta+\resp)-\dist{f}(\theta)\big)\big],
\end{equation*}
with $\operatorname{sg}$ the stop-gradient, which has the value of the minimum and the gradient of the anticipated score. The optimizer step therefore always targets the post-update position.
Because of the minimum, a predicted rise of the fact's score cannot lower the loss, which closes the shortcut of Section~\ref{sec:method:drive}.

\subsection{Retraction}\label{app:retraction}

After every optimizer step, each edited entity row is rescaled to the length it had under the original model, $\ve_v\leftarrow\|\ve^{\circ}_v\|_2\,\ve_v/\|\ve_v\|_2$. The reason is that the slot feature at $v$ scales as $1/\|\ve_v\|_2$. A model whose edited rows have grown therefore presents a smaller gradient to the update after release, on exactly the rows that were edited.
Its resistance is then a matter of scale rather than of where the edit was stored. That scale is free to grow. Under a loss that does not see the length of a row the gradient is orthogonal to the row, so the length rises on its own \citep{arora2019theoretical,heo2021adamp}. For TransE and TransH the length is a gauge, because the score normalizes the rows inside its own forward pass, and retraction only fixes that gauge. For RotatE the length enters the score, so retraction is a genuine constraint on the edit. \KAU{} retracts on all three backbones, and Appendix~\ref{app:twobytwo} reads the row without retraction beside the full method and shows that the result does not rest on norm inflation.

\subsection{Score preservation}\label{app:anchor}

The retained pass carries the ordinary margin-ranking loss on retained edges and the second term of the objective of \eqref{eq:total} in Section~\ref{sec:method}, $\beta\,\mean_i(\dist{i}(\theta)-\dist{i}(\orig))^2$, over retained edges incident to the involved entities and resampled at every step. Proposition~\ref{prop:anchor} identifies this term with the squared residual of the known channel. It is the first of the two controls of Section~\ref{sec:method:controls}. Table~\ref{tab:kau_ablation_recovery_anc03} switches it off and Table~\ref{tab:kau_ablation_full} replaces it by a relational anchor, rows that Appendix~\ref{app:twobytwo} reads beside the retraction row. A retained edge that the edit has pushed above its original score is an edge the update will pull back, and pulling it back drags the forgotten facts with it.

\subsection{Selection}\label{app:selection}

Once per epoch the run evaluates its current checkpoint and scores it by $0.7\,(1-\mathrm{MRR}_{\Fset})+0.3\,\mathrm{MRR}_R$ on the development subsets of the forget set and of the retained set that the unlearner holds, disjoint from the evaluation subsets (Appendix~\ref{app:appendix_traningl}). The released model is the maximizer over the epochs of the run. This is \GraphDPO{}'s own rule applied unchanged, and it consults neither the retrained model nor any edge of the update. Its weight $0.7$ is the value that Appendix~\ref{app:unlearning} selects for the checkpoint-rule weight. It chooses the checkpoint within a run. Hyperparameters are chosen before the run, by the criterion that every method shares (Appendix~\ref{app:unlearning}).

\subsection{The full procedure}\label{app:algorithm}

Algorithm~\ref{alg:kau} lists one run of \KAU{}.
During the run only the rows of $\Vinv$ are trainable, and the relation rows stay at their values in $\orig$.
Each epoch makes one retained pass over batches $\mathcal{B}$ of retained edges at $\Vinv$, then $J$ forgetting steps.
The margin-ranking loss on $\mathcal{B}$ is $\ell_{\mathrm{MR}}(\mathcal{B})$, and $\ell_R(\theta;S)$ is the mean in \eqref{eq:total} over a sample $S$ redrawn at every step.
The call $\textsc{Adam}(\theta,L)$ takes one step of the pass's own Adam optimizer on $L$.
The call $\textsc{Retract}$ rescales each row of $\Vinv$ to its norm under $\orig$ (Appendix~\ref{app:retraction}).
The stop-gradient $\operatorname{sg}$ keeps the features out of the gradient, and $\textsc{CG}$ solves \eqref{eq:coupled} by conjugate gradients without forming the matrix.
The checkpoint score $\textsc{Select}$ is the rule of Appendix~\ref{app:selection}, and the stand-in pools of Appendix~\ref{app:draw} are rebuilt every five epochs.
The epoch count $E$, the forgetting learning rate and the target fraction $s$ are selected per backbone by the rule that every method shares (Appendix~\ref{app:unlearning}).

\begin{algorithm}[h]
\renewcommand{\algorithmiccomment}[1]{\hfill$\triangleright$~#1}
\caption{\KAU{}, one unlearning run}
\label{alg:kau}
\begin{algorithmic}[1]
\REQUIRE original model $\orig$, forget set $\Fset$, retained edges $\Rset$, modelled class $\hat{\mathcal{N}}$, ridge scale $\ridge_{\mathrm{rel}}$, weight $\beta$, target fraction $s$, epochs $E$, forgetting steps $J$
\ENSURE released model $\unl$
\STATE $\theta\leftarrow\orig$
\STATE $t_f\leftarrow\dist{f}(\orig)+s\,\relu\big(q_f-\dist{f}(\orig)\big)$ for $f\in\Fset$ \COMMENT{(\plaineqref{eq:target})}
\FOR{$e=1$ \TO $E$}
  \FOR{each batch $\mathcal{B}$ of retained edges at $\Vinv$}
    \STATE $S\leftarrow$ fresh sample of retained edges at $\Vinv$
    \STATE $\theta\leftarrow\textsc{Retract}\big(\textsc{Adam}(\theta,\ \ell_{\mathrm{MR}}(\mathcal{B})+\beta\,\ell_R(\theta;S))\big)$ \COMMENT{(\plaineqref{eq:total})}
  \ENDFOR
  \FOR{$j=1$ \TO $J$}
    \STATE $N\sim\hat{\mathcal{N}}$ \COMMENT{\S\ref{sec:method:draw}}
    \STATE $\Phimat\leftarrow\operatorname{sg}\big[\partial\dist{N}(\theta)/\partial\ve_{\Vinv}\big]$, \quad $r\leftarrow r_N(\theta)$ \COMMENT{(\plaineqref{eq:drive})}
    \STATE $\mSigma\leftarrow\operatorname{diag}(\ridge_v\Id)$, \ $\ridge_v=\ridge_{\mathrm{rel}}\,\mean_i\|\feat{i}^{(v)}\|_2^2$
    \STATE $\Delta\leftarrow\textsc{CG}\big(\Phimat^{\top}\Phimat+\mSigma,\ -\Phimat^{\top}r\big)$ \COMMENT{(\plaineqref{eq:coupled}), differentiable in $r$}
    \STATE $\tilde d_f\leftarrow\dist{f}(\theta+\Delta)-\operatorname{sg}\big[\relu\big(\dist{f}(\theta+\Delta)-\dist{f}(\theta)\big)\big]$ for $f\in\Fset$ \COMMENT{App.~\ref{app:composition}}
    \STATE $\theta\leftarrow\textsc{Retract}\big(\textsc{Adam}\big(\theta,\ \tfrac{1}{|\Fset|}\textstyle\sum_{f}\relu(t_f-\tilde d_f)^2\big)\big)$ \COMMENT{(\plaineqref{eq:objective})}
  \ENDFOR
  \STATE $\theta_e\leftarrow\theta$, \quad $c_e\leftarrow\textsc{Select}(\theta)$
\ENDFOR
\RETURN $\unl\leftarrow\theta_{e^\star}$ with $e^\star=\arg\max_e c_e$
\end{algorithmic}
\end{algorithm}

\subsection{Backbones and where the features are taken}\label{app:backbones}

The method needs a score and its gradients with respect to the two endpoint rows, so it applies to any backbone that keeps one entity table and one relation table and is differentiable almost everywhere in those rows. We report three, all with $200$ latent coordinates per entity. TransE stores one real row of $m=200$ coordinates per entity and translates normalized rows \citep{bordes2013translating}. RotatE packs $200$ complex coordinates into a real row of $m=400$ and keeps one phase per coordinate in the relation row \citep{sun2019rotate}. Its rows are used without normalization, and its trained distances are about six times TransE's. TransH keeps entity rows of $m=200$ and normalizes them as TransE does, packs the translation and the hyperplane normal into a relation row of $400$ coordinates, and projects both endpoints onto the relation's hyperplane before translating \citep{wang2014transh}. Section~\ref{sec:setup} gives the TransE score, and the other two follow the cited definitions.

The slot feature $\feat{x}^{(v)}=\partial \dist{x}/\partial\ve_v$ is taken with respect to the raw parameter row, not with respect to whatever the model exposes as an embedding. Any transform the score applies to a row therefore sits inside the differentiated graph and reaches the feature through its own Jacobian. For a normalized row that Jacobian contributes a factor $(\Id-\bar\ve\bar\ve^{\top})/\|\ve\|_2$, which removes the radial direction the score cannot see. For a packed complex row it contributes the map from the two real coordinates of a block to that block's modulus. The response is then a displacement of the same raw rows the optimizer updates, and the anticipated score is obtained by adding the displacement to those rows and calling the score function again, with no correction term anywhere. Porting the method to a new backbone requires the score function and nothing else.

\subsection{Hyperparameters}\label{app:hyper}

Table~\ref{tab:hyper} lists the settings of the unlearner, of the update and of the models. The unlearner's values are the same on the three backbones except for the forgetting learning rate, the epoch count and the target fraction $s$, which are selected per backbone (Appendix~\ref{app:unlearning}). The retained-span projection of Table~\ref{tab:ablation_recovery_depth} shares this table and differs in two places. It applies the plain forgetting loss to the pre-update score, and it projects that loss's gradient off the span of the incident retained edges through a ridge-damped projector, with \KAU{}'s ancillaries.

\begin{table}[htbp]
\centering\small
\caption{Hyperparameters on FB15k-237 (FB). The CoDEx-L settings are selected by the same rule (Appendix~\ref{app:unlearning}) and listed in Appendix~\ref{app:budget}.}
\label{tab:hyper}
\begin{tabular}{p{0.27\linewidth}p{0.65\linewidth}}
\toprule
\multicolumn{2}{l}{\emph{Models}} \\
\midrule
Backbones & TransE, RotatE, TransH at $200$ latent coordinates, Appendix~\ref{app:backbones} \\
Training & margin ranking with Adam \citep{kingma2015adam}, loss $[\dist{x}(\theta)-\bar n_x+\gamma]_+$ with $\bar n_x$ the mean score of $K=10$ random corruptions of $x$ and margin $\gamma=8$ (Section~\ref{sec:setup}) \\
Original and retrained models & one training run each per graph and backbone with that recipe, the original model on $\mathcal{D}_{\mathrm{train}}\setminus B$, with $B$ the never-trained set of Section~\ref{sec:exp}, and the retrained model on $\Rset$ alone, never visible to the unlearner. Table~\ref{tab:oracle_seeds} adds two further retrained models on FB15k-237, at retraining seeds 2 and 3 \\
\midrule
\multicolumn{2}{l}{\emph{Update class assumed by the unlearner}} \\
\midrule
Known channel & each retained edge at $v$ with probability $\rho_{\mathrm{kn}}=0.2$ \\
Unseen channel & $32$ stand-ins per entity, best of four by plausibility under $\orig$, refreshed every $5$ epochs, each entering with probability $\min\{1,\rho_{\mathrm{un}}\deg(v)/32\}$, $\rho_{\mathrm{un}}=0.05$ \\
Cap and draw & at most $k_{\max}=96$ edges per entity, one fresh draw per forgetting step \\
\midrule
\multicolumn{2}{l}{\emph{Response, evaluation, composition}} \\
\midrule
Residual & $r_i=\relu(\dist{i}(\theta)-\dist{i}(\orig))$ on every drawn edge \\
Ridge & $\ridge_v=\ridge_{\mathrm{rel}}\cdot\mean_i\|\feat{i}^{(v)}\|_2^2$ with $\ridge_{\mathrm{rel}}=10^{-3}$ \\
Solve & $(\Phimat^\top\Phimat+\mSigma)\Delta=-\Phimat^\top r$ over all involved rows by conjugate gradients \citep{hestenes1952methods}, features recomputed at the current $\theta$ at every step and carrying no gradient \\
Evaluation & $\dist{f}$ at the shifted rows through the model's own score function \\
Composition & $\tilde d_f=\min\{\dist{f}(\theta),\dist{f}(\theta+\resp)\}$, the gradient routed through the anticipated path \\
Targets & $t_f=\dist{f}(\orig)+s\,\relu(q_f-\dist{f}(\orig))$, $s=0.15$ by default, selected as $0.15$, $0.4$ and $0.25$ on TransE, TransH and RotatE, $q_f$ the $5\%$ quantile of $\orig$'s scores over $128$ corrupted copies \\
\midrule
\multicolumn{2}{l}{\emph{Optimization and ancillaries}} \\
\midrule
Trainable rows & the entity rows of $\Vinv$, $11{,}412$ of $14{,}541$ on FB, relations frozen, released model fully trainable \\
Forgetting pass & $2$ steps per epoch, at Adam learning rate $3\times10^{-5}$ on TransE and TransH and $3\times10^{-4}$ on RotatE, following its score scale, for $20$, $10$ and $10$ epochs on TransE, TransH and RotatE (Appendix~\ref{app:unlearning}) \\
Retained pass & margin loss and score preservation over retained edges incident to $\Vinv$, Adam learning rate $10^{-4}$, once per epoch. The score-preservation term has weight $\beta=1$, the anchor weight of Appendix~\ref{app:unlearning}, and $0.3$ in the ablation tables of Section~\ref{sec:exp}. It is taken over $4096$ of those edges, resampled at every step \\
Retraction & every edited row rescaled to its original length after every step \\
Selection & $0.7\,(1-\mathrm{MRR}_{\Fset})+0.3\,\mathrm{MRR}_R$ on the development subsets the unlearner holds, once per epoch over the run \\
\midrule
\multicolumn{2}{l}{\emph{Update, and the published baselines}} \\
\midrule
Update & every parameter trainable, margin loss, Adam, on retained edges and never-trained edges, over the $48$ update configurations of Section~\ref{sec:exp}, shared by every method and by the oracle \\
\GraphDPO{} \citep{liu2026graphdpo} & the authors' implementation, under the same hyperparameter-selection criterion as \KAU{} (Appendix~\ref{app:unlearning}). It is tuned over its own ranges of Appendix~\ref{app:unlearning} and updates both the entity table and the relation table. Its feasible set therefore contains that of \KAU{}, which edits only the involved entity rows under the restriction defined in Section~\ref{sec:setup} \\
ZeroFisher \citep{xiao2025zerofisher} & the authors' implementation, tuned over its perturbation scale, update scale, number of iterations and, on RotatE, damping, under the same criterion. It edits the rows of the local neighbours of the forgotten edges (Section~\ref{sec:related}) \\
\bottomrule
\end{tabular}
\end{table}

\subsection{Cost}\label{app:cost}

This subsection gives the computational cost of \KAU{} and compares its time to release with that of the baselines.
A forgetting step draws about ten edges per involved entity, recomputes their features in one backward pass, solves for the response over all involved rows at once, and scores every forgotten fact at the shifted rows.
Each conjugate-gradient iteration costs one product with $\Phimat$ and one with its transpose, both linear in the number of drawn edges, so no matrix is ever formed.
Start-up adds an incidence list per involved entity, the first stand-in pool, and one pass of $128$ corrupted scores per forgotten fact for the targets.
Every run reported in this paper uses one NVIDIA RTX PRO 5000 Blackwell GPU.

Table~\ref{tab:cost} gives the time to release one unlearned model on CoDEx-L with TransE, over unlearning seeds 21 to 27, with four CPU cores.
\KAU{} releases $5.4$ times faster than \GraphDPO{} and $18$ times faster than retraining.

\begin{table}[htbp]
\centering\small
\caption{Time to release one unlearned model on CoDEx-L with TransE, median over seven unlearning seeds with the range in brackets.}
\label{tab:cost}
\begin{tabular}{lc}
\toprule
Method & Time to release (s) \\
\midrule
Retraining (exact) & 240.4 [240.2, 240.5] \\
\GraphDPO{} & 72.8 [72.7, 72.8] \\
\KAU{} & 13.4 [13.4, 13.4] \\
\bottomrule
\end{tabular}
\end{table}

\subsection{Constructions the derivation admits and we do not use}\label{app:variants}

Three further constructions follow from the same derivation and are not part of the method. A second Gauss--Newton round re-linearizes the response at the shifted rows on the same draw and adds the correction to the first, which models the drift of the features during the update, and it improved no reported quantity. Letting the features carry gradient opens a channel in which the unlearner reshapes the update's features instead of the residual they act on, and it was worse, so the features carry no gradient within a step. A penalty that holds the stand-ins at their original scores leaves the unseen channel free of residual at release, and it costs utility without moving the leakage. A fourth item is a hypothesis. Adam's per-coordinate scaling suggests a ridge that falls with the number of update edges at an entity, as the inverse square root of its share of the steps (Remark~\ref{rem:isotropy} in Appendix~\ref{app:isotropy}), and every experiment reported here uses the degree-independent ridge.

\section{Experimental details}\label{app:protocol}

Section~\ref{sec:setup} defines the update and the leakage, and Section~\ref{sec:exp} states the protocol in outline. This appendix gives the graphs and their splits, the edge set the update runs on, the budget each method receives, the seeds and tests behind every table, and the CoDEx-L results in full.

\subsection{Training details}
\label{app:appendix_traningl}

On FB15k-237 the training graph has 272,115 triples. The never-trained set $B$ holds 6,000 of them (2.2\%) and the forget set 27,211 (10\%), both sampled uniformly at random with a fixed seed; the retained set holds the remaining 238,904. On CoDEx-L the training graph has 551,193 triples, $B$ holds 12,000 (2.2\%), the forget set 55,119 (10\%) and the retained set 484,074. For each dataset and backbone, the original model $\orig$ is trained on $\mathcal{D}_{\mathrm{train}}\setminus B$, and the gold model $\gold$ is trained from scratch on $\Rset$, which also excludes the forget set. From each forget set a probe of 1,000 triples is drawn once and sealed; it serves only the relearning evaluation of Appendix~\ref{app:exp_protocol}.

\paragraph{Pre-training.}
Original and gold models use the \GraphDPO{} reference implementation of each backbone, with embedding dimension 200, a margin ranking loss with margin 8, ten negatives per positive, batch size 1,024, and Adam with learning rate $10^{-4}$ for 100 epochs. For each dataset and backbone, the two models use identical architecture, objective, optimisation settings and seed, differing only in their training triples. All unlearners share the original checkpoint and gold reference. Both models' training curves flatten well before the schedule ends (Figure~\ref{fig:pretrain_loss}). Every relearning quantity of Appendix~\ref{app:exp_protocol} is a difference against this gold model, so each is read relative to the reference rather than as an absolute level.

\begin{figure}[ht]
\centering
\includegraphics[width=0.6\linewidth]{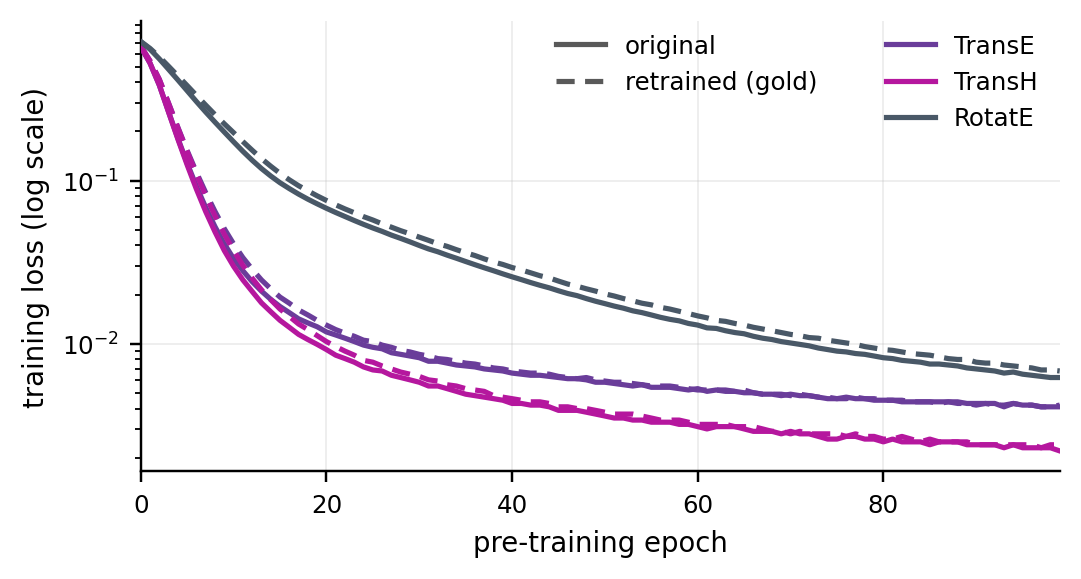}
\caption{Pre-training loss on FB15k-237, at every epoch of the schedule. Solid lines are the original model and dashed lines the gold model. The pair of each backbone shares the recipe above and its seed, and differs only in the training triples. Both curves flatten well before the schedule ends, and every quantity we report is a difference between the two.}
\label{fig:pretrain_loss}
\end{figure}

\paragraph{Evaluation.}
All MRR values are filtered and averaged over head and tail prediction. Forget-set MRR is computed on a held-out subset of the deleted triples that is never used for selecting hyperparameters or update configurations, retain-set MRR on a fixed probe of retained training triples, and test MRR on the standard test split.

\subsection{Unlearning stage}
\label{app:unlearning}

\paragraph{Metrics.}
We evaluate forgetting and utility immediately after unlearning. \textbf{Forget MRR} ($M_F$) is computed on the evaluation subset of $\Fset$ and measures how much of the deleted knowledge remains. \textbf{Retain MRR} ($M_R$) is computed on the evaluation subset of $\Rset$ and measures preservation of knowledge that was not deleted. \textbf{Test MRR} ($M_T$) is computed on the test split and measures general link-prediction utility. The original and gold models provide reference values. An unlearned model should match the gold model on all three: a forget MRR above the gold model's indicates residual knowledge, and one below it indicates over-forgetting, i.e., the deleted triples are ranked worse than by a model that never observed them.

\paragraph{Hyperparameter selection.}
\KAU{} and each baseline are tuned separately for every dataset--backbone setting, with the same criterion for all methods. Each configuration is scored by the larger of two shortfalls, one in forgetting and one in retained performance, each measured on the corresponding development subset, and the configuration with the smallest score is evaluated on disjoint held-out subsets, which produce all reported values. Selection and reporting use the same seven unlearning seeds.

For \GraphDPO{} the search covers the learning rate $\{10^{-4},3\times10^{-4},10^{-3},3\times10^{-3},10^{-2}\}$, the number of epochs $\{1,2,3,6,12,24,48\}$, the distillation weight $\{3\times10^{-4},10^{-3},3\times10^{-3}\}$, the preference weight $\{0.03,0.1,0.3\}$ and temperature $\{1,3\}$, the margin $\{4,8\}$, and the policy and side of negative sampling, followed by a local refinement of the learning rate and distillation weight around the best setting. For ZeroFisher the search covers the perturbation scale $\epsilon\in\{10^{-5},10^{-4},10^{-3}\}$, the update scale $\{0.05,0.1,0.25,0.4,0.5,1,10,35\}$, the number of iterations from $3$ to $1{,}550$, and, on RotatE, damping in $\{0,-0.01,-0.02,-0.03,-0.04\}$, the ranges recommended by its authors. For \KAU{} the forgetting learning rate $\{10^{-5},3\times10^{-5},10^{-4},3\times10^{-4},10^{-3}\}$, the number of epochs $\{2,3,5,10,20\}$ and the target fraction $s$ of \eqref{eq:target}, $s\in\{0.005,0.02,0.05,0.15,0.25,0.4,0.6\}$, are searched jointly, and the number of forgetting steps per epoch $\{1,2,4\}$, the cap $k_{\max}\in\{48,96,160\}$ on drawn edges per entity, the inclusion probabilities of the two channels, $\rho_{\mathrm{kn}}\in\{0.05,0.2,0.5,1\}$ and $\rho_{\mathrm{un}}\in\{0.01,0.05,0.3\}$, the relative ridge $\ridge_{\mathrm{rel}}\in\{10^{-4},10^{-3},10^{-2},10^{-1}\}$, the score-preservation weight $\beta\in\{0.3,1,3\}$ of \eqref{eq:total}, called the anchor weight, and the weight of the checkpoint rule of Appendix~\ref{app:selection}, $\{0.3,0.5,0.7,0.95\}$, are varied one at a time around a reference setting. The selected \KAU{} settings on FB15k-237 are, for TransE, TransH and RotatE respectively, epochs 20, 10 and 10, forgetting learning rate $3\times10^{-5}$, $3\times10^{-5}$ and $3\times10^{-4}$, and $s=0.15$, $0.4$ and $0.25$, with two forgetting steps per epoch, $k_{\max}=96$, $\rho_{\mathrm{kn}}=0.2$, $\rho_{\mathrm{un}}=0.05$, $\ridge_{\mathrm{rel}}=10^{-3}$, $\beta=1$ and a checkpoint-rule weight of $0.7$ shared across backbones.

\begin{table}[htbp]
\centering
\caption{Unlearning results on FB15k-237 with 10\% triple deletion. $M_F$, $M_R$ and $M_T$ denote filtered MRR on the forget, retain and test evaluation sets. Original and Gold are the shared reference checkpoints. Method rows report the mean over seven unlearning seeds; standard deviations are in Table~\ref{tab:main_std}. Bold marks the method closest to Gold on $M_F$ and the highest on $M_R$ and $M_T$.}
\label{tab:main}
\begin{tabular*}{\linewidth}{@{\extracolsep{\fill}}lrrrrrrrrr@{}}
\toprule
& \multicolumn{3}{c}{TransE} & \multicolumn{3}{c}{TransH} & \multicolumn{3}{c}{RotatE} \\
\cmidrule(lr){2-4}\cmidrule(lr){5-7}\cmidrule(lr){8-10}
Method & $M_F$ & $M_R\!\uparrow$ & $M_T\!\uparrow$ & $M_F$ & $M_R\!\uparrow$ & $M_T\!\uparrow$ & $M_F$ & $M_R\!\uparrow$ & $M_T\!\uparrow$ \\
\midrule
Original   & 0.2733 & 0.2800 & 0.2137 & 0.2550 & 0.2596 & 0.2038 & 0.3075 & 0.3176 & 0.2214 \\
Gold       & 0.2211 & 0.2932 & 0.2143 & 0.2097 & 0.2739 & 0.2051 & 0.2127 & 0.3301 & 0.2203 \\
\midrule
\GraphDPO{} & 0.2301 & 0.2628 & 0.2041 & 0.2121 & 0.2401 & 0.1927 & 0.2298 & 0.2531 & 0.1808 \\
ZeroFisher  & \textbf{0.2249} & 0.2322 & 0.1857 & 0.2123 & 0.2164 & 0.1818 & 0.2180 & 0.2261 & 0.1693 \\
\KAU{}      & 0.2272 & \textbf{0.2839} & \textbf{0.2137} & \textbf{0.2105} & \textbf{0.2614} & \textbf{0.2041} & \textbf{0.2112} & \textbf{0.3254} & \textbf{0.2213} \\
\bottomrule
\end{tabular*}
\end{table}

\begin{table}[htbp]
\centering
\caption{Standard deviation over the seven unlearning seeds for every method cell of Table~\ref{tab:main}. Original and Gold are single checkpoints and have no seed variation. This table prints the population standard deviation over the seven seeds (divisor $n$). Table~\ref{tab:relearn_std} and the tests of Table~\ref{tab:relearn_tests} use the sample standard deviation (divisor $n-1$).}
\label{tab:main_std}
\begin{tabular*}{\linewidth}{@{\extracolsep{\fill}}lrrrrrrrrr@{}}
\toprule
& \multicolumn{3}{c}{TransE} & \multicolumn{3}{c}{TransH} & \multicolumn{3}{c}{RotatE} \\
\cmidrule(lr){2-4}\cmidrule(lr){5-7}\cmidrule(lr){8-10}
Method & $M_F$ & $M_R$ & $M_T$ & $M_F$ & $M_R$ & $M_T$ & $M_F$ & $M_R$ & $M_T$ \\
\midrule
\GraphDPO{} & 0.0009 & 0.0012 & 0.0008 & 0.0008 & 0.0013 & 0.0005 & 0.0011 & 0.0013 & 0.0013 \\
ZeroFisher  & 0.0327 & 0.0320 & 0.0181 & 0.0340 & 0.0346 & 0.0212 & 0.0440 & 0.0445 & 0.0306 \\
\KAU{}      & 0.0006 & 0.0008 & 0.0004 & 0.0004 & 0.0005 & 0.0005 & 0.0004 & 0.0007 & 0.0002 \\
\bottomrule
\end{tabular*}
\end{table}

Seed variation is below $0.002$ MRR for \KAU{} and \GraphDPO{} in every cell, whereas ZeroFisher varies by $0.02$ to $0.04$ MRR at fixed hyperparameters, so its mean scores average over runs that barely forget and runs that over-forget.

\subsection{The update}\label{app:update}\label{app:attack}\label{app:exp_protocol}
The update is the continued training of Section~\ref{sec:exp}. It trains the released model further with the ordinary margin-ranking loss under Adam, with every parameter trainable. The update data are the retained triples together with the never-trained set $B$, and no forgotten triple is among them. The update's locality is defined against the endpoints of the probe: the 1,000 probe triples have 1,596 endpoints, and the two local components, the edges drawn from $B$ and the retained neighbours, are incident to one of these rather than to one of the 11,412 involved entities; the third axis below adds retained edges incident to none of them. Every method and the retrained oracle receive the identical edge set, and the unlearner never sees the probe. Each dataset and backbone has $48$ update configurations, shared by every method. The configurations vary four axes. The first is the number of edges drawn from the never-trained set $B$, in $\{1{,}000,\,3{,}000,\,6{,}000\}$. The second is the fraction of each probe endpoint's retained neighbours that the update trains on, in $\{0.15,\,0.25,\,0.35,\,0.5\}$. The third is the fraction of the retained edges touching no probe endpoint, in $\{0,\,0.02\}$. The fourth is the learning rate, in $\{10^{-4},\,3\times10^{-4}\}$. Every configuration runs for 20 epochs with batch size 256 and ten negatives per positive. The grid has $3\times4\times2\times2=48$ configurations. The sample of update triples is drawn per seed and is not an axis. On CoDEx-L the counts of edges drawn from $B$ are 2,000, 6,000 and 12,000, the same fractions of $B$ as on FB15k-237, and the other axes are unchanged.

All configurations run with three unlearning seeds. The probe of 1,000 deleted triples is split once into a selection half and a report half. For each method, the configuration under which it leaks most over the run on the selection half is selected; the selected configurations are then re-run with seven unlearning seeds, and each method is reported under the strongest of them against it, at each run's epoch of peak leakage. The selection half serves only this selection, and every reported number is computed on the report half. The retrained oracle (the gold model of Section~\ref{sec:exp}) receives every configuration with the same data and settings. Its runs are computed once per dataset, backbone, configuration and unlearning seed and cached. Every method is therefore read against the oracle run on the same update sample at the same epoch. Because the probe is a sample of the deleted triples, the original and gold levels on it differ from those of Table~\ref{tab:main}.

\paragraph{Relearning metrics.}
Let $\mathrm{MRR}_{\mathrm{u},e}$ and $\mathrm{MRR}_{\mathrm{gold},e}$ denote the forget-set MRR of the unlearned and gold models after $e$ epochs of the update, with $e=0$ at release, and $\mathrm{MRR}_{\mathrm{orig}}$ that of the original model. The update restores a share of each model's unlearning-induced drop,
\begin{equation}
\mathrm{Rec}_{\mathrm{u}}(e)=100\times\frac{\mathrm{MRR}_{\mathrm{u},e}-\mathrm{MRR}_{\mathrm{u},0}}{\mathrm{MRR}_{\mathrm{orig}}-\mathrm{MRR}_{\mathrm{u},0}},\qquad
\mathrm{Rec}_{\mathrm{gold}}(e)=100\times\frac{\mathrm{MRR}_{\mathrm{gold},e}-\mathrm{MRR}_{\mathrm{gold},0}}{\mathrm{MRR}_{\mathrm{orig}}-\mathrm{MRR}_{\mathrm{gold},0}},
\label{eq:exp_recovery}
\end{equation}
where $100$ is a return to the original model's forget-set MRR. The excess recovery, the leakage of \eqref{eq:leak} read at epoch $e$, and the relative excess recovery are
\begin{align}
\mathrm{ExRec}(e)&=\mathrm{Rec}_{\mathrm{u}}(e)-\mathrm{Rec}_{\mathrm{gold}}(e),\label{eq:exp_exrec}\\
\mathrm{Leak}(e)&=\bigl(\mathrm{MRR}_{\mathrm{u},e}-\mathrm{MRR}_{\mathrm{u},0}\bigr)-\bigl(\mathrm{MRR}_{\mathrm{gold},e}-\mathrm{MRR}_{\mathrm{gold},0}\bigr),\label{eq:exp_leak}\\
\mathrm{RER}(e)&=100\times\frac{\mathrm{Leak}(e)}{\mathrm{MRR}_{\mathrm{orig}}-\mathrm{MRR}_{\mathrm{gold},0}}.\label{eq:exp_rer}
\end{align}
Each is reported at $e^{\ast}=\arg\max_{e}\mathrm{Leak}(e)$, and the gold model's terms are taken at the same $e^{\ast}$ of the same update run, so they depend on the method under evaluation. RER divides Leak by one denominator per dataset and backbone, shared by every method. ExRec divides each model by its own deficit and therefore also reflects the difference in forgetting depth between the unlearned and the gold model.

\subsection{Budgets and selection}\label{app:budget}

Two selections are made. The first picks the hyperparameters of each method, and the second picks the released checkpoint within a run.

The hyperparameters are selected per method and per dataset--backbone setting by the shortfall criterion of Appendix~\ref{app:unlearning}, with the same criterion for every method. The number of epochs and the learning rates are among the searched hyperparameters. \GraphDPO{} searches its epoch count over $\{1,2,3,6,12,24,48\}$ and \KAU{} over $\{2,3,5,10,20\}$, so each method runs for the number of epochs its own selection picks. The selected \KAU{} settings on FB15k-237 are listed in Appendix~\ref{app:unlearning} and in Table~\ref{tab:hyper}. The selected settings of \GraphDPO{} on FB15k-237 are, for TransE, TransH and RotatE respectively, learning rate $10^{-4}$, $10^{-4}$ and $3\times10^{-4}$, epochs $3$, $3$ and $26$, and distillation weight $3\times10^{-3}$, $6\times10^{-3}$ and $3\times10^{-4}$. The preference weight $0.1$, the temperature $1$ and the margin $8$ are shared by the three backbones. ZeroFisher uses perturbation scale $\epsilon=10^{-3}$, update scale $10$ and $10$ iterations on TransE and TransH, and $\epsilon=10^{-4}$, update scale $1.6$ and $300$ iterations on RotatE. On CoDEx-L the selected \KAU{} settings are, for TransE, TransH and RotatE respectively, epochs 2, 10 and 20, forgetting learning rate $3\times10^{-4}$, $3\times10^{-4}$ and $3\times10^{-4}$, and $s=0.6$, $0.02$ and $0.4$, with the remaining settings as on FB15k-237. The selected settings of \GraphDPO{} are learning rate $5\times10^{-5}$, $10^{-4}$ and $10^{-3}$, epochs 48, 48 and 96, and distillation weight $3\times10^{-3}$, $3\times10^{-3}$ and $10^{-3}$, with preference weight $0.1$, temperature $1$ and margin $8$. ZeroFisher uses perturbation scale $\epsilon=10^{-5}$, update scale $0.5$ and $50$ iterations on TransE, $\epsilon=10^{-4}$, update scale $10$ and $75$ iterations on TransH, and $\epsilon=10^{-5}$, update scale $10$, $350$ iterations and damping $-0.01$ on RotatE.

The released checkpoint within a run is chosen by the rule of Appendix~\ref{app:selection}. Once per epoch the run scores its current checkpoint by $0.7\,(1-\mathrm{MRR}_{\Fset})+0.3\,\mathrm{MRR}_R$ on the development subsets the unlearner holds. The released model is the checkpoint with the highest score. The rule is applied unchanged to every method. It consults neither the oracle nor any edge of the update.

\subsection{Relearning-resistant add-ons}\label{app:addons}

Table~\ref{tab:relearn_addons} adds three relearning defences to \GraphDPO{} on FB15k-237. Each add-on is built on \GraphDPO{} at its selected setting on that backbone (Appendix~\ref{app:budget}).

\paragraph{NTK.}
We apply the scrubbing step of \citet{golatkar2020forgetting} once to \GraphDPO{}'s released model. The step moves the parameters along the forgotten facts' score gradients, projected off the span of the retained triples' score gradients, with the step length of the authors' code. It then adds Gaussian noise with covariance $\alpha F^{-1}$, where $F$ is the diagonal Fisher on the retained triples. We search the authors' step with $\alpha\in\{0,10^{-10},10^{-9}\}$ and twice and four times that step with $\alpha=0$. 

\paragraph{SAM.}
We follow Algorithm~A1 of \citet{fan2025sam} at every \GraphDPO{} step. \GraphDPO{}'s forgetting terms, its margin and preference losses on the forget batch, are evaluated at parameters perturbed by $\rho$ along their normalized gradient, and its distillation and replay terms at the unperturbed parameters. We search $\rho\in\{0.03,0.1,0.3\}$ on TransE and $\rho\in\{0.003,0.01,0.03,0.1,0.3\}$ on TransH and RotatE, adding epochs where needed so that the update is as large as \GraphDPO{}'s. We select $\rho=0.03$, $0.01$ and $0.03$ at $3$, $3$ and $27$ epochs on TransE, TransH and RotatE.

\paragraph{TAR.}
We continue \GraphDPO{}'s released model with the adversarial training of \citet{tamirisa2025tar} for 750 outer steps. Each outer step simulates one update at learning rate $10^{-4}$ or $3\times10^{-4}$, on retained edges around a batch of forgotten facts, with or without the stand-ins of Section~\ref{sec:method:draw}. The tamper-resistance loss penalizes any rise of a forgotten fact's log-odds against its 50 strongest competing answers above its level in \GraphDPO{}'s release, both along the simulated update and at the current parameters, and is differentiated to first order as in the authors' code. A retained margin loss and an anchor to \GraphDPO{}'s release hold utility. With the tamper-resistance weight fixed at $0.2$, we search the outer learning rate over $\{2,4,8,16,30\}\times10^{-5}$ and select $2\times10^{-5}$ on TransE and TransH and $8\times10^{-5}$ on RotatE. This add-on differs from the TAR-style objective of Appendix~\ref{app:alternatives}, which replaces \KAU{}'s forgetting step.

\paragraph{Selection and evaluation.}
For each add-on, the five settings are scored by the criterion of Appendix~\ref{app:unlearning}, and the selected setting is kept at every seed. Every row of Table~\ref{tab:relearn_addons}, \GraphDPO{} and \KAU{} included, is run at three of the seven unlearning seeds of Table~\ref{tab:relearn} and receives the same three to six update configurations of Appendix~\ref{app:exp_protocol} on each backbone. Each row is reported under the configuration in which it leaks most, at its epoch of peak leakage.

\subsection{Statistics}\label{app:stats}

Table~\ref{tab:main} and Table~\ref{tab:relearn} report means over seven unlearning seeds. Table~\ref{tab:main_std} gives the standard deviations of Table~\ref{tab:main} over those seeds, and Table~\ref{tab:relearn_std} those of the FB15k-237 block of Table~\ref{tab:relearn}. The standard deviation of the leakage is $0.0021$, $0.0019$ and $0.0032$ for \KAU{} on TransE, TransH and RotatE, $0.0057$, $0.0079$ and $0.0078$ for \GraphDPO{}, and $0.0470$, $0.0421$ and $0.0311$ for ZeroFisher. For ZeroFisher the excess recovery spreads far more than the leakage, because its release-time drop, the denominator of its excess recovery, varies across seeds. Table~\ref{tab:relearn} prints the excess recovery as the ratio of seed means, and Table~\ref{tab:relearn_std} gives the mean and spread of the per-seed ratios. Tables~\ref{tab:relearn_addons}, \ref{tab:kau_ablation_recovery_anc03}, \ref{tab:kau_ablation_full} and~\ref{tab:ablation_recovery_depth} and Figures~\ref{fig:depth}, \ref{fig:retain_depth}, \ref{fig:depth_heatmap_rotate} and~\ref{fig:ablation_spread} report means over three unlearning seeds. Table~\ref{tab:attacks} reports means over the seven seeds of Table~\ref{tab:relearn}. Table~\ref{tab:sweep} uses three screening seeds per configuration, and its last row uses the seven seeds of Table~\ref{tab:relearn}. Tables~\ref{tab:blocks} and~\ref{tab:predictor} use one unlearning seed. Figure~\ref{fig:curves} shows means over three unlearning seeds under one update configuration, with every method released at a forgetting rate of 90 to 106 percent of the gold model's. A seed varies the unlearner's own randomness and the sample of update triples, while the split, the original model and the update configurations stay fixed. Selection and reporting use the same seeds (Appendix~\ref{app:unlearning}). Table~\ref{tab:relearn_tests} tests the leakage differences of Table~\ref{tab:relearn}, seven \KAU{} seeds against seven baseline seeds, with Welch's $t$-test and the Mann--Whitney $U$ test, both two-sided. On every backbone every \KAU{} seed leaks less than every \GraphDPO{} seed. Both tests find \KAU{}'s leakage below \GraphDPO{}'s at the 0.1\% level and below ZeroFisher's at the 5\% level on every backbone.

\begin{table}[htbp]
\centering\small
\caption{Seed spread of Table~\ref{tab:relearn} on FB15k-237: mean $\pm$ standard deviation over the seven unlearning seeds at each method's reported update configuration. ExRec is the mean of the per-seed ratios, whereas Table~\ref{tab:relearn} prints the ratio of seed means.}
\label{tab:relearn_std}
\begin{tabular}{llcccc}
\toprule
Backbone & Method & ExRec (\%) & Leak & RER (\%) & $M_R$ \\
\midrule
TransE & \KAU{} & 48.91 $\pm$ 4.96 & 0.0215 $\pm$ 0.0021 & 37.45 $\pm$ 3.66 & 0.2839 $\pm$ 0.0008 \\
 & \GraphDPO{} & 102.13 $\pm$ 10.07 & 0.0437 $\pm$ 0.0057 & 76.12 $\pm$ 9.95 & 0.2628 $\pm$ 0.0013 \\
 & ZeroFisher & 231.11 $\pm$ 222.71 & 0.0741 $\pm$ 0.0470 & 129.00 $\pm$ 81.77 & 0.2322 $\pm$ 0.0346 \\
\midrule
TransH & \KAU{} & 6.44 $\pm$ 2.05 & 0.0211 $\pm$ 0.0019 & 71.99 $\pm$ 6.48 & 0.2614 $\pm$ 0.0005 \\
 & \GraphDPO{} & 72.95 $\pm$ 23.00 & 0.0394 $\pm$ 0.0079 & 134.22 $\pm$ 26.94 & 0.2401 $\pm$ 0.0014 \\
 & ZeroFisher & 343.97 $\pm$ 447.50 & 0.0687 $\pm$ 0.0421 & 234.02 $\pm$ 143.34 & 0.2164 $\pm$ 0.0374 \\
\midrule
RotatE & \KAU{} & 23.71 $\pm$ 2.47 & 0.0239 $\pm$ 0.0032 & 23.36 $\pm$ 3.13 & 0.3254 $\pm$ 0.0007 \\
 & \GraphDPO{} & 78.34 $\pm$ 8.47 & 0.0641 $\pm$ 0.0078 & 62.65 $\pm$ 7.66 & 0.2531 $\pm$ 0.0015 \\
 & ZeroFisher & 175.23 $\pm$ 122.56 & 0.0857 $\pm$ 0.0311 & 83.75 $\pm$ 30.34 & 0.2261 $\pm$ 0.0481 \\
\bottomrule
\end{tabular}
\end{table}

\begin{table}[htbp]
\centering\small
\caption{Significance of the leakage differences in Table~\ref{tab:relearn} on FB15k-237: \KAU{} against each baseline, seven unlearning seeds against seven, at the reported update configurations. Welch's $t$-test and the Mann--Whitney $U$ test, both two-sided.}
\label{tab:relearn_tests}
\begin{tabular}{llccccc}
\toprule
Backbone & Comparison & $\Delta$Leak & Welch $t$ & $p$ & $U$ & $p$ \\
\midrule
TransE & \KAU{} vs.\ \GraphDPO{} & $-0.0222$ & $-9.67$ & $<10^{-4}$ & 0 & 0.0006 \\
 & \KAU{} vs.\ ZeroFisher & $-0.0526$ & $-2.96$ & 0.025 & 7 & 0.026 \\
\midrule
TransH & \KAU{} vs.\ \GraphDPO{} & $-0.0183$ & $-5.96$ & 0.0007 & 0 & 0.0006 \\
 & \KAU{} vs.\ ZeroFisher & $-0.0475$ & $-2.98$ & 0.025 & 7 & 0.026 \\
\midrule
RotatE & \KAU{} vs.\ \GraphDPO{} & $-0.0402$ & $-12.62$ & $<10^{-4}$ & 0 & 0.0006 \\
 & \KAU{} vs.\ ZeroFisher & $-0.0618$ & $-5.23$ & 0.0018 & 2 & 0.0023 \\
\bottomrule
\end{tabular}
\end{table}

Section~\ref{sec:setup} places the retrained oracle at zero leakage, so the width of that zero is the oracle's own variation across retraining seeds. Table~\ref{tab:oracle_seeds} measures it. We retrain the oracle twice more per backbone with the recipe of Appendix~\ref{app:appendix_traningl}, under retraining seeds 2 and 3. Each retrained oracle receives the update of Table~\ref{tab:relearn} at every method's reported update configuration for the seven unlearning seeds. Every run is read at the same per-seed peak epoch as Table~\ref{tab:relearn}. The table reports the oracle's own rise in forget-MRR under the update at the three oracle seeds. It also reports each method's leakage recomputed against each oracle seed and each method's own rise over its seven seeds. The oracle's rise varies across retraining seeds by a standard deviation of $0.0036$ to $0.0043$ on TransE, $0.0032$ to $0.0042$ on TransH and $0.0018$ to $0.0031$ on RotatE. Its retained MRR at release varies by $0.0017$, $0.0024$ and $0.0025$. Recomputed against each oracle seed, \KAU{}'s leakage stays the smallest of the three methods on every backbone, and its reduction over \GraphDPO{} stays between 44\% and 67\%. The reductions of Table~\ref{tab:relearn} hold whichever oracle seed is the reference. \KAU{}'s leakage in Table~\ref{tab:relearn} is $5.0$, $5.0$ and $7.7$ times the largest oracle-seed standard deviation on TransE, TransH and RotatE. It is a rise beyond the oracle and not seed variation of the reference. In the last column, every method's rise exceeds the oracle's at every oracle seed on every backbone.

\begin{table}[htbp]
\centering\scriptsize
\setlength{\tabcolsep}{3pt}
\caption{The retrained oracle at three retraining seeds on FB15k-237. Each oracle receives the update of Table~\ref{tab:relearn} at every method's reported update configuration for the seven unlearning seeds and is read at the same peak epochs. Oracle rise is the oracle's own gain in forget-MRR under the update, mean over the seven update samples, at oracle seeds 1, 2 and 3. Leak is each method's leakage against each oracle seed. Method rise is the method's own gain, mean $\pm$ standard deviation over the seven seeds. The per-backbone lines give the oracle's $M_R$ at release at the three seeds and the leakage reduction of \KAU{} over \GraphDPO{} against each seed.}
\label{tab:oracle_seeds}
\begin{tabular}{llccc}
\toprule
Backbone & Method & Oracle rise (seeds 1/2/3) & Leak vs.\ seeds 1/2/3 & Method rise \\
\midrule
TransE & \GraphDPO{} & 0.0351/0.0342/0.0417 & 0.0437/0.0447/0.0372 & 0.0789 $\pm$ 0.0024 \\
 & ZeroFisher & 0.0192/0.0165/0.0235 & 0.0741/0.0768/0.0698 & 0.0933 $\pm$ 0.0499 \\
 & \KAU{} & 0.0419/0.0385/0.0471 & 0.0215/0.0249/0.0163 & 0.0634 $\pm$ 0.0049 \\
 & \multicolumn{4}{@{}l}{Oracle $M_R$ at release 0.2932/0.2899/0.2906} \\ & \multicolumn{4}{@{}l}{Leakage reduction of \KAU{} over \GraphDPO{} 50.8/44.4/56.0\%} \\
\midrule
TransH & \GraphDPO{} & 0.0303/0.0365/0.0344 & 0.0394/0.0332/0.0353 & 0.0697 $\pm$ 0.0137 \\
 & ZeroFisher & 0.0217/0.0290/0.0271 & 0.0687/0.0614/0.0632 & 0.0904 $\pm$ 0.0362 \\
 & \KAU{} & 0.0387/0.0471/0.0431 & 0.0211/0.0127/0.0167 & 0.0598 $\pm$ 0.0084 \\
 & \multicolumn{4}{@{}l}{Oracle $M_R$ at release 0.2739/0.2695/0.2703} \\ & \multicolumn{4}{@{}l}{Leakage reduction of \KAU{} over \GraphDPO{} 46.4/61.7/52.5\%} \\
\midrule
RotatE & \GraphDPO{} & 0.0308/0.0319/0.0366 & 0.0641/0.0629/0.0583 & 0.0949 $\pm$ 0.0099 \\
 & ZeroFisher & 0.0151/0.0126/0.0160 & 0.0857/0.0882/0.0848 & 0.1008 $\pm$ 0.0257 \\
 & \KAU{} & 0.0414/0.0424/0.0464 & 0.0239/0.0229/0.0190 & 0.0653 $\pm$ 0.0058 \\
 & \multicolumn{4}{@{}l}{Oracle $M_R$ at release 0.3301/0.3278/0.3328} \\ & \multicolumn{4}{@{}l}{Leakage reduction of \KAU{} over \GraphDPO{} 62.7/63.6/67.4\%} \\
\bottomrule
\end{tabular}
\end{table}

\subsection{CoDEx-L}\label{app:codexl}

CoDEx-L is run on TransE, TransH and RotatE under the protocol of this appendix, with the splits of Appendix~\ref{app:appendix_traningl} and the update grid of Appendix~\ref{app:exp_protocol}. Table~\ref{tab:main_codexl} gives link prediction at release, with Gold the gold model of Section~\ref{sec:exp}. Table~\ref{tab:relearn_codexl_std} gives the seed spread and the tests behind the CoDEx-L block of Table~\ref{tab:relearn}.

\begin{table}[t]
\centering\small
\caption{Link prediction at release on CoDEx-L with TransE, TransH and RotatE: filtered MRR on the forget, retain and test evaluation sets. Original and Gold are single checkpoints. Method rows are mean $\pm$ standard deviation over seven unlearning seeds. Bold marks the method closest to Gold on $M_F$ and the highest on $M_R$ and $M_T$.}
\label{tab:main_codexl}
\begin{tabular}{llccc}
\toprule
Backbone & Method & $M_F$ & $M_R\uparrow$ & $M_T\uparrow$ \\
\midrule
TransE & Original & 0.2354 & 0.2462 & 0.1501 \\
       & Gold     & 0.1487 & 0.2655 & 0.1492 \\
\cmidrule{2-5}
       & GraphDPO   & 0.1630 $\pm$ 0.0004 & 0.2154 $\pm$ 0.0013 & 0.1389 $\pm$ 0.0006 \\
       & ZeroFisher & 0.1655 $\pm$ 0.0655 & 0.1734 $\pm$ 0.0683 & 0.1081 $\pm$ 0.0404 \\
       & KAU        & \textbf{0.1504} $\pm$ 0.0002 & \textbf{0.2474} $\pm$ 0.0005 & \textbf{0.1466} $\pm$ 0.0002 \\
\midrule
TransH & Original & 0.2412 & 0.2535 & 0.1505 \\
       & Gold     & 0.1509 & 0.2704 & 0.1534 \\
\cmidrule{2-5}
       & GraphDPO   & 0.1609 $\pm$ 0.0011 & 0.2091 $\pm$ 0.0008 & 0.1334 $\pm$ 0.0011 \\
       & ZeroFisher & 0.1446 $\pm$ 0.0607 & 0.1515 $\pm$ 0.0641 & 0.0956 $\pm$ 0.0355 \\
       & KAU        & \textbf{0.1475} $\pm$ 0.0004 & \textbf{0.2530} $\pm$ 0.0008 & \textbf{0.1461} $\pm$ 0.0003 \\
\midrule
RotatE & Original & 0.2783 & 0.2916 & 0.1557 \\
       & Gold     & 0.1554 & 0.3235 & 0.1574 \\
\cmidrule{2-5}
       & GraphDPO   & 0.1900 $\pm$ 0.0011 & 0.2019 $\pm$ 0.0010 & 0.1041 $\pm$ 0.0007 \\
       & ZeroFisher & 0.1586 $\pm$ 0.0532 & 0.1666 $\pm$ 0.0563 & 0.1006 $\pm$ 0.0324 \\
       & KAU        & \textbf{0.1561} $\pm$ 0.0002 & \textbf{0.3051} $\pm$ 0.0004 & \textbf{0.1556} $\pm$ 0.0001 \\
\bottomrule
\end{tabular}
\end{table}

\begin{table}[htbp]
\centering\small
\caption{Seed spread and significance of the CoDEx-L block of Table~\ref{tab:relearn}. Cells give mean $\pm$ standard deviation over seven unlearning seeds at each method's reported update configuration. The test lines give two-sided Welch $t$ and Mann--Whitney $U$ tests of \KAU{} against each baseline on the leakage. ExRec is the mean of the per-seed ratios.}
\label{tab:relearn_codexl_std}
\begin{tabular}{llcccc}
\toprule
Backbone & Method & ExRec (\%) & Leak & RER (\%) & $M_R$ \\
\midrule
TransE & \GraphDPO{} & 77.35 $\pm$ 3.60 & 0.0524 $\pm$ 0.0031 & 31.30 $\pm$ 1.83 & 0.2154 $\pm$ 0.0013 \\
 & ZeroFisher & 218.34 $\pm$ 122.10 & 0.1080 $\pm$ 0.0844 & 64.49 $\pm$ 50.37 & 0.1734 $\pm$ 0.0683 \\
 & \KAU{} & 26.94 $\pm$ 1.60 & 0.0214 $\pm$ 0.0012 & 12.80 $\pm$ 0.71 & 0.2474 $\pm$ 0.0005 \\
\midrule
\multicolumn{6}{@{}l}{Leak, \KAU{} vs.\ \GraphDPO{}: Welch $t=-24.93$, $p<10^{-4}$, and $U=0$, $p=0.0006$} \\
\multicolumn{6}{@{}l}{Leak, \KAU{} vs.\ ZeroFisher: Welch $t=-2.71$, $p=0.035$, and $U=7$, $p=0.026$} \\
\midrule
TransH & \GraphDPO{} & 75.61 $\pm$ 4.80 & 0.0621 $\pm$ 0.0034 & 39.41 $\pm$ 2.15 & 0.2091 $\pm$ 0.0008 \\
 & ZeroFisher & 168.63 $\pm$ 65.18 & 0.1438 $\pm$ 0.0438 & 91.25 $\pm$ 27.80 & 0.1515 $\pm$ 0.0641 \\
 & \KAU{} & 27.10 $\pm$ 1.57 & 0.0310 $\pm$ 0.0034 & 19.68 $\pm$ 2.14 & 0.2530 $\pm$ 0.0008 \\
\midrule
\multicolumn{6}{@{}l}{Leak, \KAU{} vs.\ \GraphDPO{}: Welch $t=-17.22$, $p<10^{-4}$, and $U=0$, $p=0.0006$} \\
\multicolumn{6}{@{}l}{Leak, \KAU{} vs.\ ZeroFisher: Welch $t=-6.79$, $p=0.0005$, and $U=0$, $p=0.0006$} \\
\midrule
RotatE & \GraphDPO{} & 108.82 $\pm$ 6.45 & 0.1000 $\pm$ 0.0063 & 53.66 $\pm$ 3.37 & 0.2019 $\pm$ 0.0010 \\
 & ZeroFisher & 192.86 $\pm$ 133.16 & 0.1328 $\pm$ 0.0210 & 71.29 $\pm$ 11.26 & 0.1666 $\pm$ 0.0563 \\
 & \KAU{} & 24.76 $\pm$ 4.30 & 0.0179 $\pm$ 0.0052 & 9.62 $\pm$ 2.79 & 0.3051 $\pm$ 0.0004 \\
\midrule
\multicolumn{6}{@{}l}{Leak, \KAU{} vs.\ \GraphDPO{}: Welch $t=-26.63$, $p<10^{-4}$, and $U=0$, $p=0.0006$} \\
\multicolumn{6}{@{}l}{Leak, \KAU{} vs.\ ZeroFisher: Welch $t=-14.06$, $p<10^{-4}$, and $U=0$, $p=0.0006$} \\
\bottomrule
\end{tabular}
\end{table}

The ordering of Table~\ref{tab:relearn} holds on a second graph with over five times the entities and under a third of the relations, on all three backbones. \KAU{} attains the lowest excess recovery, leakage and relative excess recovery on every backbone. Its leakage is 59\%, 50\% and 82\% below \GraphDPO{}'s on TransE, TransH and RotatE. Its retained MRR exceeds \GraphDPO{}'s by $0.032$, $0.044$ and $0.103$. On TransE it is $0.018$ below the gold model's, where \GraphDPO{} and ZeroFisher lose $0.050$ and $0.092$, and \KAU{}'s forget-MRR at release is within $0.002$ of the gold model's. On TransE all three methods are read at one update configuration, the one that leaks most against each of them on the selection half. On every backbone every \KAU{} seed leaks less than every \GraphDPO{} seed. Both tests find \KAU{}'s leakage below ZeroFisher's at the 5\% level on every backbone.

\section{Additional results}\label{app:results}

This appendix holds the measurements that the main text uses in summary. The first is the leakage across the 48 update configurations, and the second reads the update built from never-trained edges alone. The next two attribute the TransE result between the forgetting term and the ancillaries, and the recovery between parameter blocks. The next two are the calibration of the anticipated response as a ranking and the relation of forgetting depth to leakage and retained utility. The definitions and settings behind the ablation tables of Section~\ref{sec:exp} follow, and a note on evaluating the anticipated score at the shifted rows closes the appendix. Leakage is Leak of \eqref{eq:exp_leak}, in MRR points, and RER of \eqref{eq:exp_rer} is leakage as a share of the erased knowledge.

\subsection{Leakage across update configurations}\label{app:sweep}

Section~\ref{sec:exp} runs 48 update configurations per dataset and backbone, shared by every method. Each configuration is one draw of the update of Section~\ref{sec:setup}, the continued training of Section~\ref{sec:exp}, run on retained triples together with the never-trained triples $B$. The retrained oracle, the gold model of Section~\ref{sec:exp}, receives every configuration with identical data and settings. Table~\ref{tab:relearn} reports each method under the configuration that leaks most against it on the selection half of the probe, read on the report half at the epoch of peak leakage. Table~\ref{tab:sweep} shows what lies behind that selected cell on TransE. The grid is that of Appendix~\ref{app:exp_protocol}, and every configuration runs at three screening seeds. A configuration's leakage is the mean over those seeds of the leakage at each seed's peak epoch. Each cell of the table is the largest such value over the configurations of its group. The last row gives each method under its own configuration in Table~\ref{tab:relearn}, over seven unlearning seeds. It can therefore lie outside the three-seed range of the rows above it, as it does for ZeroFisher, whose seed spread is the largest (Table~\ref{tab:relearn_std}).

\begin{table}[htbp]
\centering\small
\caption{Leakage on FB15k-237 with TransE by group of update configurations. Each cell is the largest leakage over the configurations of the group, where a configuration's leakage is the mean over the three screening seeds of the leakage at that seed's peak epoch. The last column is the ratio of \KAU{}'s cell to \GraphDPO{}'s. The last row gives each method under its own configuration in Table~\ref{tab:relearn}, over seven seeds.}
\label{tab:sweep}
\setlength{\tabcolsep}{6pt}
\begin{tabular}{llcccc}
\toprule
Axis & Value & \GraphDPO{} & ZeroFisher & \KAU{} & \KAU{}/\GraphDPO{} \\
\midrule
$|B|$ drawn & 1,000 & 0.0467 & 0.0469 & 0.0248 & 0.53 \\
 & 3,000 & 0.0462 & 0.0475 & 0.0247 & 0.53 \\
 & 6,000 & 0.0470 & 0.0440 & 0.0249 & 0.53 \\
\midrule
neighbour fraction & 0.15 & 0.0467 & 0.0471 & 0.0249 & 0.53 \\
 & 0.25 & 0.0456 & 0.0421 & 0.0215 & 0.47 \\
 & 0.35 & 0.0470 & 0.0475 & 0.0248 & 0.53 \\
 & 0.5 & 0.0462 & 0.0401 & 0.0247 & 0.53 \\
\midrule
random retained fraction & 0 & 0.0470 & 0.0475 & 0.0248 & 0.53 \\
 & 0.02 & 0.0425 & 0.0395 & 0.0249 & 0.58 \\
\midrule
learning rate & $10^{-4}$ & 0.0470 & 0.0440 & 0.0247 & 0.53 \\
 & $3\times10^{-4}$ & 0.0462 & 0.0475 & 0.0249 & 0.54 \\
\midrule
All 48 & & 0.0470 & 0.0475 & 0.0249 & 0.53 \\
Reported (7 seeds) & & 0.0437 & 0.0741 & 0.0215 & 0.49 \\
\bottomrule
\end{tabular}
\end{table}

The table is read down each column. \KAU{} leaks less than \GraphDPO{} and less than ZeroFisher in every one of the 48 configurations. Its leakage is 0.47 to 0.58 of \GraphDPO{}'s in every group. Leakage is flat along every axis. \GraphDPO{}'s group maxima span 0.0425 to 0.0470, and \KAU{}'s span 0.0215 to 0.0249. The worst configuration therefore sets the scale, and no axis moves \KAU{} from the bottom of the ordering. \KAU{}'s place below both baselines in Table~\ref{tab:relearn} is not a property of one configuration. The configuration reported in Table~\ref{tab:relearn} is chosen by the selection of Appendix~\ref{app:exp_protocol}, which reads leakage over the run on the selection half, whereas this table reads peak leakage on the report half; the two readings need not pick the same configuration, and under either \KAU{} is lowest in every group.

\subsection{The never-trained-edge update}\label{app:attackvar}

Table~\ref{tab:attacks} reads one update outside the 48 configurations on FB15k-237 with TransE, beside the standard update, over the seven unlearning seeds of Table~\ref{tab:relearn}. Each update is applied to the released checkpoint of every method and to the retrained oracle. ExRec, Leak and RER are read at the epoch of peak leakage as in Table~\ref{tab:relearn}. The standard update is the configuration of Table~\ref{tab:blocks}. It draws 3,000 edges from the never-trained set $B$ and 25 percent of the retained neighbours of each probe endpoint, 40,021 edges in all, at learning rate $10^{-4}$ for 20 epochs. The never-trained-only update holds the 6,000 edges of $B$ and nothing else, under the same learning rate and epoch budget.

\begin{table}[htbp]
\centering\small
\caption{The never-trained-edge update on FB15k-237 with TransE, means over seven unlearning seeds. Each row applies one update to the released checkpoint of every method and to the retrained oracle. ExRec, Leak and RER are read at the epoch of peak leakage as in Table~\ref{tab:relearn}. The standard update is the configuration of Table~\ref{tab:blocks}, and the other lies outside the grid. Columns follow equations~\ref{eq:exp_exrec}--\ref{eq:exp_rer}.}
\label{tab:attacks}
\setlength{\tabcolsep}{4pt}
\begin{tabular}{lrrrrrrrrr}
\toprule
 & \multicolumn{3}{c}{\GraphDPO{}} & \multicolumn{3}{c}{ZeroFisher} & \multicolumn{3}{c}{\KAU{}} \\
\cmidrule(lr){2-4}\cmidrule(lr){5-7}\cmidrule(lr){8-10}
Update & ExRec$\downarrow$ & Leak$\downarrow$ & RER$\downarrow$ & ExRec$\downarrow$ & Leak$\downarrow$ & RER$\downarrow$ & ExRec$\downarrow$ & Leak$\downarrow$ & RER$\downarrow$ \\
\midrule
standard & 95.26 & 0.0402 & 69.93 & 132.11 & 0.0697 & 121.35 & 48.53 & 0.0215 & 37.45 \\
never-trained edges only & 61.12 & 0.0274 & 47.75 & 104.15 & 0.0553 & 96.30 & 19.36 & 0.0089 & 15.55 \\
\bottomrule
\end{tabular}
\end{table}

The never-trained-only update holds only edges that no model has ever trained on. This is the channel the unlearner can model only through stand-ins (Section~\ref{sec:method:draw}). Under it \GraphDPO{} keeps 68 percent of its leakage and ZeroFisher 79 percent. Removing the retained channel therefore does not remove the baselines' leakage. The residual they leave on the involved rows is restored by any true edge at those rows, as Section~\ref{sec:method:controls} says. \KAU{} keeps 42 percent of its leakage. Its leakage is 0.33 of \GraphDPO{}'s under this update, against 0.54 under the standard one. This is what the stand-in channel of Section~\ref{sec:method:draw} predicts, together with the observation of Section~\ref{sec:theory} that the retained span misses these directions. At the unlearning seed of Table~\ref{tab:blocks}, the retrained oracle's own rise ranges from $+0.022$ to $+0.041$ under the standard update. Under the never-trained-only update it falls to a range of $+0.011$ to $+0.015$. The ratio between the methods is the reading.

Running the standard update for longer changes the leakage little. Over 40 epochs, \KAU{}'s leakage is $0.0217$ against $0.0215$ at 20 epochs, and \GraphDPO{}'s is $0.0412$ against $0.0402$. ZeroFisher's is unchanged at $0.0697$. The peak-epoch reading of Section~\ref{sec:exp} therefore captures nearly all the leakage a longer update produces, and \KAU{} still leaks about half as much as \GraphDPO{}.

\subsection{The forgetting term against the ancillaries}\label{app:twobytwo}

\KAU{} runs its anticipated forgetting loss with two ancillaries, the score-preservation term of \eqref{eq:total} and the norm retraction of Appendix~\ref{app:retraction}.
The TransE result therefore has to be attributed among the three.
The rows of Tables~\ref{tab:kau_ablation_recovery_anc03}, \ref{tab:ablation_recovery_depth} and~\ref{tab:kau_ablation_full} that change one of them at a time settle the attribution.
Every such row except the relational anchor and the row without a forgetting term is released at \KAU{}'s forgetting depth, so the rows compare leakage at equal forgetting.
Section~\ref{sec:method:controls} predicts the pattern they show.
Its first control is the residual the edit leaves on the known edges, which Proposition~\ref{prop:anchor} ties to the score-preservation term.
Without that term the update pulls the raised retained edges back and drags the forgotten facts with them.
The row without a score anchor in Table~\ref{tab:kau_ablation_recovery_anc03} shows exactly this.
Its leakage rises by almost half and its retained utility falls by $0.007$, the lowest utility among \KAU{}'s variants.
The relational anchor of Table~\ref{tab:kau_ablation_full} holds pairwise distances rather than scores, so it leaves the residual free.
It leaks the most of all \KAU{}'s variants except the rows without the anticipated response or with a damped response, although it forgets the deepest.
What matters on the known channel is therefore the score residual and not the geometry of the rows, as the first control says.
The second control is where the forgetting is placed.
The plain forgetting loss of Table~\ref{tab:ablation_recovery_depth}, which is \KAU{} without the anticipated response, leaks more than the anticipated loss at the same depth.
The anticipated term therefore lowers leakage on its own and not through a deeper release.

The same rows test the two statements about the filter.
Section~\ref{sec:theory} reads subspace methods as hard thresholds of the survival filter, and Corollary~\ref{cor:thresholds} places null-space projection at its zero threshold.
The retained-span projection of Table~\ref{tab:ablation_recovery_depth}, the hard threshold without stand-ins, leaks more than \KAU{} at the same depth.
Adding the projector on top of the anticipated loss, the projector row of Table~\ref{tab:kau_ablation_recovery_anc03}, raises leakage instead of lowering it.
The hard threshold therefore adds nothing that the soft weights of the filter do not already provide, and it removes directions that the update would have left alone (Remark~\ref{rem:cone}).
Section~\ref{sec:method:anc} keeps the norm retraction because norm inflation could mimic resistance on TransE and TransH (Appendix~\ref{app:retraction}).
The row without retraction in Table~\ref{tab:kau_ablation_full} leaks almost the same as the full method at the same utility, so the low leakage does not rest on norm control.
The anticipated term is what lowers leakage. The score anchor holds utility while that term acts, and the retraction is a safeguard rather than a source of the result.
One reading note applies at unequal depth.
The excess recovery of the relational anchor barely moves, because a deeper release enlarges the denominator of \eqref{eq:exp_recovery}.
Leakage in MRR points is therefore the column to compare there.

\subsection{Which parameters the update restores}\label{app:blocks}

The unlearner edits only the rows of the involved entities, and the update restores the forgotten facts through the same rows. To read off which block carries the recovery we build hybrid models. A hybrid model takes one parameter block from the updated model and every other block from the released model, and the forget-MRR is evaluated on it. Table~\ref{tab:blocks} gives the result on TransE for \KAU{}, \GraphDPO{}, ZeroFisher and the retrained oracle. All four face one update configuration with one unlearning seed, and each model is read at its own epoch of peak leakage. Forget-MRR is measured on the 500-edge report half of the held-out forget probe. The blocks are the involved rows, the relation rows and the entity rows the edit never touched. The involved rows are the endpoint rows of the forgotten facts, 11,412 of the 14,541 entities. The involved rows alone reproduce 95 to 105 percent of every model's full rise. The relation rows reproduce at most 5.4 percent of the rise for the three unlearners and 9.5 percent for the oracle. The other entity rows move the forget-MRR by at most 0.0014. Each hybrid replaces one block at a time, so the three rises need not add up to the full rise, and two rows exceed 100 percent. Over the seven seeds of Table~\ref{tab:relearn}, the involved rows reproduce 94 to 105 percent of the rise for \KAU{}, \GraphDPO{} and the oracle, and 91 to 105 for ZeroFisher. A forgotten fact therefore relearns through its own two endpoint rows, for an edited model and for the retrained oracle alike. This is the premise of Section~\ref{sec:setup} and of the anticipated response.

\begin{table}[htbp]
\centering\small
\caption{Block attribution of the rise in forget-MRR under the update, FB15k-237 with TransE, one update configuration and one unlearning seed. Released is the forget-MRR of the released model. Each further column gives the rise in forget-MRR of the hybrid that takes that block alone from the updated model, and the every-parameter column gives the rise of the updated model itself. Each model is read at its own epoch of peak leakage.}
\label{tab:blocks}
\setlength{\tabcolsep}{6pt}
\begin{tabular}{lccccc}
\toprule
Model & Released & Every parameter & Involved rows & Relation rows & Other entity rows \\
\midrule
\KAU{} & 0.2265 & $+0.0580$ & $+0.0553$ & $+0.0002$ & $+0.0005$ \\
\GraphDPO{} & 0.2292 & $+0.0875$ & $+0.0845$ & $+0.0012$ & $+0.0014$ \\
ZeroFisher & 0.2527 & $+0.0632$ & $+0.0635$ & $+0.0034$ & $-0.0011$ \\
retrained oracle & 0.2240 & $+0.0381$ & $+0.0400$ & $+0.0036$ & $+0.0007$ \\
\bottomrule
\end{tabular}
\end{table}

\subsection{The anticipated response as a per-fact ranking}\label{app:predictor}

The objective uses the response to order the forgotten facts by how much the update would restore them, so we evaluate it as a ranking. We take \KAU{}'s released TransE checkpoint and its updated checkpoint at the epoch of peak leakage under one update configuration and one unlearning seed. The predicted change of a forgotten fact's score is the response at the released checkpoint, averaged over 32 draws. The realized change is the fact's score at release minus its score after the update, so a positive change is a recovery. Table~\ref{tab:predictor} compares the two over the $n=1{,}000$ held-out forgotten facts for the three ways of computing the response.

\begin{table}[htbp]
\centering\small
\caption{The anticipated response as a per-fact ranking on \KAU{}'s released TransE checkpoint on FB15k-237, under one update configuration and one unlearning seed. Per-fact correlation between the predicted and the realized change of the forgotten facts' scores under the update, at the epoch of peak leakage, for the three ways of computing the response. The last two columns give the mean predicted and the mean realized change, a positive change being a recovery. $n=1{,}000$ facts, response averaged over 32 draws, 95 percent bootstrap intervals in brackets. The last line gives the mean realized recovery within each quintile of predicted recovery.}
\label{tab:predictor}
\footnotesize
\setlength{\tabcolsep}{3pt}
\begin{tabular}{@{}lcccc@{}}
\toprule
Response & Pearson & Spearman & Mean predicted & Mean realized \\
\midrule
joint solve, exact evaluation (\KAU{}) & 0.526 {\scriptsize[0.294,\,0.674]} & 0.207 {\scriptsize[0.145,\,0.266]} & $-0.009$ & $+0.159$ \\
per-entity block solve & 0.303 {\scriptsize[0.167,\,0.422]} & 0.197 {\scriptsize[0.135,\,0.259]} & $-0.308$ & $+0.159$ \\
first-order evaluation & 0.605 {\scriptsize[0.333,\,0.746]} & 0.125 {\scriptsize[0.061,\,0.188]} & $+0.145$ & $+0.159$ \\
\midrule
\multicolumn{5}{@{}l}{Realized recovery by quintile of prediction (joint solve): $+0.003,\,+0.052,\,+0.085,\,+0.196,\,+0.459$} \\
\bottomrule
\end{tabular}
\end{table}

The joint response with exact evaluation reaches a Pearson correlation of 0.53 and a Spearman correlation of 0.21 per fact. Its quintiles of predicted recovery are monotone in realized recovery, from $+0.003$ in the lowest quintile to $+0.459$ in the highest. This is the ranking property the objective uses. The per-entity block solve lowers the Pearson correlation to 0.30. A fact whose two endpoints are both edited therefore has to be solved once, jointly, for the prediction to be right. First-order evaluation ranks the facts worse, at a Spearman correlation of 0.13, although its Pearson correlation of 0.61 is higher. Over the seven unlearning seeds of Table~\ref{tab:relearn}, the joint response has median Pearson and Spearman correlations of 0.50 and 0.23. At every one of those seeds it has a higher Pearson correlation than the per-entity block solve and a higher Spearman correlation than first-order evaluation. First-order evaluation matches the level of the change, with a mean predicted change of $+0.145$ against a realized $+0.159$. The exactly evaluated response predicts the ranking and not the level, with a mean predicted change of $-0.009$. The target $t_f$ therefore supplies the level, as Section~\ref{sec:method:pess} states, and the response supplies the ranking. Table~\ref{tab:kau_ablation_recovery_anc03} places the joint solve with exact evaluation first in leakage as well. First-order evaluation raises RER from 19.38 to 29.07, and the per-entity block solve raises it to 30.50.

\subsection{Leakage and retained utility against forgetting depth}\label{app:floor}

Leakage depends on how deeply a method forgets at release, so Figure~\ref{fig:depth} in Section~\ref{sec:exp} plots RER against forgetting depth. Depth is the forgetting rate FR, the forget-MRR drop at release as a percentage of the retrained oracle's. Each method is released at three depths by varying only its budget knob, under one update configuration per backbone. Figure~\ref{fig:depth} shows means over three unlearning seeds. The rung nearest each of the three bands, 40 to 60, 60 to 85 and 85 to 110 percent, is shown.  On TransE, from the shallowest to the deepest band the mean RER rises by about 59 points for ZeroFisher, 50 for \GraphDPO{} and 15 for \KAU{}. In the deepest band it is about 131 for ZeroFisher, 77 for \GraphDPO{} and 26 for \KAU{}. The mean ordering of the three methods holds in every band on every backbone. The baselines' RER grows with depth more than twice as fast as \KAU{}'s, so the gap widens toward the deepest band.

The figure is the reason every row of Tables~\ref{tab:kau_ablation_recovery_anc03} and~\ref{tab:ablation_recovery_depth} is released at \KAU{}'s forgetting depth. A method released deeper earns part of its leakage from that depth alone, so leakage is compared at equal FR.

\begin{figure}[htbp]
\centering
\includegraphics[width=\linewidth]{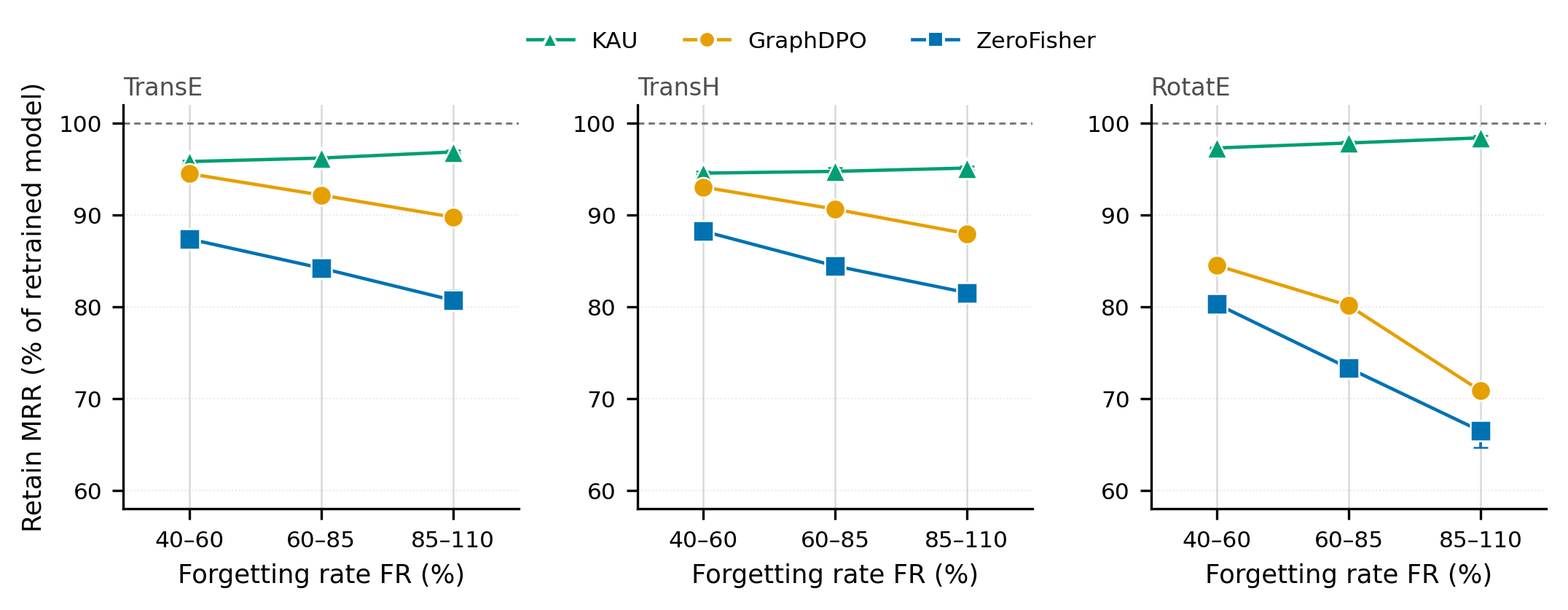}
\caption{Retained MRR at release against forgetting depth on FB15k-237, as a percentage of the gold model's retained MRR, for the releases of Figure~\ref{fig:depth}. Points average three unlearning seeds and bars span the seeds; the dashed line at 100 is the gold model. \KAU{} keeps more than 94\% of the gold model's retained MRR at every depth, whereas both baselines lose retained utility as they forget deeper.}
\label{fig:retain_depth}
\end{figure}

Forgetting deeper does not cost \KAU{} retained utility. Figure~\ref{fig:retain_depth} gives the retained MRR at release of the same releases as Figure~\ref{fig:depth}, as a percentage of the gold model's. \KAU{} keeps more than 94 percent of the gold model's retained MRR in every band on every backbone, and its retention rises slightly with depth. The baselines pay for depth in retained utility. From the shallowest to the deepest band, \GraphDPO{}, the stronger baseline, loses 4.7, 5.1 and 13.6 points on TransE, TransH and RotatE and ends at 89.8, 88.0 and 70.8 percent, and ZeroFisher loses 6.7, 6.7 and 13.8 points. At every depth \KAU{} therefore both leaks less and retains more.

\begin{figure}[ht]
\centering
\includegraphics[width=\linewidth]{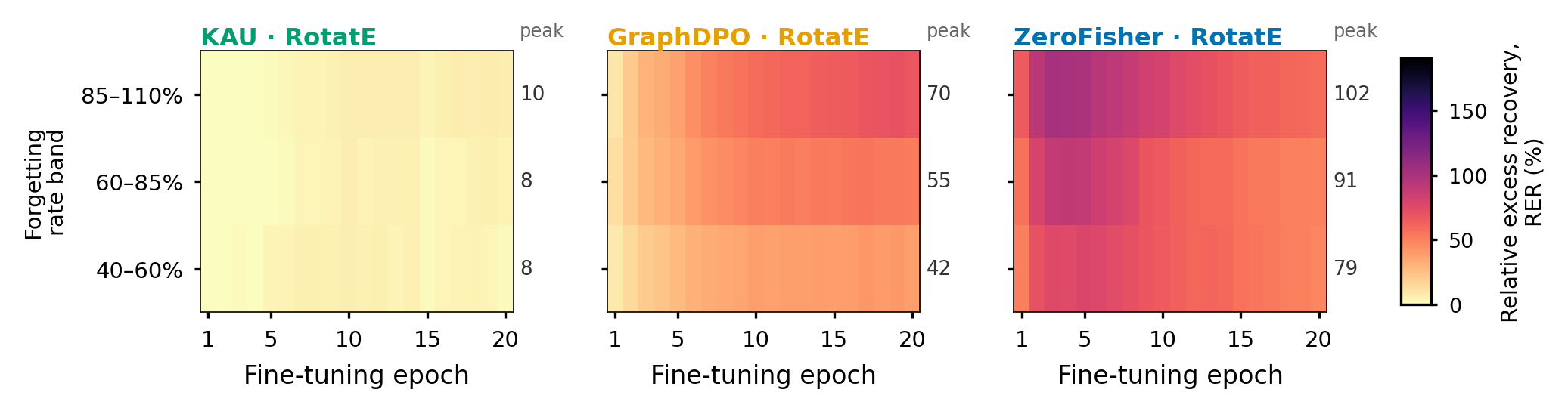}
\caption{RER along the update on FB15k-237 with RotatE, by forgetting-rate band at release, under the update configuration of Figure~\ref{fig:depth}. Each cell is the mean RER over three unlearning seeds at that epoch, and the number after each row is the mean over seeds of each seed's peak RER. An RER of zero is the retrained oracle's own recovery. The three panels share one colour scale.}
\label{fig:depth_heatmap_rotate}
\end{figure}

Figure~\ref{fig:depth_heatmap_rotate} follows the relearning epoch by epoch on RotatE, under the update configuration of Figure~\ref{fig:depth} and at three unlearning seeds.
ZeroFisher's RER peaks within the first five epochs, and \GraphDPO{}'s keeps rising until the last epochs.
In every band and at every seed, \KAU{}'s peak RER stays below 15, whereas the lowest peak of either baseline is 36.

\subsection{Ablation details}\label{app:kau_ablation}

\begin{figure}[htbp]
\centering
\includegraphics[width=0.86\linewidth]{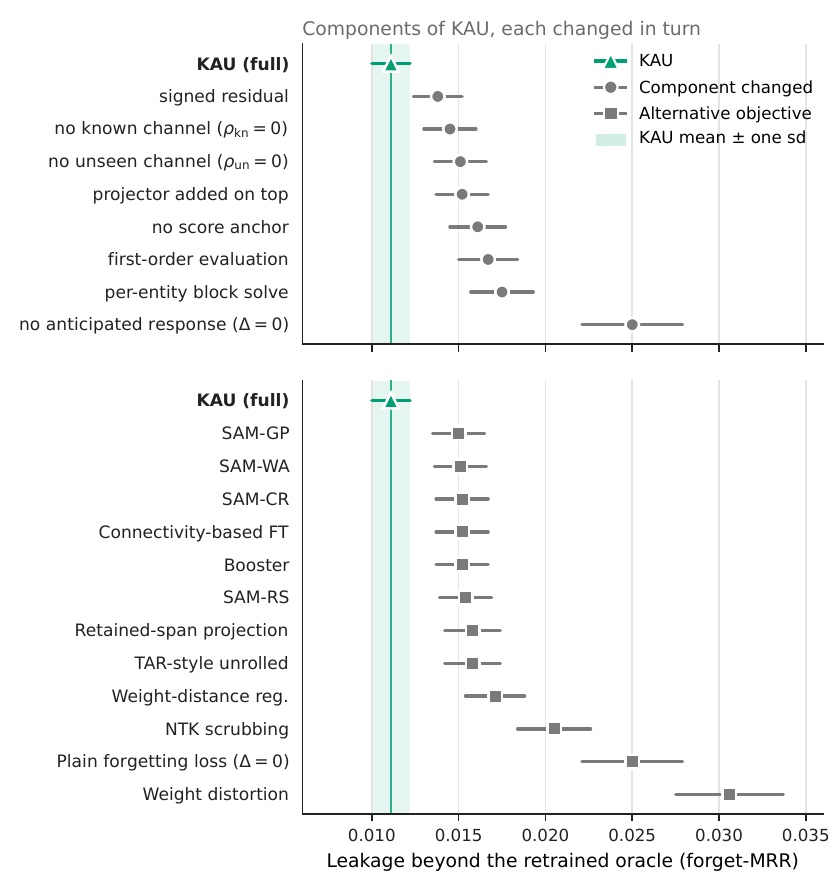}
\caption{Leakage of every row of Tables~\ref{tab:kau_ablation_recovery_anc03} and~\ref{tab:ablation_recovery_depth} on FB15k-237 with TransE, in the ablation setting of those tables. The upper panel holds the rows of Table~\ref{tab:kau_ablation_recovery_anc03} and the lower panel those of Table~\ref{tab:ablation_recovery_depth}. Markers are means over three unlearning seeds, and whiskers span one standard deviation over the seeds. The shaded band is \KAU{}'s mean plus or minus one standard deviation. Rows are sorted by leakage within each table.}
\label{fig:ablation_spread}
\end{figure}

Tables~\ref{tab:kau_ablation_recovery_anc03} and~\ref{tab:ablation_recovery_depth} are run in one ablation setting. Every variant starts from \KAU{}'s forgetting budget and keeps the retained pass, per-fact targets and release rule at \KAU{}'s selected TransE hyperparameters. The score-preservation weight is $\beta=0.3$, the ablation setting of Table~\ref{tab:hyper}. All rows, including \KAU{}, are released within a fixed band of forgetting rates, so that leakage is compared at equal forgetting, which is why \KAU{}'s values in these tables differ from those of Table~\ref{tab:relearn}. All rows face one update configuration of Table~\ref{tab:relearn} with three unlearning seeds and are read at the epoch of peak leakage on the report half. A variant that forgets less than \KAU{} at its budget is given a larger forgetting budget, set from its forgetting rate at release alone, so that it falls within the band. Figure~\ref{fig:ablation_spread} shows the leakage of every row of Tables~\ref{tab:kau_ablation_recovery_anc03} and~\ref{tab:ablation_recovery_depth} with its spread over the three seeds.

\begin{table}[htbp]
\centering
\caption{\KAU{} ablations on FB15k-237 with TransE, means over three unlearning seeds. Figure~\ref{fig:ablation_spread} gives both tables' seed spreads. Each row changes one component. All rows, including \KAU{}, are released within a fixed forgetting-rate band, so \KAU{}'s values differ from Table~\ref{tab:relearn}.}
\label{tab:kau_ablation_recovery_anc03}
\footnotesize
\setlength{\tabcolsep}{4pt}
\begin{tabular*}{\linewidth}{@{\extracolsep{\fill}}lrrr@{}}
\toprule
Variant & ExRec$\downarrow$ & Leak$\downarrow$ & RER$\downarrow$ \\
\midrule
\KAU{} (full) & 31.91 & 0.0111 & 19.38 \\
\midrule
no anticipated response ($\Delta{=}0$) & 48.60 & 0.0250 & 43.53 \\
first-order evaluation & 37.40 & 0.0167 & 29.07 \\
per-entity block solve & 42.24 & 0.0175 & 30.50 \\
signed residual & 30.78 & 0.0138 & 24.05 \\
no unseen channel ($\rho_{\mathrm{un}}{=}0$) & 37.55 & 0.0151 & 26.29 \\
no known channel ($\rho_{\mathrm{kn}}{=}0$) & 36.16 & 0.0145 & 25.28 \\
projector added on top & 31.20 & 0.0152 & 26.53 \\
no score anchor & 36.57 & 0.0161 & 28.06 \\
\bottomrule
\end{tabular*}
\end{table}

\paragraph{Variants of \KAU{}.}
\emph{No anticipated response} sets the response to zero, so that the forgetting loss acts on the current score. \emph{Response damped} raises the relative ridge from $10^{-3}$ to $10^{3}$, which shrinks the response to a small fraction of its size. \emph{First-order evaluation} replaces the score at the shifted rows by its linearisation around the current rows. \emph{Per-entity block solve} solves the response separately for each entity's block instead of jointly. It therefore counts the residual of a drawn edge between two involved entities once at each endpoint (Lemma~\ref{lem:jacobi}). \emph{Signed residual} takes the residual of every drawn edge with its sign, crediting the update with restoring lowered edges as well as raised ones. \emph{No pessimistic composition} applies the loss to the anticipated score alone rather than to the smaller of the current and anticipated scores. \emph{Per-relation gate} scales each edge's residual by its margin-violation probability; it binds on 4\% of the modelled update edges. \emph{No unseen channel} and \emph{no known channel} remove, respectively, the stand-in edges and the retained neighbours from the draw. \emph{No best-of-$Q$ stand-in choice} keeps the first plausible stand-in candidate instead of the best of four. \emph{Hinge of the expected draw} applies the loss to the response of the expected draw, with inclusion probabilities as weights, instead of averaging the loss over sampled draws. \emph{Ordered $k_{\max}$ cut} keeps the first $k_{\max}$ candidate edges in a fixed order instead of a random subset. The rows $\rho_{\mathrm{kn}}=1$ and $k_{\max}=16$ draw every retained edge and cap the draw at 16 edges, respectively. These two and the $Q{=}1$ row also draw fewer stand-in edges. \emph{No score anchor} sets the score-preservation weight to zero. \emph{Relational anchor} holds pairwise distances between rows instead of scores; it is released deeper than the gold model. \emph{No norm retraction} omits the rescaling of edited rows, \emph{projector added on top} applies the retained-span projector to \KAU{}'s update, and \emph{no forgetting term} runs the retained pass alone; it forgets less than half of what the gold model forgets and cannot be brought to \KAU{}'s depth. Every other forgetting rate lies within 9 points of \KAU{}'s. Table~\ref{tab:kau_ablation_full} gives the variants that Table~\ref{tab:kau_ablation_recovery_anc03} omits.

\begin{table}[htbp]
\centering
\caption{Remaining variants of \KAU{} on FB15k-237 with TransE, in the setting of Table~\ref{tab:kau_ablation_recovery_anc03}; the \KAU{} row is repeated for reference. Means over three unlearning seeds.}
\label{tab:kau_ablation_full}
\footnotesize
\setlength{\tabcolsep}{4pt}
\begin{tabular*}{\linewidth}{@{\extracolsep{\fill}}lrrrr@{}}
\toprule
Variant & ExRec$\downarrow$ & Leak$\downarrow$ & RER$\downarrow$ & $M_R\!\uparrow$ \\
\midrule
\KAU{} (full) & 31.91 & 0.0111 & 19.38 & 0.2860 \\
\midrule
response damped (relative ridge $10^3$) & 47.13 & 0.0238 & 41.45 & 0.2869 \\
no pessimistic composition & 32.60 & 0.0110 & 19.18 & 0.2854 \\
per-relation gate & 34.18 & 0.0110 & 19.12 & 0.2859 \\
no best-of-$Q$ stand-in choice ($Q{=}1$) & 37.90 & 0.0147 & 25.52 & 0.2857 \\
hinge of the expected draw & 31.98 & 0.0148 & 25.81 & 0.2857 \\
ordered $k_{\max}$ cut & 36.97 & 0.0155 & 27.02 & 0.2860 \\
$\rho_{\mathrm{kn}}=1$ & 33.92 & 0.0126 & 22.00 & 0.2864 \\
$k_{\max}=16$ & 38.56 & 0.0151 & 26.34 & 0.2856 \\
relational anchor & 33.49 & 0.0213 & 37.01 & 0.2868 \\
no norm retraction & 32.92 & 0.0121 & 21.00 & 0.2864 \\
no forgetting term & 77.39 & 0.0013 & 2.27 & 0.2830 \\
\bottomrule
\end{tabular*}
\end{table}

\begin{table}[htbp]
\centering
\caption{Alternative objectives replacing only \KAU{}'s forgetting step, in the setting of Table~\ref{tab:kau_ablation_recovery_anc03}. Means over three unlearning seeds.}
\label{tab:ablation_recovery_depth}
\footnotesize
\setlength{\tabcolsep}{4pt}
\begin{tabular*}{\linewidth}{@{\extracolsep{\fill}}lrrr@{}}
\toprule
Forgetting objective & ExRec$\downarrow$ & Leak$\downarrow$ & RER$\downarrow$ \\
\midrule
Plain forgetting loss ($\Delta{=}0$) & 48.60 & 0.0250 & 43.53 \\
Retained-span projection & 33.32 & 0.0158 & 27.51 \\
SAM-RS & 40.15 & 0.0154 & 26.85 \\
SAM-GP & 42.51 & 0.0150 & 26.09 \\
SAM-CR & 36.58 & 0.0152 & 26.55 \\
SAM-WA & 43.00 & 0.0151 & 26.30 \\
Weight-distance regularisation & 40.30 & 0.0171 & 29.73 \\
Connectivity-based fine-tuning & 42.31 & 0.0152 & 26.39 \\
TAR-style unrolled & 46.15 & 0.0158 & 27.54 \\
Booster & 43.73 & 0.0152 & 26.42 \\
NTK scrubbing & 55.20 & 0.0205 & 35.63 \\
Weight distortion & 53.56 & 0.0306 & 53.23 \\
\midrule
\KAU{} & 31.91 & 0.0111 & 19.38 \\
\bottomrule
\end{tabular*}
\end{table}

\paragraph{Alternative forgetting objectives.}\label{app:alternatives}
Each alternative in Table~\ref{tab:ablation_recovery_depth} replaces \KAU{}'s forgetting step and keeps everything else. The setting of each objective's own hyperparameters is given in brackets. \emph{Plain forgetting loss} is the variant of Table~\ref{tab:kau_ablation_recovery_anc03} without the anticipated response. \emph{Retained-span projection} projects the forgetting gradient on each edited entity row off the span of the score gradients of that entity's retained facts \citep{chen2024unsc,hoang2024pgu,zhou2025geometric}, with projectors rebuilt every five epochs (ridge $10^{-2}$). The \emph{sharpness-aware} rows follow \citet{fan2025sam}: SAM-RS averages the forgetting gradient over Gaussian perturbations of the edited rows ($\rho=10^{-4}$, one perturbation), SAM-GP adds a penalty on the norm of the forgetting gradient ($\rho=3\times10^{-5}$), SAM-CR adds a finite-difference curvature penalty ($w=10^{-6}$, $\rho=10^{-2}$), and SAM-WA writes an exponential average of the edited rows back once per epoch (decay $0.9$). \emph{Weight-distance regularisation} subtracts a multiple of the mean squared displacement of the edited rows from the original model \citep{siddiqui2025dormant} ($\gamma=10$). \emph{Connectivity-based fine-tuning} adds, for the retained and the forgetting loss, a barrier term at the midpoint between the current and the original rows \citep{lubana2023mechanistic} ($w=10^{-3}$). \emph{TAR-style unrolled} copies the entity table, runs Adam steps of the update's own loss on the retained facts of the current draw, and evaluates the forgetting loss at the displaced rows \citep{tamirisa2025tar}.
\emph{Booster} adds a penalty on the reduction of the retained loss that one normalised gradient step would obtain \citep{huang2025booster} ($\alpha=10^{-2}$). \emph{NTK scrubbing} is a one-shot edit \citep{golatkar2020forgetting}: for each deleted fact its two endpoint rows are moved along the component of the fact's score gradient orthogonal to the span of the retained facts' gradients until the fact reaches its target to first order ($\epsilon=0.32$), after which Fisher-scaled Gaussian noise is added at the largest scale on a ladder that keeps the retain-set MRR within 1\% of the noise-free edit's; no non-zero scale did, so the noise term is zero. \emph{Weight distortion} adds isotropic Gaussian noise to the edited rows ($\sigma=10^{-2}$) and then runs the retain fine-tuning through the retained pass, as in the original method \citep{siddiqui2025dormant}.

The forgetting rates at release lie between 88 and 100 percent of the gold model's, against 93 for \KAU{}.

\subsection{Why exact evaluation on piecewise-linear scores}\label{app:pwl}

The anticipated score is computed by adding the response to the rows and calling the score function again, rather than by adding the first-order term $\feat{f}^{\top}\resp$ to the current score. Two facts motivate this on any backbone. The score is not linear in the rows, because of the row transform and of the form of the score itself, and the displacement the update is predicted to make is not small on the scale of the coordinate differences from which the score is built. Appendix~\ref{app:predictor} measures the consequence, which is that the exactly evaluated response ranks facts by realized recovery better than the linear one.

On an $L_1$ score the size of the difference can be written down. Let $\vx=\bar\ve_h+\bar\vw_r-\bar\ve_t$ be the residual vector of a triple, so that $\dist{x}=\|\vx\|_1$, and let $\bm{D}$ be the displacement of that vector produced by the response. Lemma~\ref{lem:pwl}, applied coordinate by coordinate, reads
\begin{equation}\label{eq:pwl}
|x_j+D_j|-|x_j| \;=\; \mathrm{sign}(x_j)\,D_j \;+\; 2\big(|D_j|-|x_j|\big)_+\,\mathbf 1\big[\mathrm{sign}(D_j)\neq \mathrm{sign}(x_j)\big].
\end{equation}
Up to the normalization's second-order term, the first term is what a first-order evaluation sees. The second is non-negative and is present only in the coordinates where the displacement opposes the residual and exceeds it in size, that is, where the coordinate crosses zero. Whether it matters is a question about two scales, the size of a coordinate $|x_j|$ against the size of the displacement $|D_j|$ the response predicts. A first-order evaluation of the anticipated score is blind to the second term by construction, and evaluating at the shifted rows captures it at the cost of one further call to the score function.

\section{Extended related work}\label{app:related}

This appendix gives the full related work that Section~\ref{sec:related} summarizes.
It places \KAU{} among the relearning-robust unlearning methods and the unlearners for graphs.
Table~\ref{tab:related} records, for each study, the update it resists or measures and the object its objective reasons about.
It also records whether that object couples retained to forgotten data, whether it computes what the update's data do to the forgotten items, and how the study is evaluated.
Appendix~\ref{app:related:gaps} states what the table shows, and Appendix~\ref{app:related:scope} states the scope of this paper relative to these studies.
Appendices~\ref{app:related:graphs}, \ref{app:related:update}, \ref{app:related:robust} and~\ref{app:related:operator} expand Section~\ref{sec:related}.
They cover unlearning in knowledge graphs and graph networks, benign relearning, the relearning-robust methods and the tangent-kernel response.
Appendices~\ref{app:related:design} and~\ref{app:related:remaining} place the closest methods and the remaining studies one at a time.

\begin{landscape}
\begingroup
\scriptsize
\setlength{\tabcolsep}{3pt}
\setlength{\LTcapwidth}{\linewidth}
\makeatletter
\let\related@parboxrestore\@arrayparboxrestore
\def\@arrayparboxrestore{\related@parboxrestore\raggedright\let\\\tabularnewline}
\makeatother
\begin{longtable}{@{}p{2.0cm}p{1.2cm}p{2.7cm}p{2.9cm}p{2.3cm}p{2.3cm}p{1.5cm}p{1.6cm}p{2.0cm}@{}}
\caption{Relearning-robust unlearning methods, retained-span projection methods and unlearners for graphs, by what their objective reasons about and how they are evaluated. Update model: the update the study resists or measures, benign (no forgotten item in it) or forget-set (forgotten items in it), and whether it adapts to the method. Coupling: whether the objective relates retained to forgotten data, and through what. Data effect: whether the objective computes what training on the update's data does to the forgotten items. Reference: whether a retrained model is the reference. Matched: whether the unlearned and the retrained model receive the same fine-tune.}
\label{tab:related}\\
\toprule
Study & Domain & Update model & Object reasoned about & Coupling & Data effect & Reference & Matched & Cost \\
\midrule
\endfirsthead
\caption[]{Relearning-robust unlearning methods and unlearners for graphs (continued).}\\
\toprule
Study & Domain & Update model & Object reasoned about & Coupling & Data effect & Reference & Matched & Cost \\
\midrule
\endhead
\midrule
\multicolumn{9}{r}{Continued on next page}\\
\endfoot
\bottomrule
\endlastfoot
\multicolumn{9}{@{}l}{\emph{Relearning-robust unlearning methods}}\\
TAR \citep{tamirisa2025tar} & LLM & forget-set, on forget, retain and mixed corpora, adaptive only by adding fine-tunes that broke earlier versions & weights and residual-stream activations, entropy of one fixed forget batch along a simulated multi-step trajectory & no, retain loss at the weights before the simulated fine-tune & partial, simulated, one fixed batch, straight-through gradient & no & none & a simulated fine-tune per outer step, multi-GPU \\
Meta-unlearning \citep{gao2025meta} & diffusion & forget-set, on forget images, benign set as test only, not adaptive & weights, two whole-set gradients, a few unrolled inner steps & partial, one inner product of whole-set gradients & partial, forget-data fine-tune, unrolled & no & none & second-order autodiff through the inner steps \\
ResAlign \citep{li2025resalign} & diffusion & forget-set, sampled fine-tunes on a benign and harmful pool, benign at test, two adaptive attacks & weights, harmful loss at a simulated end point, implicit proximal Jacobian & no, retain as distillation at the current weights & partial, simulated, mean harmful loss at the end point & no & none & three simulated fine-tunes per outer step, about one GPU hour in total \\
MUDMAN \citep{sondej2025mudman} & LLM & forget-set, on the forget set, fixed rate, not adaptive & per-parameter gradients, retain momentum, a forked copy & partial, per-coordinate sign agreement & partial, first-order, interleaved fine-tune on the forget set & no, unedited model after the same relearning & none & several passes per loop, hours of search \\
CIR \citep{sondej2025cir} & LLM & forget-set, long fine-tune on most of the forget set, not adaptive & principal components of forget-set activations and gradients & no, retain step optional & no, a hypothesis that broken representations are repaired & no & none & seconds per fact \\
RepSelect \citep{sondej2026repselect} & LLM & forget-set, held-out same-domain data, an adaptive variant is tested & singular vectors of the accumulated forget-set gradient after a LoRA fine-tune & no, needs no retain set & partial, LoRA fine-tune on the forget set plus a proportionality assumption & no, base model & none & minutes per run, no search \\
SAM \citep{fan2025sam} & LLM & forget-set, few forget samples, benign in one figure, not adaptive & weight vector, forget-loss gradient and Hessian inside a norm ball & no, additive retain loss at the unperturbed weights & no, the perturbation is the unlearner's own forget gradient & no & none & about twice the base cost \\
ILU \citep{wang2025invariance} & LLM & benign, downstream tasks to convergence, forget relearning secondary, not adaptive & weight vector, one stationarity scalar per environment & no, additive terms & no, data enter through one scalar & no & none & several times the base cost, needs an auxiliary set \\
\citet{siddiqui2025dormant} & vision & benign, retain-only, one rate, not adaptive & whole weight vector, distance and loss barrier & no, norms are direction-blind & no & yes & yes, and the reference does not move & long unlearning runs \\
\citet{rosati2026limits} & LLM, diffusion & forget-set and adaptive, rate sweeps, second-order optimizers, layer insertion, no benign & weight-matrix spectrum, global over all data & no & no, data-blind & no & none & seconds \\
MCU \citep{xiao2026mcu} & LLM & forget-set, on most of the forget set, adaptive representation-matching variant, no benign & principal components of forget-set representations, isotropic kernel assumed & no, retain enters only the host's norm penalty & no, shared eigenstructure assumed & no, original model & none & one randomized SVD, rank chosen by full relearning runs \\
FRAG and FRP \citep{chen2026frag} & LLM & benign retain-only by default, also forget-only and mixed, rate, optimizer and epoch sweeps, not adaptive & per-weight importances from forget-to-retain channel activation ratios & partial, channel-support ratio, no relation among examples & partial, which channels the retain data activate, not how far outputs move & yes, retain-trained, identically fine-tuned & yes, but the ranked metric is the method's own score change & seconds to minutes, no training \\
UNDO \citep{lee2025undo} & LLM & forget-set, forget-set relearning, worst case over rates, not adaptive & weights and output distributions, noise and distillation & no & no & yes & yes, under forget-set relearning & a large fraction of retraining compute \\
\citet{yoon2026rethinking} & LLM & benign, topical and syntactic relearn sets, not adaptive & dataset-mean hidden states and gradients, as a diagnostic & partial, aggregate alignment, post hoc & no & yes, as a benignness check & yes, to certify benignness only & one paraphrase pass \\
\midrule
\multicolumn{9}{@{}l}{\emph{Retained-span projection and its origin}}\\
\citet{golatkar2020forgetting} & vision, small subsets & relearning time on the complete data, forget set included, no benign & sample-gradient tangent kernel, projector off the retain span, Fisher noise & yes, pairwise kernel, used to compute the edit & no, analysis stops at release & yes & yes, under relearning on the complete data & memory quadratic in the retained set \\
UNSC \citep{chen2024unsc} & vision & none, no fine-tune after release & per-layer retained activation spans per class, projector fixed at the original weights & partial, class activation span & no & yes, at release & no & one SVD per layer per class \\
GU \citep{zhou2025geometric} & LLM & none, no relearning run & retain-gradient span in the Adam metric & partial, one-step entanglement norm & no & no & no & small overhead \\
GPM \citep{saha2021gradient} & continual learning & none, GPM is itself the later optimizer & per-layer activation bases of past tasks & partial, span of old inputs & no & no & no & one SVD per layer per task \\
Adam-NSCL \citep{wang2021adamnscl} & continual learning & none, Adam-NSCL is itself the later optimizer & null space of per-layer feature covariance & partial, covariance span of old features & no & no & no & one SVD per layer per task \\
\midrule
\multicolumn{9}{@{}l}{\emph{Unlearners for graphs}}\\
\GraphDPO{} \citep{liu2026graphdpo} & KGE & none & entity and relation rows, one-hop boundary replay, pointwise distillation to a frozen original & partial, boundary replay & no & yes, at release & no & faster than fine-tuning, frozen copy doubles memory \\
ZeroFisher \citep{xiao2025zerofisher} & KGE & none & Fisher curvature of the full loss, $k$-hop restriction & partial, implicit through the Fisher & no & yes, at release & no & iterations tuned from $3$ to $1{,}550$ (Appendix~\ref{app:unlearning}) \\
Schema \citep{xiao2025schema} & KGE & none & pairwise difference alignment in the two-hop neighbourhood to a schema-matched subgraph & partial, pairwise geometry, no forget-score term & no & yes, at release & no & minutes \\
MetaEU \citep{xu2024metaeu} & KGE & none & embeddings generated from relations and neighbours, loss maximization on forget queries & no & no & yes, at release & no & not reported \\
GNNDelete \citep{cheng2023gnndelete} & GNN & none & layer-wise node representations on the $k$-hop subgraph, a deletion operator & partial, neighbourhood-influence term & no & yes, at release & no & much faster than retraining \\
GIF \citep{wu2023gif} & GNN & none & Hessian of the full loss, neighbour loss change within $k$ hops & yes, neighbour loss term, aimed at the retraining optimum & no & yes, at release & no & about a second per request \\
Projector \citep{cong2023projector} & linear GNN & benign fine-tune, as a utility check only & weight vector in the feature span, projection off deleted features & no, node dependencies folded into scalars & no & yes, at release & no, the retrained model is not fine-tuned & well under a second \\
\midrule
\KAU{} (this paper) & KGE & benign, $48$ update configurations shared by every method, the strongest against each method chosen on a held-out selection half, not adaptive & block-sparse tangent kernel over triples, per-fact closed-form response, score at the shifted rows & yes, per-fact kernel blocks over shared rows & yes, the linearized response to a fresh draw each step & yes & yes & one linear solve per step, 13 s per release on CoDEx-L \\
\end{longtable}
\endgroup
\end{landscape}

\subsection{What the table shows}\label{app:related:gaps}

No knowledge-graph or GNN unlearner has been evaluated for recovery of the forgotten facts under continued training.
\GraphDPO{}, ZeroFisher, Schema and MetaEU report the released model's forget and retained ranking and no training after the edit \citep{liu2026graphdpo,xiao2025schema,xiao2025zerofisher,xu2024metaeu}.
GNNDelete, GIF and the OpenGU benchmark report membership or ranking readouts of the released model and no continued training \citep{cheng2023gnndelete,wu2023gif,fan2025opengu}.
Projector fine-tunes after its edit, as a check that fine-tuning from the unlearned solution is cheaper than retraining, and reads utility from that fine-tune \citep{cong2023projector}.

Almost no relearning-robust unlearning method reports its gap to an identically fine-tuned retrained model.
TAR compares against an undefended model and a random baseline, MUDMAN against the unedited model after the same relearning, MCU against the original model and RepSelect against the base model \citep{tamirisa2025tar,sondej2025mudman,xiao2026mcu,sondej2026repselect}.
Meta-unlearning, CIR, SAM, ILU, the spectral method of \citet{rosati2026limits}, GU and UNSC report no retrained reference under the fine-tune \citep{gao2025meta,sondej2025cir,fan2025sam,wang2025invariance,zhou2025geometric,chen2024unsc}.
FRAG fine-tunes a retrained model identically but ranks methods by their own score change rather than by their distance to it \citep{chen2026frag}.
\citet{siddiqui2025dormant}, \citet{lee2025undo} and \citet{dehghani2026sourcefree} use the reference, with a reference that does not move under the fine-tune, a fine-tune on forget data, and a head-only audit with a frozen extractor respectively.
The reference matters here because the retrained oracle itself gains forget-MRR under the update on FB15k-237 with TransE (Figure~\ref{fig:curves}), so an absolute post-update score cannot separate a leak from legitimate learning.

No prior method writes in closed form what the update's data do to each forgotten item.
The update enters as a simulation, an assumption, or a data-free proxy.
TAR back-propagates the entropy of one fixed forget batch through a simulated fine-tune with a straight-through estimator \citep{tamirisa2025tar}.
Meta-unlearning unrolls a fine-tune on a subset of the forget data, MUDMAN interleaves a first-order fine-tune on the forget set, and RepSelect fits a LoRA fine-tune on the forget set and assumes the update's weight change is proportional to the accumulated gradient \citep{gao2025meta,sondej2025mudman,sondej2026repselect}.
MCU assumes the relearning distribution shares its dominant eigenstructure with the forget representation, FRAG reads channel activation statistics and needs no relearning run, and CIR rests on the hypothesis that broken representations are repaired \citep{xiao2026mcu,chen2026frag,sondej2025cir}.
SAM perturbs along the unlearner's own forget gradient, ILU reduces each environment to one stationarity scalar, \citet{siddiqui2025dormant} use a weight distance and \citet{rosati2026limits} a spectrum \citep{fan2025sam,wang2025invariance}.
None of these is closed form, none is per fact, and none uses a relation among examples.

Coupling between retained and forgotten data is absent or a single aggregate quantity.
Meta-unlearning reduces it to one inner product of whole-set gradients, MUDMAN to a per-coordinate sign test, GU to a one-step entanglement norm, and FRAG to a channel-support ratio \citep{gao2025meta,sondej2025mudman,zhou2025geometric,chen2026frag}.
TAR evaluates the retain loss at the weights before the simulated fine-tune, SAM and ILU add retain terms, the retain data of MCU enter neither its principal components nor its projector, RepSelect needs no retain set, and the retain step of CIR is optional \citep{tamirisa2025tar,fan2025sam,wang2025invariance,xiao2026mcu,sondej2026repselect,sondej2025cir}.
\citet{golatkar2020forgetting} alone carry pairwise couplings, through the tangent kernel between forget and retain samples, and they spend them on computing the scrubbing edit, after which the analysis ends.
A knowledge graph needs the pairwise object, because an entity row's retained gradient aggregates over every incident edge, so no scalar and no sign mask can say which neighbour edge restores which fact.

Projection off the retained span cancels the edit where the forgotten and the retained gradients share rows.
The appendix of GU records that the projected step vanishes when the forget gradient lies in the retain tangent space \citep{zhou2025geometric}.
Adam-NSCL states that zero singular values of the feature covariance cannot be guaranteed, and GPM reports that most of the gradient space is constrained after a handful of dissimilar tasks \citep{wang2021adamnscl,saha2021gradient}.
On a knowledge graph a forgotten triple's gradient shares its rows with every retained neighbour, so this case is the entity of high degree.
That is why the operator of Theorem~\ref{thm:operator} is ridge-shrunk rather than a projector, and why the loss acts on the anticipated score rather than on the update direction.
It is also why the retained-span projection, which approaches the second limit of Corollary~\ref{cor:soft} as its ridge vanishes, leaks more than \KAU{} at matched forgetting depth on TransE, $0.0158$ against $0.0111$ forget-MRR (Table~\ref{tab:ablation_recovery_depth}).

Benign relearning is measured by several methods and modelled by none of the unrolled or subspace methods.
Meta-unlearning states that self-destruction is meant to occur only under harmful fine-tuning and confirms that it does not occur on the benign set \citep{gao2025meta}.
MUDMAN, RepSelect, MCU, CIR and UNDO fine-tune on forget-domain data, and TAR includes retain-only fine-tunes in its sampled set but optimizes only their effect on the fixed forget batch \citep{sondej2025mudman,sondej2026repselect,xiao2026mcu,sondej2025cir,lee2025undo,tamirisa2025tar}.
ILU trains on a benign fine-tune and FRAG evaluates one, so benign relearning is modelled there, through a global stationarity scalar and through channel support, never through the response of specific retained examples \citep{wang2025invariance,chen2026frag}.

Resistance is often bought with utility or with self-destruction.
TAR lowers retained accuracy and accepts a low post-update retained accuracy by design, and meta-unlearning makes the retained knowledge self-destruct under the fine-tune \citep{tamirisa2025tar,gao2025meta}.
\citet{siddiqui2025dormant} find that the methods best at resisting relearning are those with the lowest test accuracy. The utility drop of FRP exceeds that of the unlearners it is compared with, and UNDO pays in general capability \citep{chen2026frag,lee2025undo}.
\KAU{} keeps the retained utility within $0.01$ of the oracle's on TransE, $0.2839$ against $0.2932$, where \GraphDPO{} and ZeroFisher lose $0.03$ and $0.06$ (Table~\ref{tab:main}).

The robustness of a method that samples fine-tunes is tied to the sampled distribution.
TAR reports sensitivity to parameter-efficient fine-tunes and to learning rates outside its sampled set and closes each hole by adding the offending fine-tune to that set, and meta-unlearning weakens against malicious fine-tuning as its number of inner steps grows \citep{tamirisa2025tar,gao2025meta}.
\KAU{} also draws from a class, but the draw is over the data of a benign recipe and the response to a draw is computed in closed form to first order, not sampled as a trajectory.

A method that only slows relearning is provably undone.
\citet{rosati2026limits} prove that a fine-tune designed against the method restores fast convergence at a cost linear in model size, and their sweep finds no earlier method that withstands standard fine-tuning once the learning rate varies.
\KAU{} relocates the edit rather than slowing convergence, and the results read each method under the configuration that leaks most against it on the selection half, at the epoch where its leakage peaks (Section~\ref{sec:exp}).

Curvature and norm-ball proxies do not describe the update on a graph.
The perturbation ball of SAM is small on the scale of the weights, its own results show its unlearning efficacy falling as the number of forget samples grows, and its mechanism, a bound on Hessian-vector norms, says nothing about which rows two triples share \citep{fan2025sam}.

\subsection{Scope relative to these studies}\label{app:related:scope}

The relearning-robust unlearning methods for language and diffusion models handle deep networks with no closed-form kernel, and several of them face fine-tunes that adapt to the method.
The closed form of Lemma~\ref{lem:response} holds for any differentiable score.
It is explicit in the rows and cheap enough for every step because the score depends on three parameter rows, which a deep encoder does not offer.
Our update is a benign recipe whose $48$ configurations are fixed in advance and shared by every method.
Each method is reported under the one that leaks most against it on the selection half of the probe (Section~\ref{sec:exp}).

Several of these methods face fine-tunes that hold forgotten data.
A fine-tune that includes forgotten edges re-inserts the data, the retrained oracle learns them too, and the gap then measures relearning speed against the oracle, which is the question of \citet{lee2025undo}.

Several methods are evaluated under multi-step trajectories, unseen update types, and the worst case over learning rates, epochs and seeds.
TAR simulates a multi-step fine-tune with momentum optimizers, and ILU fine-tunes to convergence and holds on unseen downstream tasks.
Four studies report the maximum over rates, epochs, seeds or a suite of fine-tunes \citep{deeb2024unlearning,lee2025undo,qi2025durability,che2025model}.
Here the response is the linearized response of the whole run of the update under frozen features (Appendix~\ref{app:rounds}), and Figure~\ref{fig:curves} shows each model's forget-MRR along the update, so the reader sees where the linearization stops predicting.
The never-trained-edge update is the unseen case, and an update outside the class, on edges that share no entity and no relation row with a forgotten fact, has zero first-order coupling to that fact.

Several studies compare more unlearners and more checkpoints.
\citet{siddiqui2025dormant} cover many unlearners across datasets and architectures, \citet{che2025model} many checkpoints, and OpenGU many graph unlearners \citep{fan2025opengu}.
This paper compares two published knowledge-graph unlearners, \GraphDPO{} and ZeroFisher, three relearning-resistant add-ons on \GraphDPO{} (Table~\ref{tab:relearn_addons}), and the alternative forgetting objectives of Table~\ref{tab:ablation_recovery_depth}, among them the retained-span projection.
The projector of that row approaches, as its ridge vanishes, the zero-ridge limit of the per-entity survival filter with every retained edge drawn and no stand-ins (Proposition~\ref{prop:release} and Theorem~\ref{thm:operator}).
It is the per-entity counterpart of the threshold at level zero in Corollary~\ref{cor:thresholds}, and its step direction is the matching limit of Corollary~\ref{cor:soft}.

\citet{siddiqui2025dormant} find that data unseen during training relearn less than seen data, and Appendix~\ref{app:attackvar} reads every method under an update built from never-trained edges alone (Table~\ref{tab:attacks}).
An unseen true edge lands on the same shared rows as a seen one with a larger residual, because the edit never saw it.

FRAG gives a predictor of relearning that needs no relearning run \citep{chen2026frag}.
The anticipated post-update score of \KAU{} is such a predictor, computed per fact, and it ranks forgotten facts by their realized recovery.
Its per-fact Pearson correlation with that recovery is $0.53$, and its Spearman correlation is $0.21$ (Appendix~\ref{app:predictor}).

GU projects in the optimizer's own metric and Adam-NSCL guarantees plasticity for the later task \citep{zhou2025geometric,wang2021adamnscl}.
The response here is Euclidean, and a degree-dependent ridge for Adam is stated as a hypothesis in Appendix~\ref{app:isotropy}.

\citet{golatkar2020forgetting} bound what a reader of the scrubbed weights can extract, and GNNDelete, OpenGU, Projector and UNSC audit by membership inference \citep{cheng2023gnndelete,fan2025opengu,cong2023projector,chen2024unsc}.
The metric here is a ranking gap against the updated oracle, and a membership readout on a released embedding is a separate question.

\GraphDPO{} and GNNDelete handle sequential requests, \GraphDPO{} runs on bilinear backbones as well, and ZeroFisher and Schema delete entities and relations \citep{liu2026graphdpo,cheng2023gnndelete,xiao2025schema,xiao2025zerofisher}.
\KAU{} is defined per request for any score that depends on three rows, and a new request reruns it with the new forget set.

ZeroFisher, RepSelect, CIR and the spectral method edit in seconds to minutes \citep{xiao2025zerofisher,sondej2026repselect,sondej2025cir,rosati2026limits}.
A \KAU{} step costs one linear solve over all involved rows and no simulated fine-tune (Appendix~\ref{app:cost}).

Meta-unlearning uses self-destruction as a deterrent and the spectral method uses slowing \citep{gao2025meta,rosati2026limits}.
The goal here is equivalence to the oracle under benign use, so the released model must remain fine-tunable at a retained utility close to the oracle's, which rules out both.

\citet{lee2025undo} show that a model matched to the oracle's outputs relearns faster than the oracle.
That corroborates the reading that a pointwise anchor to a reference, such as the distillation of \GraphDPO{} to the frozen original, confers no persistence, and it is why the score-preservation term of \KAU{} holds the scores of retained edges rather than the rows (Proposition~\ref{prop:anchor}).

\subsection{Unlearning in knowledge graphs and graph networks}\label{app:related:graphs}

Exact unlearning retrains the model on the retained data \citep{cao2015towards,bourtoule2021machine}.
Approximate unlearning edits the released model instead.
One branch takes a single step that approximates the effect of removing the forgotten examples, with the inverse Hessian or its Fisher approximation \citep{koh2017understanding,guo2020certified,izzo2021approximate,sekhari2021remember,golatkar2020eternal}.
The other optimizes an objective for a few steps on the forgotten examples, sometimes with retained ones \citep{jang2023knowledge,zhang2024npo}.

The object of the edit depends on the model.
In the translational and rotational backbones we study (Section~\ref{sec:setup}), the trainable parameters are the entity and relation tables.
Their score function has no weights of its own \citep{bordes2013translating,wang2014transh,sun2019rotate}.
A row edit changes the scores of the facts at that row alone, so the change and its cost can be confined to the neighbourhood of the deletion.
A graph neural network instead computes every node's representation from shared encoder weights by message passing.
An edit to those weights changes every representation, so the collateral change, and the recomputation needed to control it, grow with the graph.
Knowledge-graph unlearners therefore edit the tables themselves, and they differ in how much of the tables they update.

\citet{liu2026graphdpo} state the task as an update of the two tables, and their method \GraphDPO{} updates both tables.
It trains a preference objective that prefers a reconstructed alternative over each forgotten triple, with the original model as reference.
It also replays the retained triples adjacent to the forgotten ones.
ZeroFisher \citep{xiao2025zerofisher} applies an influence update in which a Fisher matrix replaces the Hessian and zeroth-order estimates replace the gradients.
It iterates that update for a tuned number of steps (Appendix~\ref{app:unlearning}).
It restricts the update to the rows of the local neighbours of the forgotten edges, because the cost of the full update grows with the size of the graph.
Schema \citep{xiao2025schema} re-embeds the two-hop neighbourhood of the deleted entities so that it matches a subgraph with the same schema and without the deleted facts.
MetaEU \citep{xu2024metaeu} meta-learns a shared generator that regenerates the rows of the forgotten entities from their relations and neighbours.

Graph-neural-network unlearners edit the encoder or its output representations.
GNNDelete inserts a deletion operator on the node representations of the affected subgraph and freezes the encoder \citep{cheng2023gnndelete}.
GIF applies an influence update to the encoder weights with a term for the affected neighbours \citep{wu2023gif}.
Projector projects the weights of a linear encoder off the deleted node features \citep{cong2023projector}, and OpenGU benchmarks such methods \citep{fan2025opengu}.

These knowledge-graph and graph-network unlearners use graph structure as the scope of the edit or as a preservation term for neighbours.
None has a term that computes what training on those neighbours does to the forgotten score.
They report the ranking the released model gives before any fine-tune (Appendix~\ref{app:related:gaps}).
We extend that reading and ask what the edit is worth once training resumes.

\subsection{Benign relearning and the update model}\label{app:related:update}

An unlearned model gives back what it was asked to forget once training resumes.
In language models, fine-tuning on data that contains none of the forgotten answers restores them \citep{hu2025jogging,lucki2025adversarial,yoon2026rethinking}.
\citet{hu2025jogging} are the origin of benign relearning and of the reading that unlearning may obfuscate rather than delete.
We transplant that setting to knowledge-graph embeddings, score it against a retrained model that receives the same fine-tune, and replace an informal correlation between forgotten and retained data with a closed-form account of the linearized response of the fine-tune.
\citet{deeb2024unlearning} show that fine-tuning on an accessible subset of the forget set restores held-out forgotten facts, and they name the absence of a reliable reference as the open problem.
Our update holds no forgotten fact at all, and the reference is the retrained model after the same update.

Recovery grows as the fine-tune proceeds, so evaluations read the worst point over its steps and learning rates \citep{deeb2024unlearning,lee2025undo}.
Others read the worst case over seeds, configurations and update recipes \citep{qi2025durability,che2025model}.
Section~\ref{sec:exp} follows both, reading the epoch of peak leakage and the worst update configuration, but averages over seeds.

\citet{siddiqui2025dormant} establish the reading we adopt.
They show that retain-only fine-tuning restores forget accuracy for most vision unlearners while a retrained model stays robust, they define tamper resistance as the requirement that the relearned unlearner not beat the relearned retrained model, which is the leakage of \eqref{eq:leak} read against the oracle's zero here, and they report that unseen data relearn less.
Two things differ on a knowledge graph.
The retrained model itself recovers under the update (Figure~\ref{fig:curves}), so the reference is not a fixed point.
And an update built only from never-trained edges is read for every method in Appendix~\ref{app:attackvar} (Table~\ref{tab:attacks}).

\citet{yoon2026rethinking} attribute benign relearning in language models to syntactic alignment between the relearning data and the forgotten targets, and they defend by paraphrasing the forget set.
A triple cannot be paraphrased, and the alignment between a forgotten fact and its retained neighbours is exact, which removes the data-side remedy and forces a storage-side one.
\citet{suriyakumar2025unstable} show the same failure in diffusion models, where fine-tuning on concepts disjoint from the erased one resurrects it, and they lower-bound the leaked gradient mass by the overlap of the two spans.
We sharpen that observation into an objective in which the edit is chosen against the overlap rather than measured after the fact.

\citet{dehghani2026sourcefree} recover forgotten classes with no forget, retain or auxiliary data, by probing the representation space and updating the head alone, and they measure excess recoverability relative to a retrained model under an identical audit.
We follow that convention and report the leakage beyond the retrained oracle in MRR points and as a share of the erased knowledge, beside the forgetting at release, because a method can buy one with the other (Section~\ref{sec:exp}).

\subsection{Relearning-robust unlearning methods}\label{app:related:robust}

Three families anticipate the update, and they differ in whether they simulate it, replace it by a proxy, or choose the subspace the edit occupies.
Appendix~\ref{app:related:gaps} sorts the same methods by how the update enters their objective, as a simulation, an assumption or a data-free proxy.

Unrolled methods anticipate the update by simulation.
They simulate the update inside the objective and evaluate the forgetting loss at the simulated weights.
Meta-unlearning takes the gradient through the simulation exactly \citep{gao2025meta}, and TAR takes it by a straight-through estimator \citep{tamirisa2025tar}.
ResAlign does similarly, taking the gradient implicitly through a proximal model of the simulation's end point \citep{li2025resalign}.
MUDMAN does not differentiate through the simulation.
It evaluates the forgetting gradient at a simulated copy and applies it to the original weights \citep{sondej2025mudman}.
\KAU{} replaces the simulation by the closed-form linearized response of the update for a shallow relational kernel.

Proxy methods replace the update by a cheaper quantity that does not describe the update itself.
SAM penalizes the sharpness of the forgetting loss under weight perturbations over a finite ball \citep{fan2025sam}.
ILU penalizes the variation of the loss across auxiliary datasets, a stationarity penalty with one scalar per dataset \citep{wang2025invariance}.
\citet{siddiqui2025dormant} regularize a weight-space distance to the original model.
\citet{rosati2026limits} control the weight spectrum and show that a fine-tune designed against any method that only slows recovery undoes it.
That result, detailed in Appendix~\ref{app:related:gaps}, is why we report the gap along the fine-tune and not only at its end.
Relearning-speed measures \citep{lee2025undo,tran2026efficient} are the unpaired version of that reading.

Subspace methods, the third family, choose where the edit lives.
Retained-span projection places the edit off the span of the retained gradients to protect utility \citep{golatkar2020forgetting,chen2024unsc,zhou2025geometric}.
Slow-direction methods place the edit in directions that an update restores slowly.
MCU and CIR take those directions from the principal components of the forget set's representations and activations respectively \citep{xiao2026mcu,sondej2025cir}.
RepSelect takes them from the weight gradient that a low-rank fine-tune on the forget set accumulates \citep{sondej2026repselect}.
FRAG places the edit on weights whose input channels the retained data barely activate \citep{chen2026frag}.
Appendix~\ref{app:related:operator} traces the retained-span projector to continual learning and discusses MCU, and Appendix~\ref{app:related:design} discusses FRAG and RepSelect.

These methods are built for deep networks.
Each reads its subspace or its proxy off the network's weights and activations, or runs one simulated update.
None of these quantities says which training example moved which parameter.
A deep model does not expose that attribution in closed form.
Carried over to a knowledge-graph embedding, they leave its structure unused.
An unrolled method returns the rows after one simulated update and no expression for which edge moved which row.
A proxy method reads sharpness, weight distance or spectrum off the whole table, and those quantities say nothing about which edges drive which rows.
A projection method projects off the retained-gradient span over all rows rather than off the span of the update's edges at the shared rows.
\KAU{} instead writes the update's effect on each forgotten fact in closed form, as a function of drawn edges the update may train on, and differentiates through it.
Appendix~\ref{app:related:design} reads both projection families through the survival filter of Section~\ref{sec:theory}, the slow-direction ones on a spectrum estimated from the forget or retained data.

\subsection{The tangent-kernel response and its origin}\label{app:related:operator}

The function \KAU{} differentiates through is the linearized response of training, and it can be built up from one gradient step.
Take one gradient step on one example.
To first order, the model's output on a second example moves in proportion to the inner product of the two examples' output gradients.
That inner product is the tangent kernel \citep{jacot2018neural}, and it says how much training on one example moves the prediction for another.
Over a run of small steps the same rule holds at every step, with the kernel held at the current parameters.
To first order the total change in any output is then the solution of a ridge-regularized linear system in that kernel (Section~\ref{sec:method:response}).
The ridge term stands in for the finite length of the run.
In a knowledge-graph embedding a triple's score depends on three rows.
The kernel between two triples is then an explicit expression in the rows of the two triples, and it is zero unless they share an entity or a relation.
The kernel therefore names, for each forgotten fact, which trained edges move it and through which rows.
That is the attribution the methods of Appendix~\ref{app:related:robust} lack.
The effect of the update on a forgotten fact can then be written down as a linear system in that kernel rather than simulated.

The linearized cross-task response we differentiate through is not ours.
\citet{doan2021theoretical} and \citet{bennani2020generalisation} derive it for continual learning, where one phase of training moves the function on an earlier phase's data through the tangent kernel.
Both use it to bound the resulting interference.
\citet{hidekel2026catastrophic} give a recent function-space account of the same motion.
\citet{malladi2023kernel} describe the fine-tuning of a pretrained model by the same kernel.
\citet{li2025resalign} differentiate a forgetting loss through a proximal model of a simulated fine-tune, after \citet{rajeswaran2019imaml}.
What is new here is the use we make of it.
The second phase is the downstream fine-tune, and the residual that drives it on the retained edges is created by the unlearning edit instead of being given by the data.
The resulting closed form is placed inside an objective the unlearner minimizes over a class of draws.

\citet{golatkar2020forgetting} run the same kernel algebra in the opposite direction.
They derive a one-shot scrubbing update from the linearized dynamics, project the forgetting gradient off the span of the retained gradients, add Fisher-shaped noise, and bound what a reader of the scrubbed weights can extract.
Their kernel couples each forget sample to each retain sample pairwise, and that coupling is spent on computing the edit.
Their analysis ends at the scrubbed weights, and their relearning readout fine-tunes on the complete data, forget set included.
Ours begins with the benign fine-tune after release.
We use the same algebra with the roles inverted, so that the kernel predicts the update instead of computing the edit.
Corollary~\ref{cor:soft} identifies their projector, restricted to one entity row, with the zero-ridge limit of our operator once the unseen channel is switched off.
Read through the survival filter of Corollary~\ref{cor:thresholds}, a projector introduced to protect utility also marks the edits that continued training on retained data cannot undo to first order.

The projector off the span of earlier data has a second origin in continual learning.
GPM keeps a new task's gradient off the activation bases of past tasks and Adam-NSCL keeps it in the approximate null space of their feature covariance \citep{saha2021gradient,wang2021adamnscl}.
Both are optimizers of the later phase, so their invariance holds while they control the update, and the unlearners that reuse the construction \citep{chen2024unsc,zhou2025geometric} protect the retained loss from the edit.

\citet{xiao2026mcu} give the closest statement about which components a fine-tune restores.
Under an isotropic-kernel linearization, an edit placed in a high-variance component of the representation is reversed at a rate proportional to that variance, which is what their minor-component method exploits.
The isotropic kernel drops exactly the off-diagonal blocks that carry the coupling in an embedding model, where the kernel between two triples is nonzero only when they share a row.
Our ridge $\ridge_v$ plays a comparable role in the eigenvalues of a block-sparse graph-local kernel, and the subspace we avoid is the span of the update's data rather than the forget representation's principal directions.
In the survival filter of Proposition~\ref{prop:release} their minor components correspond to the directions of small eigenvalue.
Confining the edit to such directions is a hard threshold of the filter (Corollary~\ref{cor:thresholds}).
Their shared-eigenstructure assumption is what lets the spectrum of the forget representation stand for the spectrum of the update.

\subsection{Methods closest in design}\label{app:related:design}

\citet{chen2026frag} argue that robustness is governed by where an update lives rather than by how far it moves, and they give a score computed per checkpoint that needs no relearning run.
Their pruning rule places the edit on weights that the retain data barely activate, so that a retain-only fine-tune leaves them inert.
In a knowledge graph the rows a forgotten fact uses are the rows its retained neighbours use, so no row is inert under a retain-only update.
The placement question then has to be asked inside the span of an entity's edges rather than across channels.
For a shallow embedding model that question can be made exact, because the span of the update's data at an entity is a closed-form graph-local tangent space.
The response goes one step further by evaluating the score at the shifted rows instead of along the tangent.

\citet{sondej2026repselect} confine the edit to the low-variance complement of the forget gradient's principal subspace and summarize the principle as selectivity rather than update strength.
We share the idea of storing suppression outside the subspace a later fine-tune reaches.
We define that subspace by drawn edges the update may train on, through a kernel available in closed form.
In the linearized model the anticipated loss is the plain forgetting objective on the part of the edit that survives the update (Corollary~\ref{cor:composition}).
Projection onto the complement is the zero-ridge threshold of that survival filter, a consequence instead of a starting point (Corollary~\ref{cor:thresholds}).

\citet{yin2026margin} calibrate a one-sided loss against a reference model's margins with a budget-bounded certificate.
Our loss is one-sided in the same way and is applied to the anticipated post-update score, and its per-fact targets are computed from quantiles of corrupted variants under the original model, so it needs no access to the retrained model.

These three, together with the retained-span methods above, share one object with us and use it differently.
A null space enters prior work in three roles.
It is a fixed layer operator that completes an ill-posed inverse \citep{yin2025rangenull,fang2025alphaedit}.
It is the null space of retained features or gradients that keeps a later update off earlier data in continual-learning optimizers \citep{saha2021gradient,wang2021adamnscl} and that protects utility from the unlearning edit \citep{golatkar2020forgetting,chen2024unsc,hoang2024pgu,zhou2025geometric}.
It is the low-variance part of the forget representation that a fine-tune restores slowly \citep{xiao2026mcu,sondej2026repselect,sondej2025cir}.
Ours is the complement of the span of the update's data, chosen for persistence.
In the linearized model it is the set of edits the update leaves untouched (Proposition~\ref{prop:release}).
Read through the survival filter, the second role is its hard threshold at level zero, restricted to the involved entities' rows, when the draw is exactly their retained edges.
The third has the form of a hard threshold at a positive level, taken on a proxy spectrum (Corollary~\ref{cor:thresholds}).
In the linearized model under a uniform ridge, \KAU{}'s gradient instead weights each direction by its surviving fraction, computed from the spectrum of its own update draws.
Unseen edges in the update can move edits that retained-span projection preserves, whereas \KAU{}'s draws include stand-ins for those edges (Section~\ref{sec:method:draw}).
Where the forgotten gradient lies inside the retained span, a hard projector removes the whole edit, which the appendix of GU records as a local trade-off \citep{zhou2025geometric}.
On a graph that is the entity of high degree.
There the cap $k_{\max}$ keeps every draw below full rank, and the ridge of Theorem~\ref{thm:operator} keeps the operator from annihilating the forgetting direction inside the drawn span (Section~\ref{sec:method:response}).

\subsection{Remaining placements}\label{app:related:remaining}

Three further threads touch the mechanism.
The image of an alignment edit that is pulled back toward its origin is due to \citet{ji2025resist}.
\citet{arora2019theoretical} and \citet{heo2021adamp} study how a growing row norm shrinks an effective learning rate.
The retraction rule of Appendix~\ref{app:retraction} removes this effect.
Studies of when an unlearned model is reversible \citep{ha2026blind,chang2026resist,xu2026unlearning} ask the same question at the level of the loss landscape.

\end{document}